\documentclass{bmvaannals}

\usepackage{amssymb}

\usepackage{amsthm}

\usepackage{amsfonts}
\usepackage[cmex10]{amsmath}

\usepackage{dsfont}
\usepackage{upgreek}

\usepackage[utf8]{inputenc} % set input encoding (not needed with XeLaTeX)

\usepackage{graphicx}
\makeatletter
\newcommand{\figcaption}{\def\@captype{figure}\caption}
\newcommand{\tabcaption}{\def\@captype{table}\caption}
\makeatother

\usepackage{float}

\usepackage{subfigure}
\usepackage[vlined,titlenumbered,algoruled]{algorithm2e}

\usepackage{colortbl}
\usepackage{ctable}
\newcolumntype{+}{>{\global\let\currentrowstyle\relax}}
\newcolumntype{^}{>{\currentrowstyle}}
\newcommand{\rowstyle}[1]{\gdef\currentrowstyle{#1}#1\ignorespaces}
\newcommand{\PreserveBackslash}[1]{\let\temp=\\#1\let\\=\temp}
\newcolumntype{C}[1]{>{\PreserveBackslash\centering}p{#1}}
\newcolumntype{R}[1]{>{\PreserveBackslash\raggedleft}p{#1}}
\newcolumntype{L}[1]{>{\PreserveBackslash\raggedright}p{#1}}

\usepackage{url}

\usepackage{booktabs} % for much better looking tables
\usepackage{array} % for better arrays (eg matrices) in maths
\usepackage{paralist} % very flexible & customisable lists (eg. enumerate/itemize, etc.)
\usepackage{verbatim} % adds environment for commenting out blocks of text & for better verbatim
\title{Numerical Methods for Coupled Reconstruction and Registration in Digital Breast Tomosynthesis}
\author{Guang~Yang, John~H.~Hipwell, David~J.~Hawkes and Simon~R.~Arridge\\*[0.5\baselineskip] Centre for Medical Image Computing, Department of Computer Science and Medical Physics, University College London, Gower Street, London, UK.\\ \email{g.yang@cs.ucl.ac.uk} \\ \url{www.cs.ucl.ac.uk/people/G.Yang.html}}
\papernumber{2012}{x}{2012-00xx}
\runninghead{Yang et al.}{Numerical Coupled Reconstruction and Registration in DBT}

\begin{document}
\maketitle


%______________________________________________________________________________________________________________________
% Abstract
\begin{abstract}
Digital Breast Tomosynthesis (DBT) provides an insight into the fine details of normal fibroglandular tissues and abnormal lesions by reconstructing a pseudo-3D image of the breast. In this respect, DBT overcomes a major limitation of conventional X-ray mammography by reducing the confounding effects caused by the superposition of breast tissue. In a breast cancer screening or diagnostic context, a radiologist is interested in detecting change, which might be indicative of malignant disease. To help automate this task image registration is required to establish spatial correspondence between time points. Typically, images, such as MRI or CT, are first reconstructed and then registered. This approach can be effective if reconstructing using a complete set of data. However, for ill-posed, limited-angle problems such as DBT, estimating the deformation is complicated by the significant artefacts associated with the reconstruction, leading to severe inaccuracies in the registration.

This paper presents a mathematical framework, which couples the two tasks and jointly estimates both image intensities and the parameters of a transformation. Under this framework, we compare an iterative method and a simultaneous method, both of which  tackle the problem of comparing DBT data by combining reconstruction of a pair of temporal volumes with their registration.

We evaluate our methods using various computational digital phantoms, uncompressed breast MR images, and in-vivo DBT simulations. Firstly, we compare both iterative and simultaneous methods to the conventional, sequential method using an affine transformation model. We show that jointly estimating image intensities and parametric transformations gives superior results with respect to reconstruction fidelity and registration accuracy. Also, we incorporate a non-rigid B-spline transformation model into our simultaneous method. The results demonstrate a visually plausible recovery of the deformation with preservation of the reconstruction fidelity. 

%In addition, due to the usage of two temporal data sets, the unified reconstruction can be interpreted as an intermediate step in the detection of change in the breast.
\end{abstract}

%______________________________________________________________________________________________________________________
% Keyword
%\begin{keyword}
%Image reconstruction \sep image registration \sep digital breast tomosynthesis \sep decoupled optimisation
%% \sep limited angle tomography \sep inverse problem \sep fully coupled optimisation \sep decoupled approach
%%% keywords here, in the form: keyword \sep keyword
%
%%% MSC codes here, in the form: \MSC code \sep code
%%% or \MSC[2008] code \sep code (2000 is the default)
%\end{keyword}

%% \pagewiselinenumbers % Uncomment this one only
%\linenumbers

%______________________________________________________________________________________________________________________
% Section 1: Introduction
\section{Introduction}
\label{sec:introduction}

\noindent Limited angle transmission tomography, {\it i.e.,} {\it tomosynthesis}, is playing an increasingly significant  research role across a wide range of clinical imaging tasks, including coronary angiography, cerebral angiography, and chest, breast, dental and orthopaedic applications \citep{DobbinsIII2003}.

Digital Breast Tomosynthesis (DBT) involves acquiring a small number of low dose X-ray images, over a limited angle, and reconstructing this data into a pseudo-3D image of the breast. This offers potential sensitivity and specificity gains to be made over conventional X-ray mammography in the management of breast cancer, by reducing the confounding effects associated with superimposed breast tissue. Increased sensitivity would increase survival rates, which are known to be associated with early  detection of the disease, whilst increased specificity would reduce recall rates, the associated patient anxiety and clinical costs \citep{Poplack2007, Gur2009, Spangler2011}.

In a breast cancer screening or diagnostic setting, radiologists routinely compare conventional current and prior mammograms to detect suspicious changes that might be indicative of malignancy. The workflow in which DBT would be used clinically, involves two key tasks: reconstruction, to generate a 3D image of the breast, and registration, to enable images from different visits to be compared, as is routinely performed by radiologists working with conventional mammograms. In established medical image modalities these tasks are normally performed sequentially; the images are reconstructed and then registered. In this paper, we hypothesise that, for DBT in particular, combining the optimisation processes of reconstruction and registration into a single algorithm will offer benefits for both tasks. Based on this hypothesis, we have devised a mathematical framework to combine these two tasks iteratively and simultaneously, and have implemented both affine and non-rigid B-spline transformation models as plug-ins. By applying our algorithm to various simulated data, we demonstrate the success of our method in terms of both reconstruction fidelity and in the registration accuracy of the recovered transformation parameters.

This paper is organised as follows. Section \ref{sec:related_work} surveys previous studies on tomographic reconstruction and image registration techniques, which are applicable to DBT. In Section \ref{sec:conventional_method}, we briefly recapitulate the {\it conventional sequential method}. Section \ref{sec:iterative_method} describes our {\it iterative reconstruction and registration method} and in Section \ref{sec:simultaneous_method}, we propose a {\it simultaneous method} to solve the fully-coupled reconstruction and registration problem. Section \ref{sec:results} describes the experimental results obtained, and this is followed by the discussion, Section \ref{sec:discussion}, and conclusion, Section \ref{sec:conclusion_and_perspectives}.

%______________________________________________________________________________________________________________________
% Section 2: Related Work
\section{Related Work}
\label{sec:related_work}

\subsection{Tomographic Reconstruction in DBT}

The development and performance of algorithms for the reconstruction of DBT have been extensively investigated over the last two decades. Most existing tomographic reconstruction algorithms fall into four categories: back-projection (BP) \citep{Kak2001, Herman2010} based methods including filtered back-projection (FBP); algebraic reconstruction techniques (ART) \citep{Kak2001, Herman2010} such as simultaneous algebraic reconstruction technique (SART) \citep{Mueller1999}; least squares (LS) based optimisation methods \citep{Fessler1994} and maximum likelihood (ML) techniques \citep{Shepp1982, Hudson1994}.

BP-based algorithms are also classified as analytical or transform methods. They are naturally simple and operate by smearing line integral values of the forward projections back into the image volume. In \citep{Kolitsi1992} Kolitsi et al. carried out an early study of BP-based digital tomosynthesis (DTS) reconstruction. They achieved an optimised efficiency by dividing the reconstruction process into discrete groups of pixels rather than performing a pixelwise operation. The traditional shift-and-add (SAA) method and the BP method are equivalent excepting a spatial scaling factor in the context of DTS reconstruction \citep{Wu2004}. This equivalence is only valid when the motion of the X-ray focal-spot is parallel to the detector, {\it i.e.,} a linear motion at a fixed height above the detector. One of the major disadvantages of the BP method is that the reconstructed images are over-smooth. The FBP method, which is the most widely-used method in parallel beam tomographic reconstruction, is a means of correcting this blurring effect. In the early 1990s, Matsuo et al. \citep{Matsuo1993} proposed a reconstruction method that utilised a 3D convolution process with an inverse filter function. This process was analytically derived from the point spread function of the projection geometry, and was well adapted to both phantom experiments and clinical evaluations. Stevens et al. \citep{Stevens2001} devised a filtering technique to blur out-of-plane objects whilst preserving in-plane features using a circular tomosynthesis setup. Recent studies on FBP are mainly divided into two categories: new filter designs and hardware acceleration, {\it e.g.,} field-programmable gate array (FPGA), graphics processing unit (GPU) and others. For example, Mertelmeier et al. \citep{Mertelmeier2006} published the filter design for their FBP reconstruction of DBT, and in \citep{Yan2007} Yan et al. adopted GPU programming for high performance DTS reconstruction using commercial PC graphics hardware.

Unlike one-step BP and FBP algorithms, iterative methods are deliberately modelled and mathematically complex. They recursively update the reconstructed estimation until the model reaches convergence according to a given criteria, {\it e.g.,} objective function tolerance, optimised value tolerance, or maximum number of iterations. ART, LS or Maximum Likelihood \citep{Dempster1977} (ML) algorithms can be used to build the model and instantiate the objective function mathematically. The proponents of iterative methods claim superior reconstruction accuracy compared to analytical methods {\it e.g.,} \citep{Wu2004, Zhang2006}; but their higher computational cost has been a major impediment to their adoption in commercial systems. Wu et al. \citep{Wu2004a} developed an iterative ML based method to reconstruct DBT using parallel computing. This method reduced the execution time from 187 minutes using a CPU to 6.5 minutes without reducing restoration quality. In \citep{Zhang2006} Zhang et al. concluded that both the SART and ML-convex methods increased the contrast and edges of high-contrast features, but decreased the signal to noise ratio. In addition, Kastanis et al. \citep{Kastanis2008} and Sidky et al. \citep{Sidky2009} implemented total variation based reconstruction methods for DBT.

A recent investigation by Cand\`{e}s, Romberg and Tao \citep{Candes2006} into compressed sensing (CS), indicates that it is possible to recover the original signal exactly, using a linear measurement model with incomplete data. This theoretical derivation is applicable to DBT reconstructions, which are computed given incomplete forward projections. Therefore, mathematically, we can solve the DBT reconstruction problem perfectly, with a limited angle set of projections, given judicious choice of appropriate constraints such as regularisation.

Most recently, Van de Sompel et al. \citep{VandeSompel2011} have developed a task-driven evaluation study of FBP, SART and ML for DBT reconstructions. They have concluded that DBT reconstructions are highly dependent on the choice of particular acquisitions and reconstruction parameters. This is an expected but also a non-trivial observation. Although numerous iterative methods have been proposed for DBT application, FBP types of methods still dominate the industry. The reason for this is the ease of implementation and computation,  combined with the efficacious reconstructions of these FBP methods. Quantitative comparison of DBT reconstruction methods using clinical data is still an open topic for research however.

\subsection{DBT Registration}

Early breast cancer detection requires the recognition of subtle pathological changes, such as those due to tumour growth, over time. These abnormal changes and deformations of the breast tissue must be distinguished from normal deformations caused by differences in breast position, compression and other imaging acquisition parameters between time-points. In the high throughput breast screening context, the greater volume of data generated by DBT must be integrated into the workflow in a way that enhances performance but does not increase the workload of the clinicians involved \citep{CRUK2010}. In this respect, image registration could play an important role in eliminating differences between temporal DBT data sets due to patient position, allowing the observer to focus on identifying those changes that might be indicative of disease.

Previous work on DBT image registration is limited. Sinha et al. \citep{Sinha2009} described an application of a thin-plate spline registration of corresponding manually selected control points, using mutual information as the objective function. They applied this method to seven subjects' data sets, which were acquired between one year and a few minutes apart and estimated the registration accuracy to be 1.8mm $\pm1.4$. Zhang and Brady \citep{Zhang2010} proposed a method for feature point extraction and use the resulting landmarks to drive a polyaffine registration of a single pair of DBT data sets.

\subsection{Combined Reconstruction and Registration}

\begin{table*}
	\begin{center}
    \setlength{\floatsep}{10pt plus 3pt minus 2pt} % Set the length between the table and the text body
        \tabcaption{\small\emph{Comparison of different applications of simultaneous inverse problem. (SR: super-resolution; LR: low resolution; fwdProjs: forward projections; Recon.+Regn.: reconstruction and registration; ``\textendash'': not mentioned).}}
        \scalebox{0.598}{ % Resize the whole block
        \begin{tabular}{L{5.2cm}^C{2.5cm}^C{2.5cm}^C{2.5cm}^C{4.6cm}^C{4.8cm}}
            \toprule\rowstyle{\bfseries}
            \footnotesize
            Publications 		 & Application     & Dimension        & Optimisation 	   & Optimiser 	    & Data       \\
            \midrule\rowstyle{\mdseries}
            Chung et al. 2006   				   & SR  			  & 2D Affine 		   & Decoupled 	    & Gauss-Newton   			 & $32$ LR images                                  \\
            \midrule\rowstyle{\mdseries}
            He et al. 2007   					   & SR  			  & 2D Rigid 		   & Decoupled 	    & Conjugate Gradient   	& $5$ LR images                       \\
            \midrule\rowstyle{\mdseries}
            Yap et al. 2009  					   & SR  			  & 2D Rigid 		   & Decoupled 	    & Linear Interior Point 	& $5$ LR images                           \\
            \midrule\rowstyle{\mdseries}
            Jacobson and Fessler 2003 	           & PET  			  & 3D Affine 		   & Decoupled 	    & Gradient Descent   				 & $64$ fwdProjs $180^\mathrm{o}$                           \\
            \midrule\rowstyle{\mdseries}
            Fessler  2010	    		           & PET 			  & 3D \textendash	   & Decoupled 	    & Conjugate Gradient   	& \textendash                         \\
            \midrule\rowstyle{\mdseries}
            Odille et al. 2008   			       & MRI  			  & 3D Affine 		   & Decoupled	    & GMRES   				 & \textendash                                     \\
            \midrule\rowstyle{\mdseries}
            Schumacher et al. 2009                 & SPECT 			  & 3D Rigid 		   & Decoupled 	    & Gauss-Newton   		    & $60$ to $64$ fwdProjs $360^\mathrm{o}$                   \\
            \midrule\rowstyle{\mdseries}
            Yang et al. 2005   					   & Cryo-EM 		  & 3D Rotation		   & Decoupled 	    & Quasi-Newton (L-BFGS)  	& $84$ fwdProjs                       \\
            \midrule\rowstyle{\mdseries}
            Chung et al. 2010  				       & Cryo-EM 		  & 3D Rigid		   & Decoupled 	    & Quasi-Newton (L-BFGS)  	& $799$ fwdProjs                      \\
            \midrule\rowstyle{\mdseries}
            Our Recon.+Regn. Model  			   & DBT 			  & 3D Affine \& B-spline		   & Decoupled				& Conjugate Gradient or L-BFGS  						 & $22$ fwdProjs $50^\mathrm{o}$ ($\pm25^\mathrm{o}$)                \\
            \bottomrule
        \end{tabular}
        }
        \label{table:LiteratureCompare}
	\end{center}
\end{table*}

There is little or no previous research in combining reconstruction and registration of DBT. In Table \ref{table:LiteratureCompare}, we summarise relevant publications for other image modalities and non-breast applications. There are three primary applications that have been explored to date. These are super-resolution, motion-correction for medical imaging modalities like PET, SPECT and MRI, and 3D density map reconstruction from 2D cryo-electron microscopy (Cryo-EM) images.

First, the process of combining a set of low resolution images into a single high-resolution image is often referred to as super-resolution (SR). The SR problem involves registration and restoration. Most of the previous research separated these two tasks for the SR problem. Chung et al. \citep{Chung2006} elucidated a simultaneous mathematical framework that enabled combination of the problem of estimating the displacements with restoring the high-resolution image. Other studies, {\it e.g.,} \citep{He2007,Yap2009}, also proposed algorithms to integrate 2D rigid image registration into the image SR problem.

Second, due to the long acquisition times in medical imaging modalities such as PET, SPECT and MRI, patient motion is inevitable and constitutes a serious problem for any reconstruction algorithm. Many algorithms use a gating system or even breath-holding to mitigate the motion effect. In \citep{Jacobson2003}, the authors reported a method to jointly estimate image and deformation parameters in motion-corrected PET imaging. Odille et al., \citep{Odille2008} presented a coupled system to perform a motion-compensated reconstruction, and subsequently optimised the motion model for MRI. Schumacher et al. \citep{Schumacher2009} used the combined reconstruction and motion correction method in SPECT imaging. Recently, Fessler \citep{Fessler2010} proposed the novel idea of using an optimisation transfer, {\it a.k.a.,} majorise-minimise method, to find a surrogate objective function to simplify the original simultaneous functional for motion-compensated PET reconstruction.

For the Cryo-EM imaging application, Yang et al. \citep{Yang2005} described a simultaneous method to refine a 3D density map and the orientation parameters of the 2D projections, which were used to reconstruct this map. Chung et al. \citep{Chung2010} extended this idea using parallel computing to speed up the application.

We summarise these previous research in Table \ref{table:LiteratureCompare} and compare them with our fully coupled inverse problem for the DBT reconstruction and registration.

%______________________________________________________________________________________________________________________
% Section 3: Conventional Method
\section{Conventional Method}
\label{sec:conventional_method}

\subsection {Forward Problem}

\begin{figure*}[!htb]
  %\begin{center}
  \centering % \setlength{\floatsep}{10pt plus 3pt minus 2pt}
        \subfigure[]{\label{fig:subfig:projFixed1}
    \includegraphics[width=0.15\textwidth]{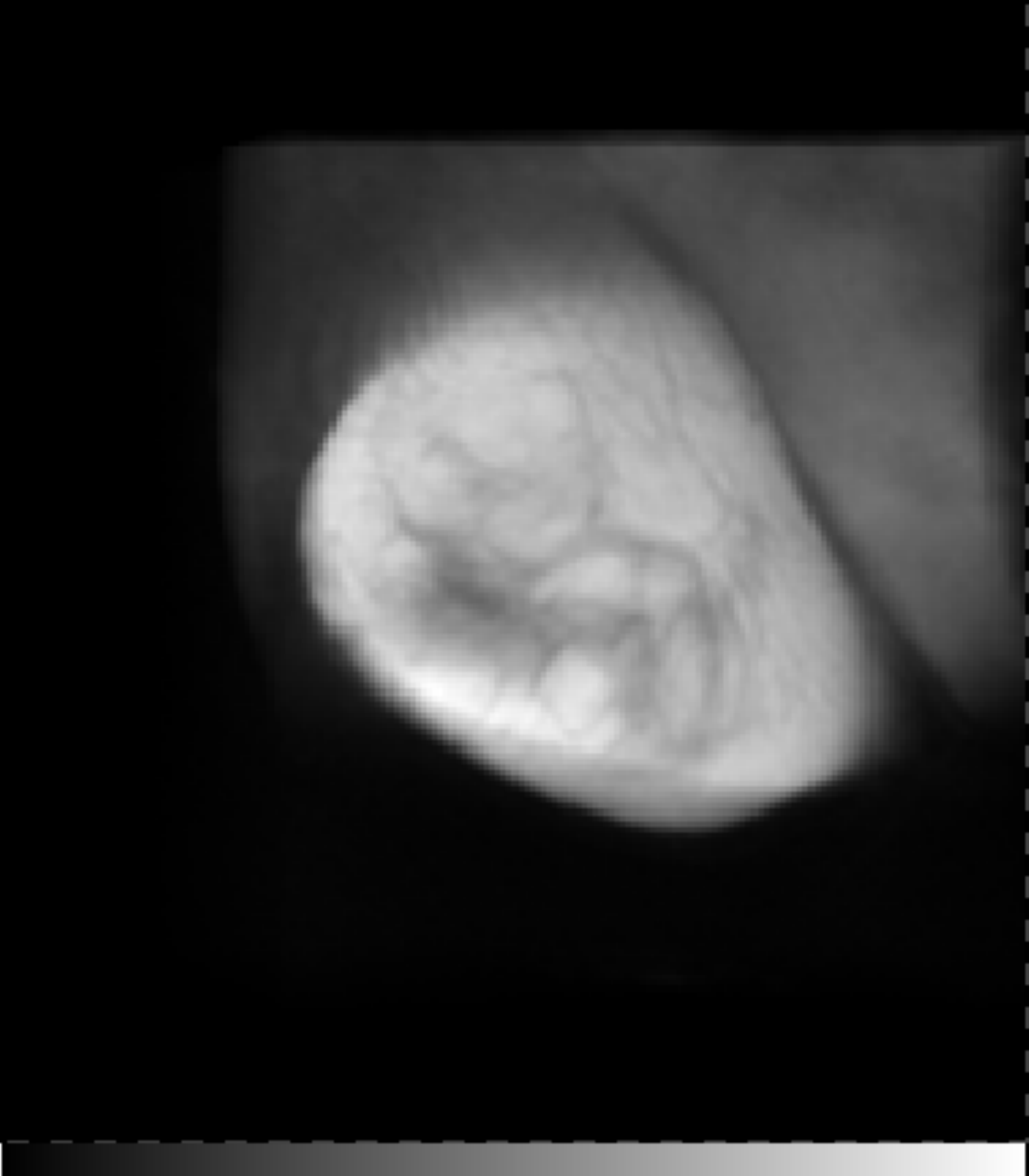}}
        \subfigure[]{
    \includegraphics[width=0.15\textwidth]{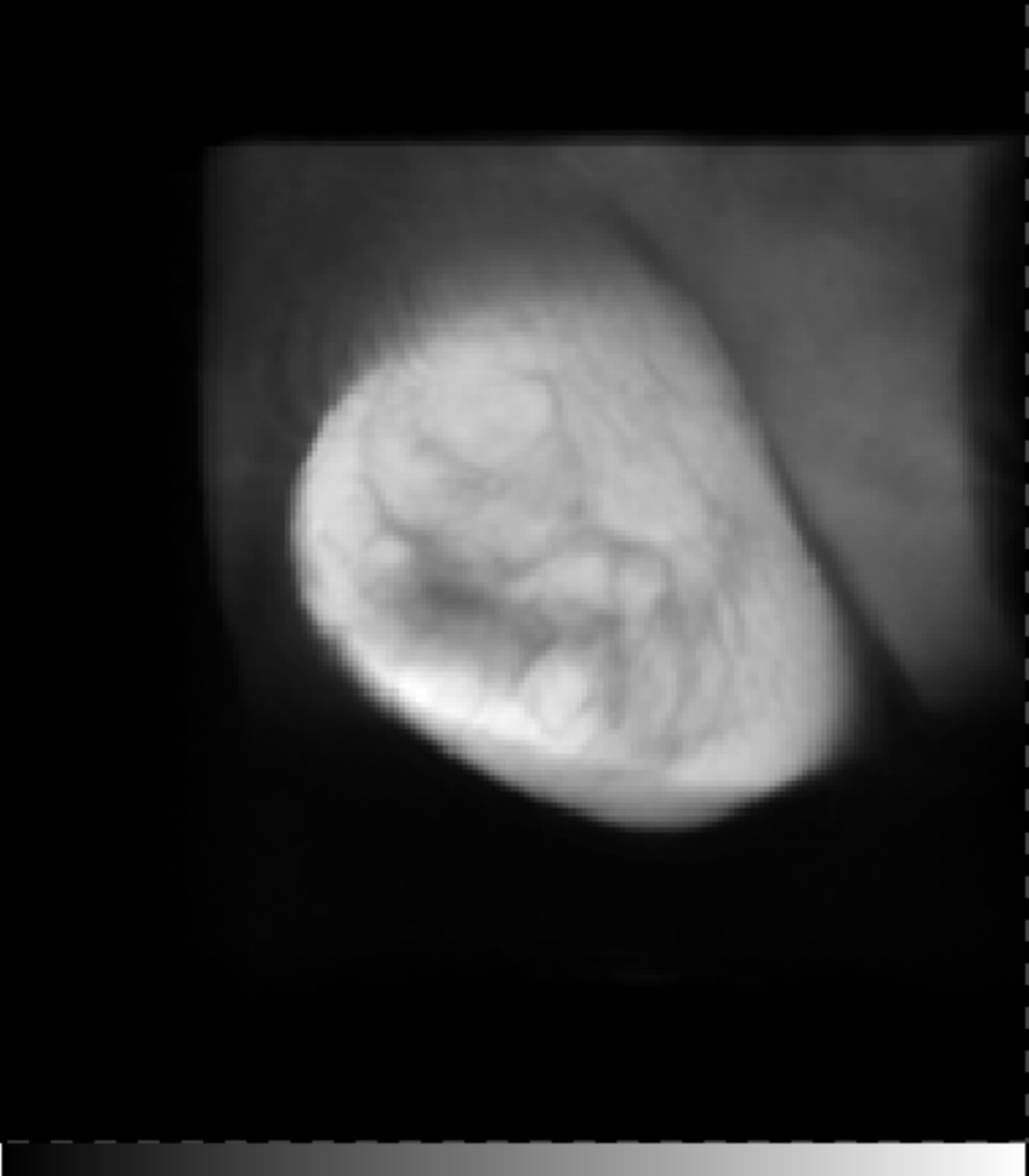}}
        \subfigure[]{
    \includegraphics[width=0.15\textwidth]{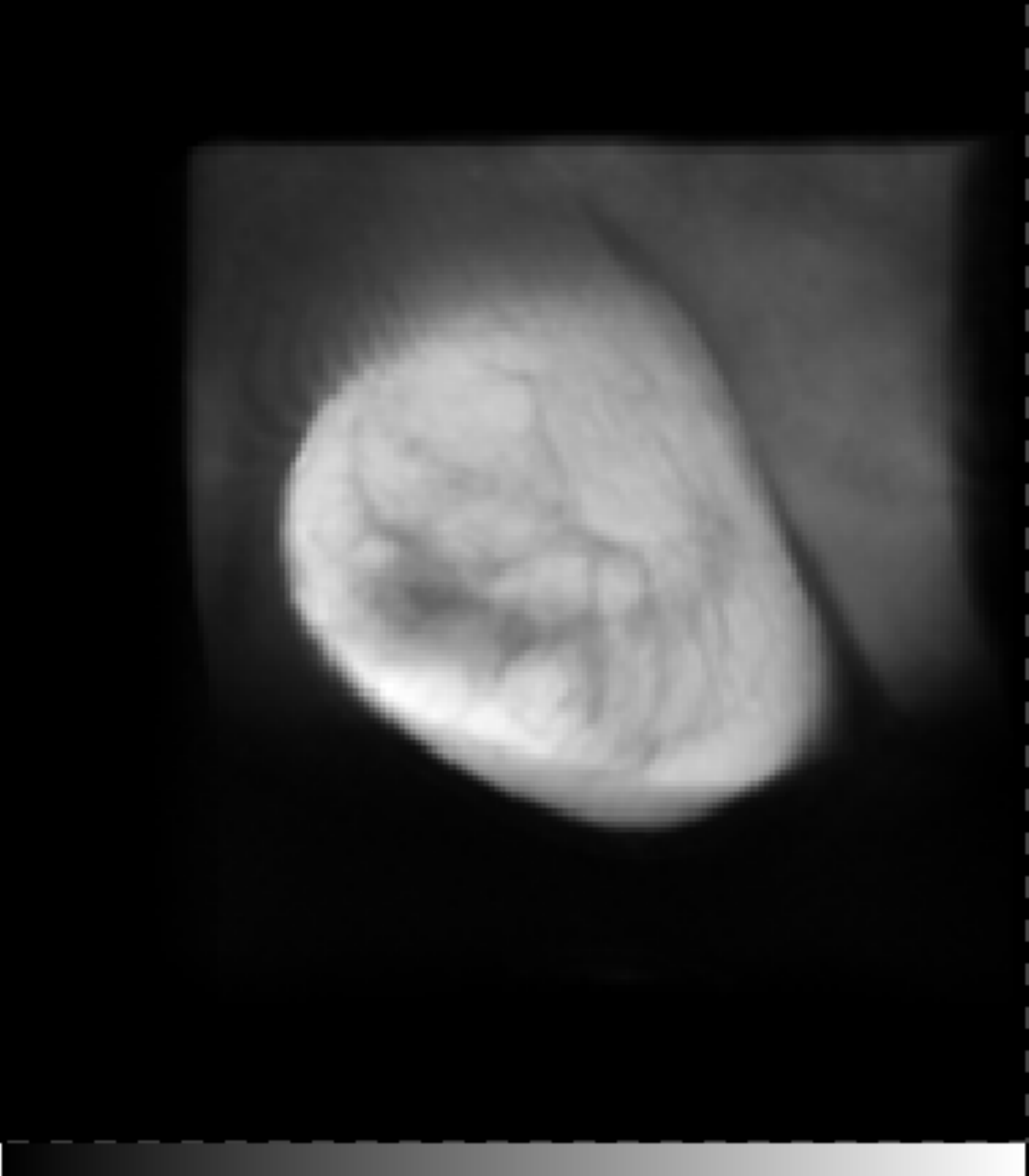}}
        \subfigure[]{
    \includegraphics[width=0.15\textwidth]{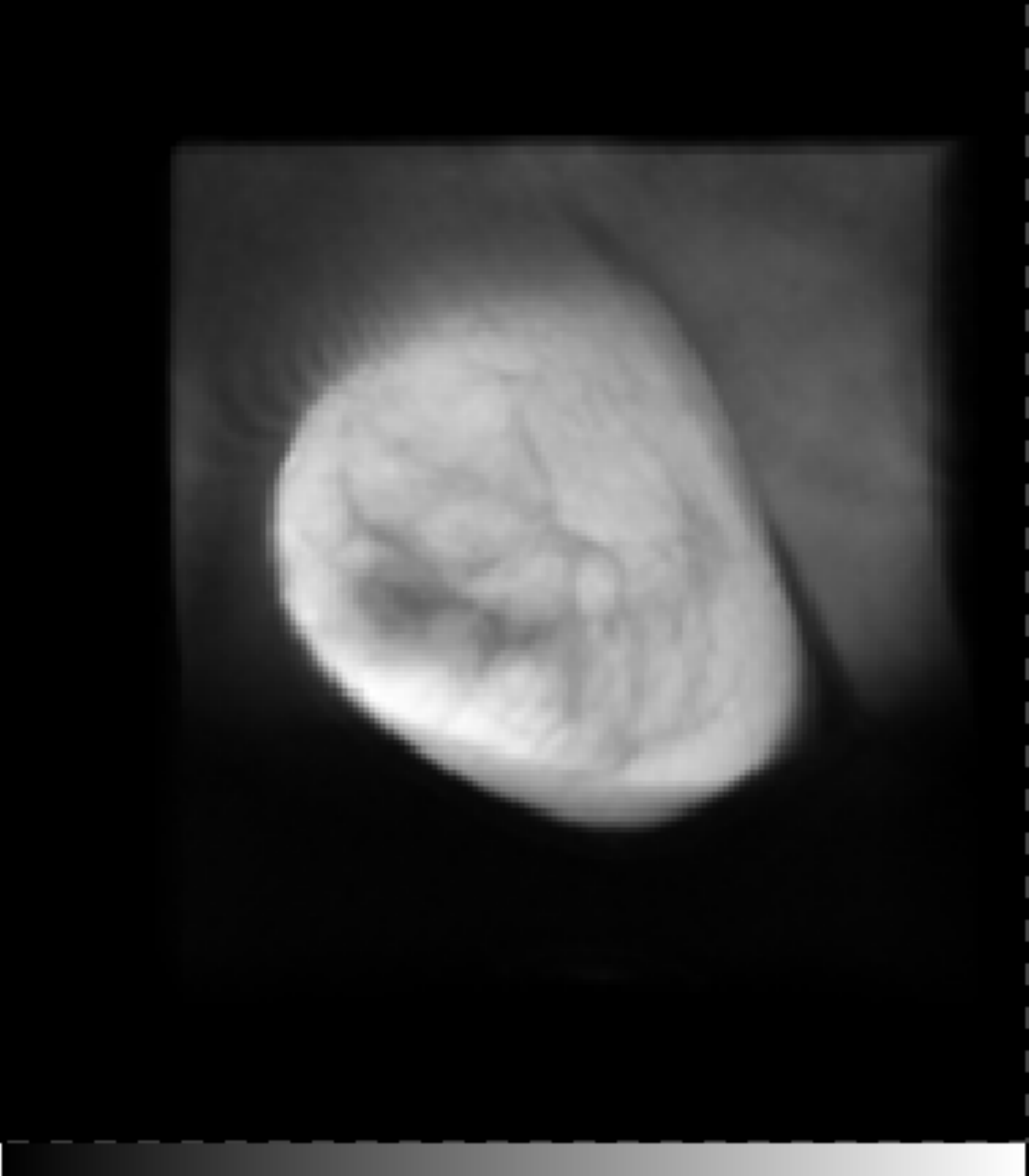}}
        \subfigure[]{
    \includegraphics[width=0.15\textwidth]{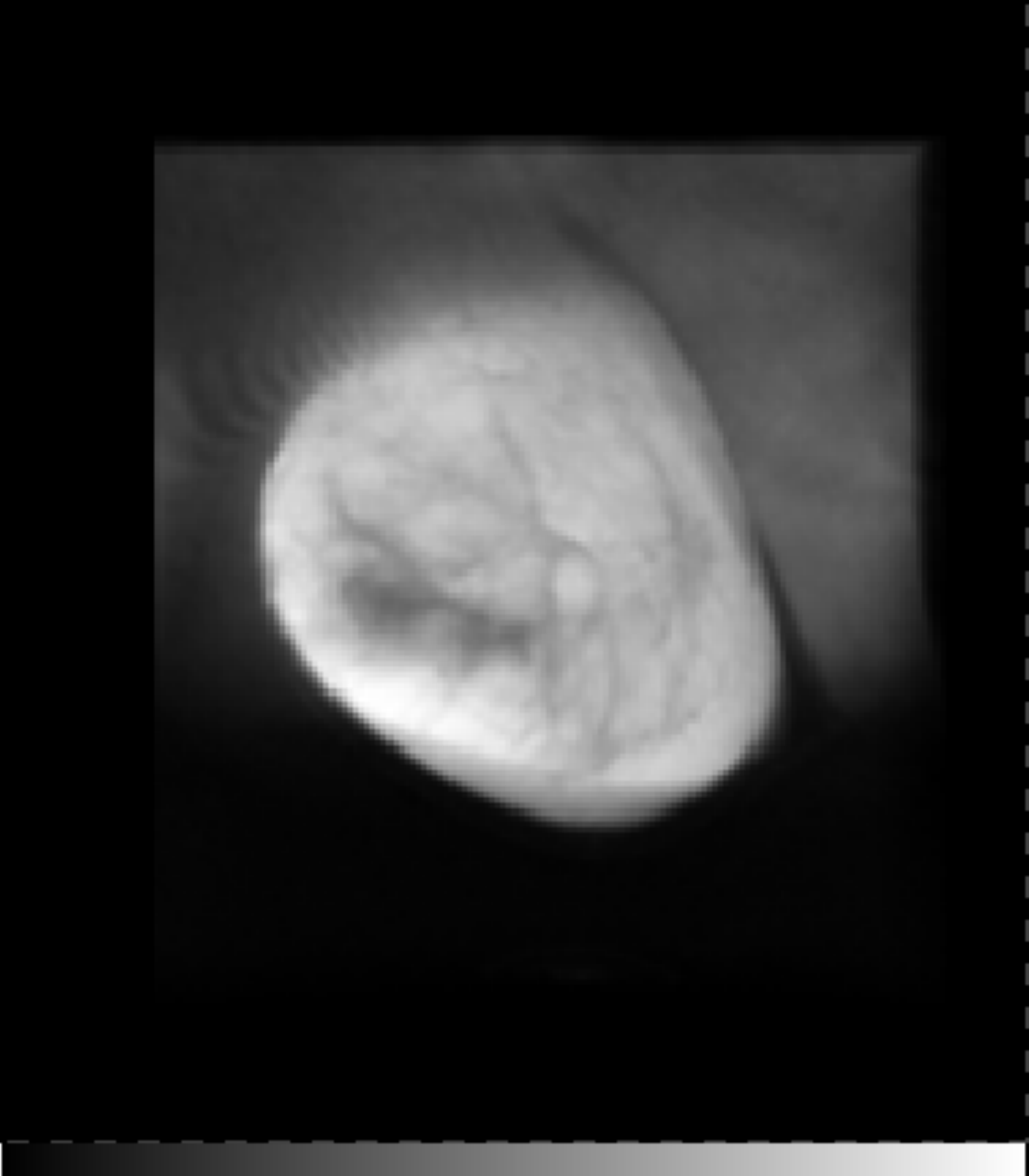}}
        \subfigure[]{
    \includegraphics[width=0.15\textwidth]{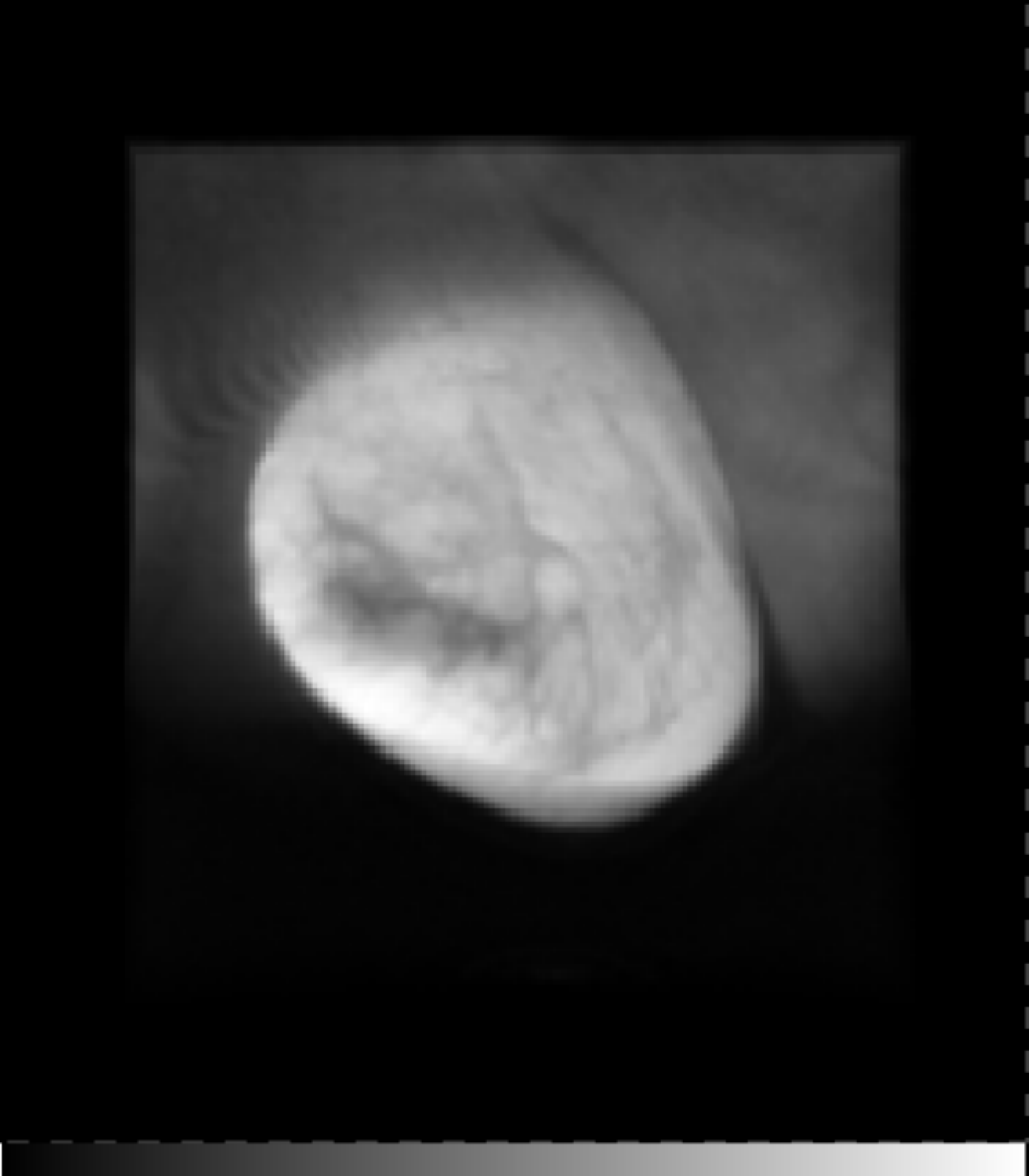}}
        \subfigure[]{
    \includegraphics[width=0.15\textwidth]{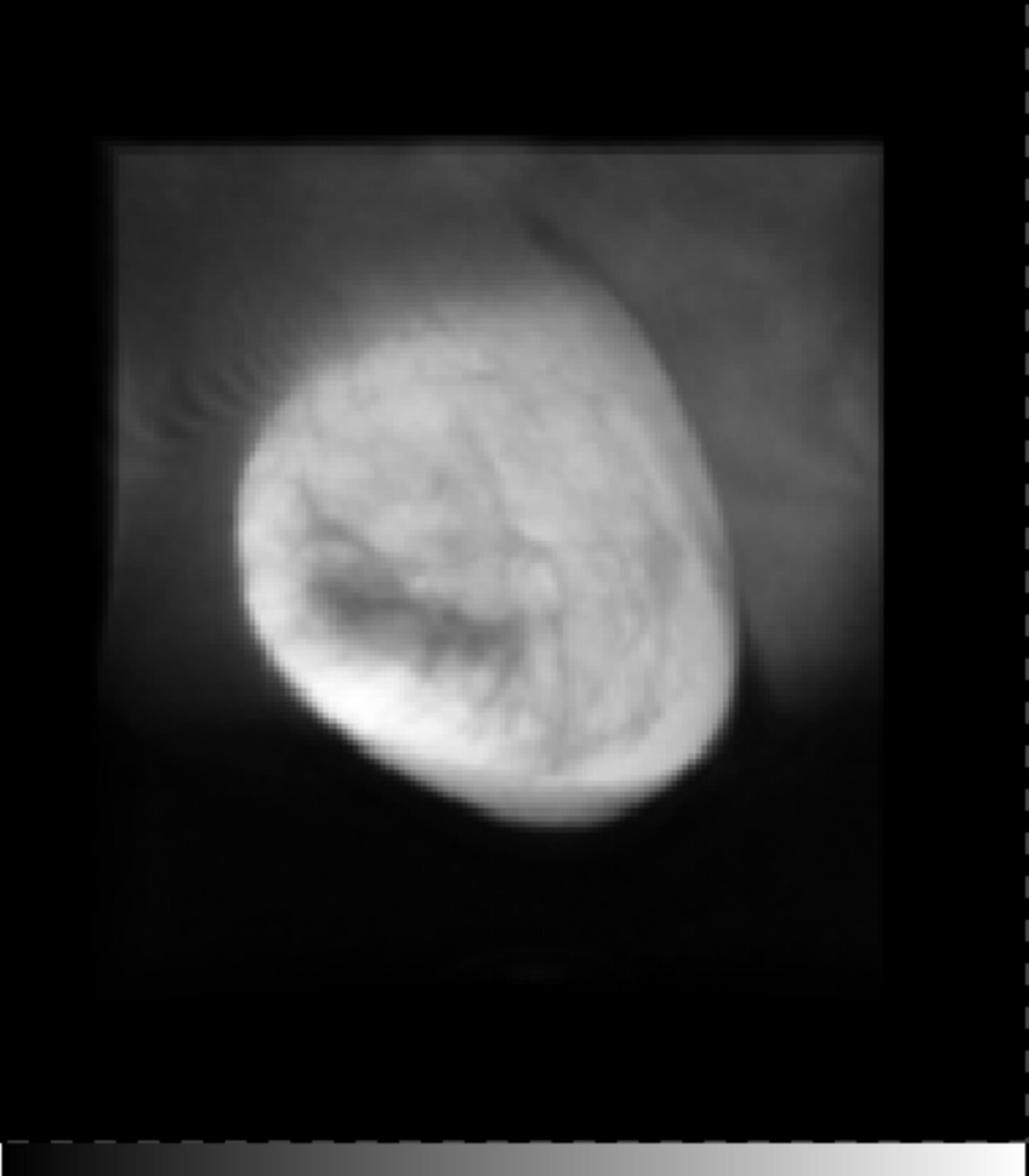}}
        \subfigure[]{
    \includegraphics[width=0.15\textwidth]{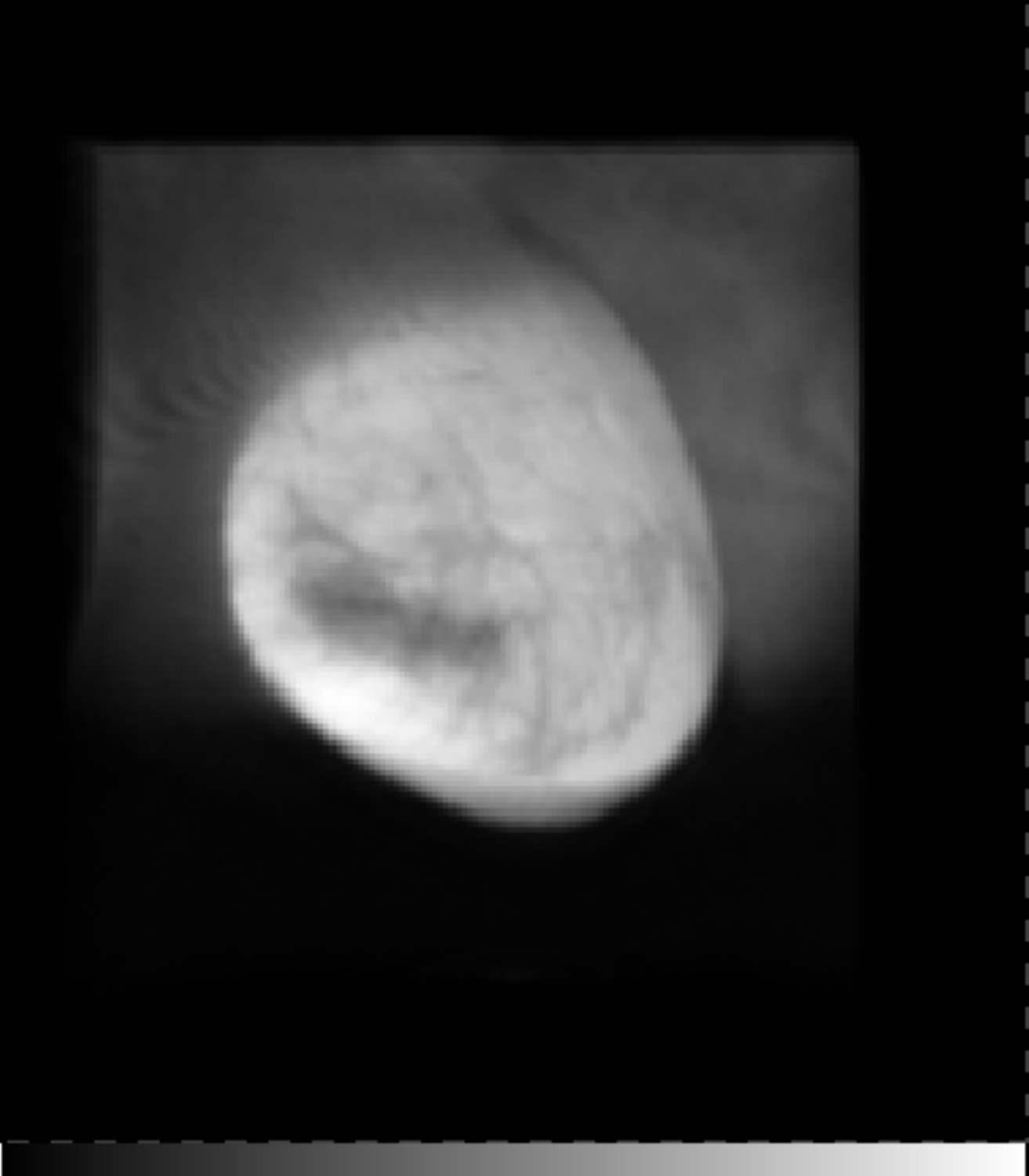}}
        \subfigure[]{
    \includegraphics[width=0.15\textwidth]{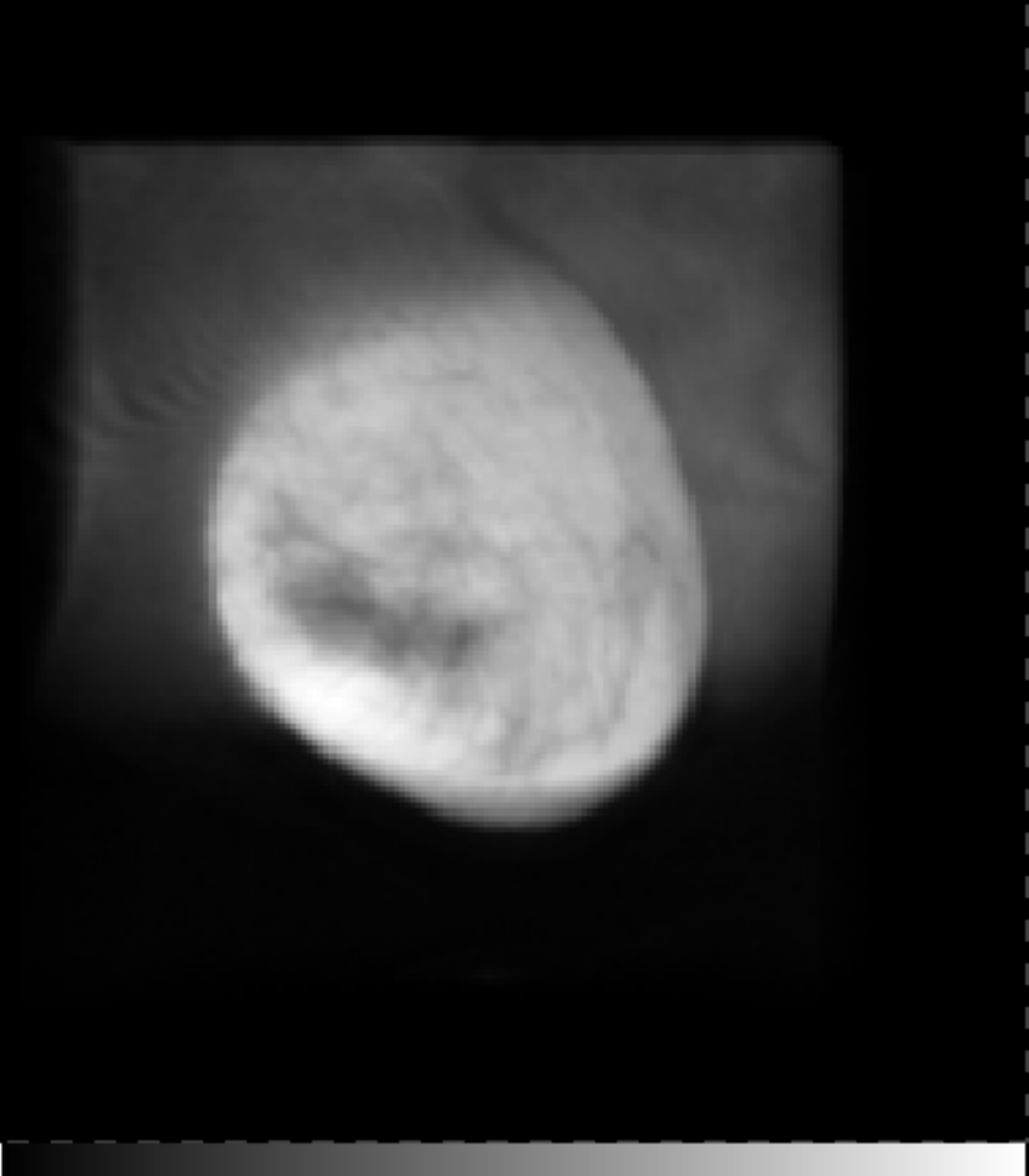}}
        \subfigure[]{
    \includegraphics[width=0.15\textwidth]{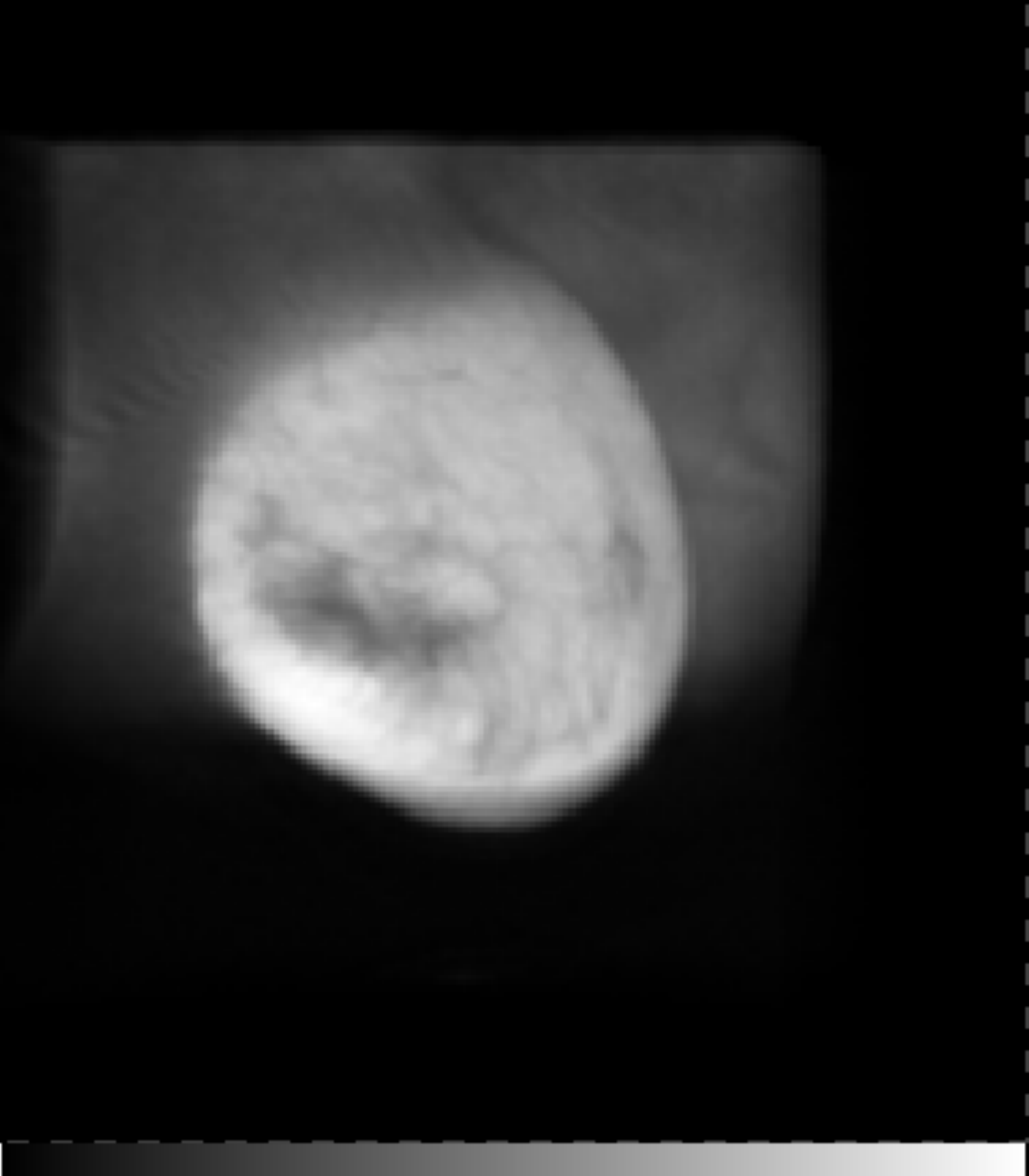}}
        \subfigure[]{\label{fig:subfig:projFixed11}
    \includegraphics[width=0.15\textwidth]{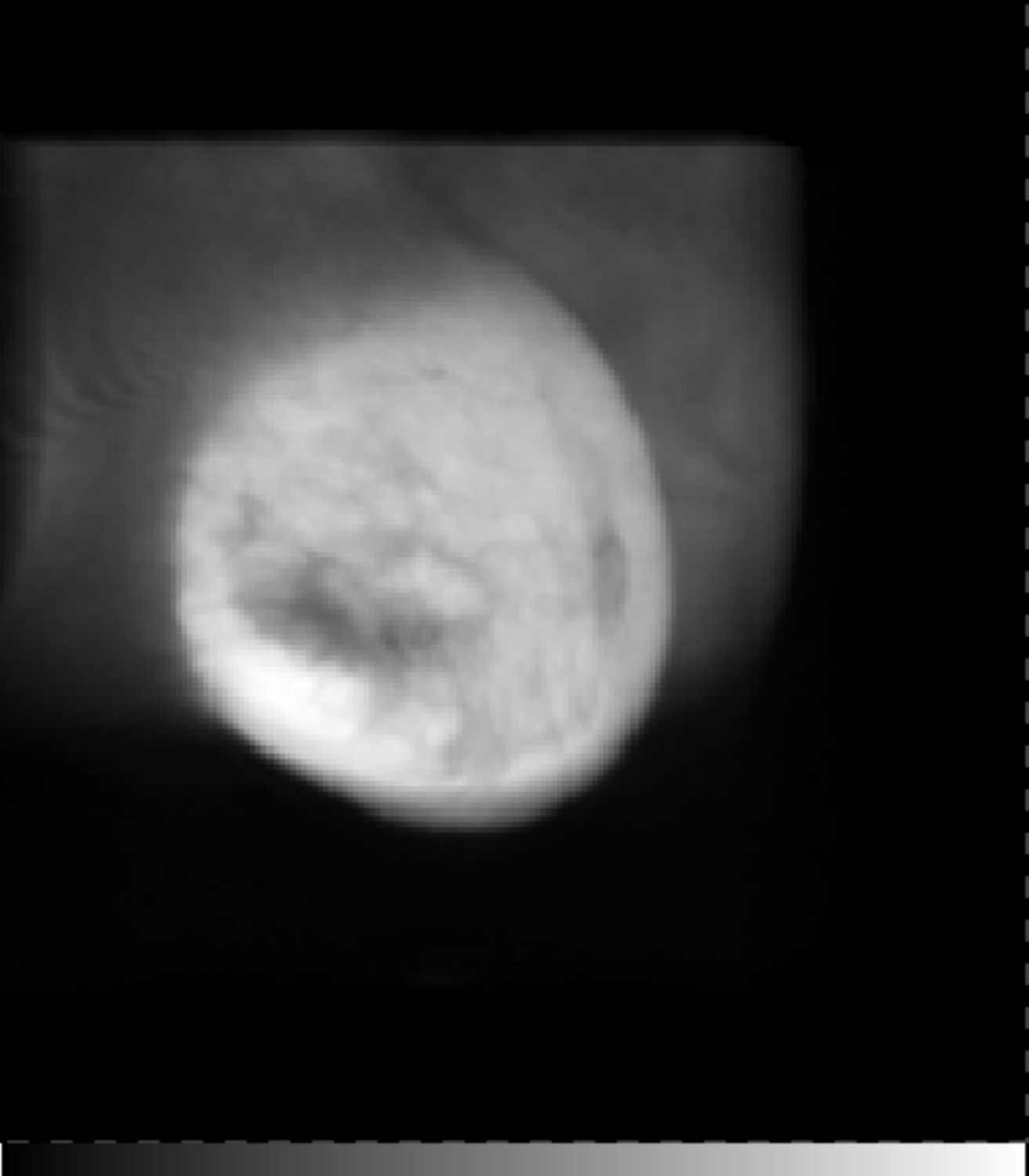}}  \\
    	  \subfigure[]{\label{fig:subfig:projMoving1}
    \includegraphics[width=0.15\textwidth]{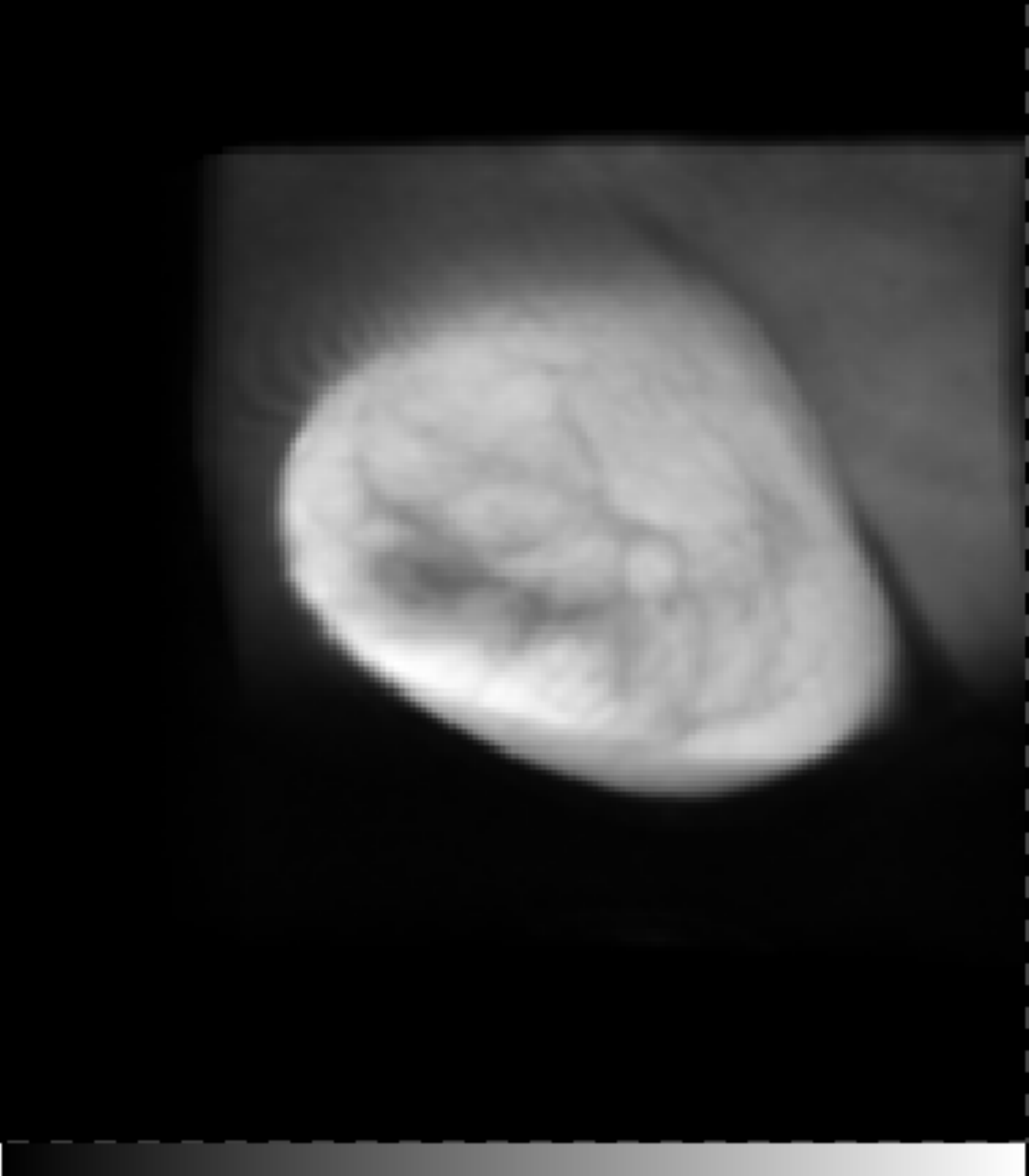}}
        \subfigure[]{
    \includegraphics[width=0.15\textwidth]{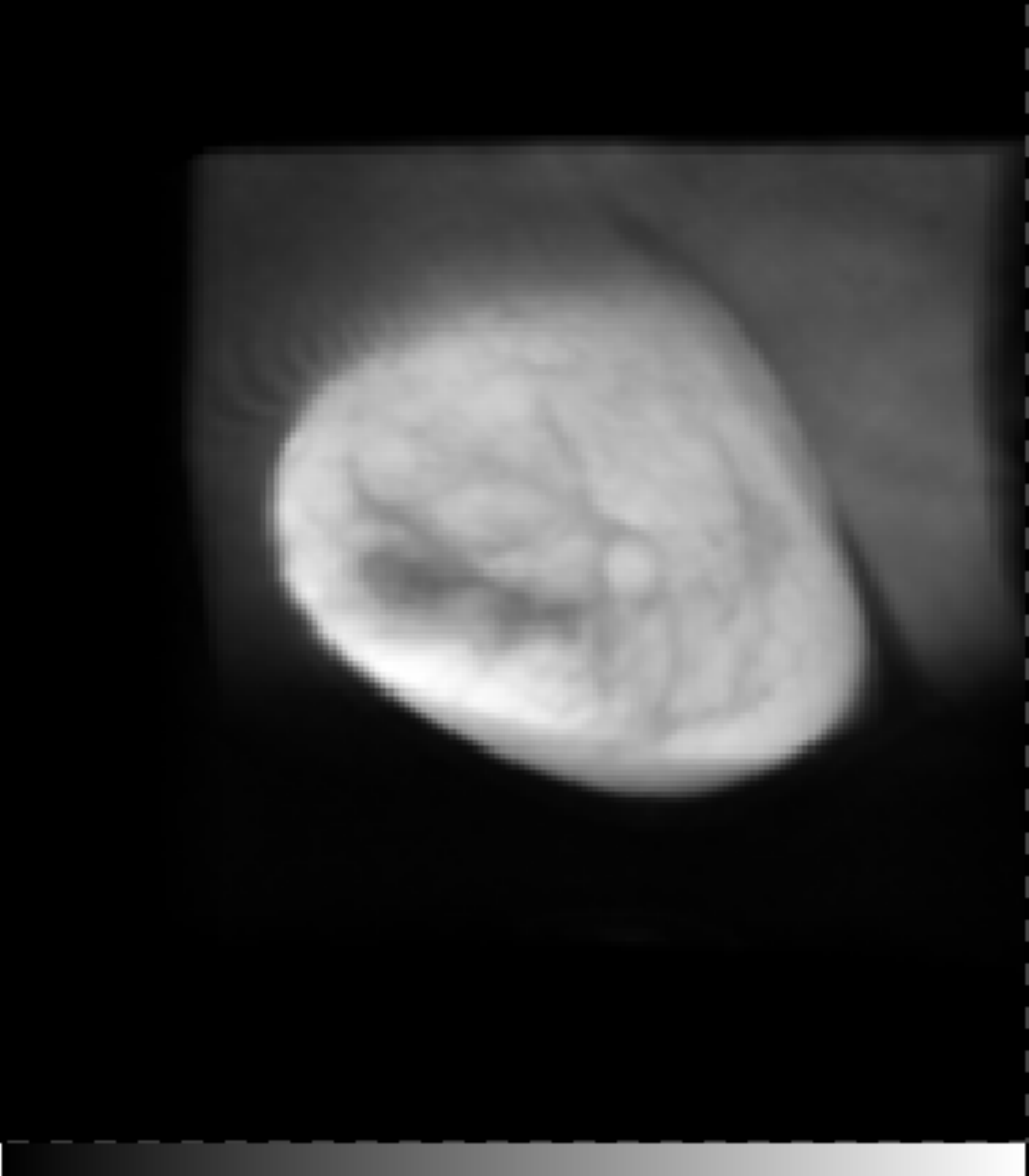}}
        \subfigure[]{
    \includegraphics[width=0.15\textwidth]{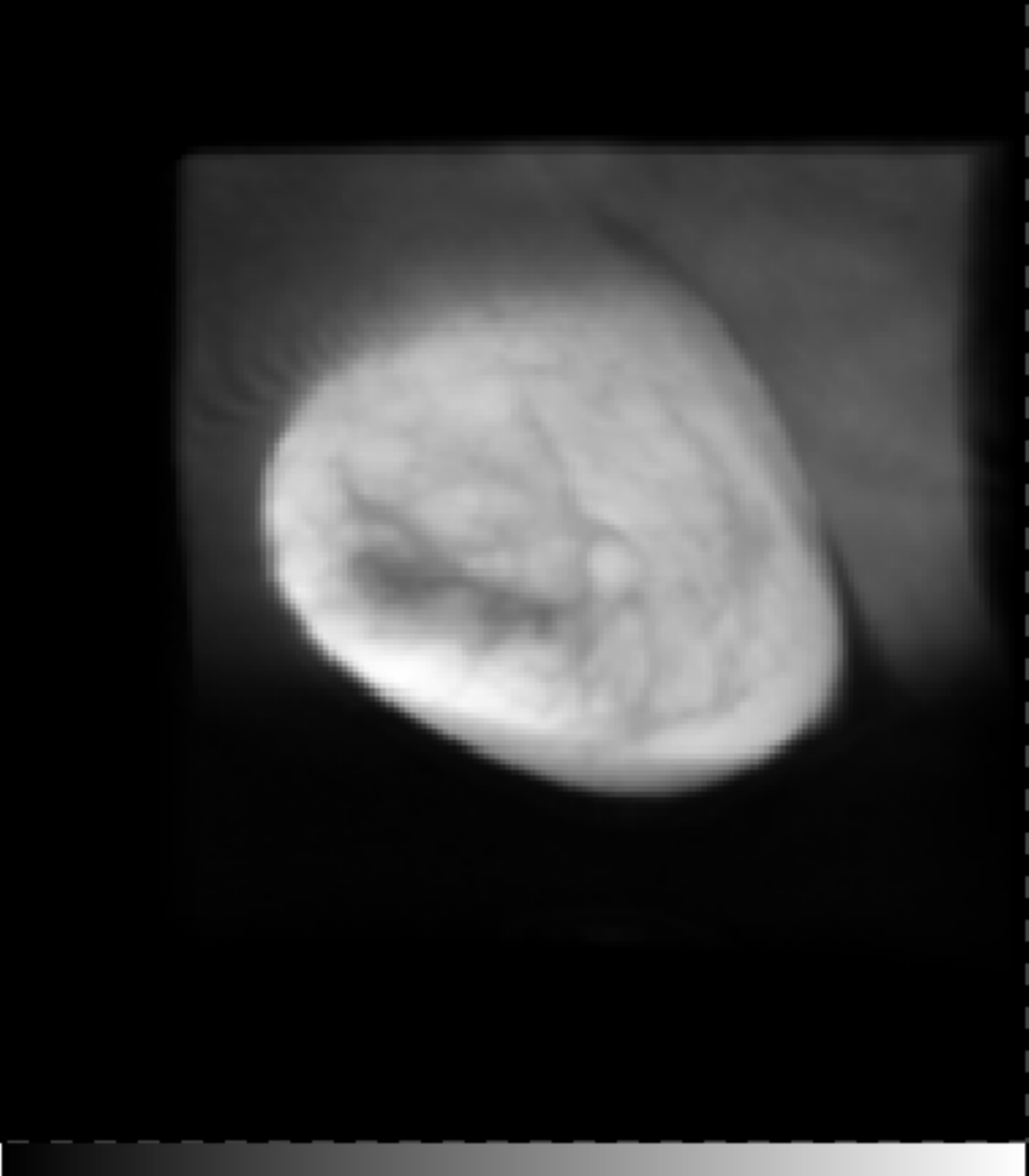}}
        \subfigure[]{
    \includegraphics[width=0.15\textwidth]{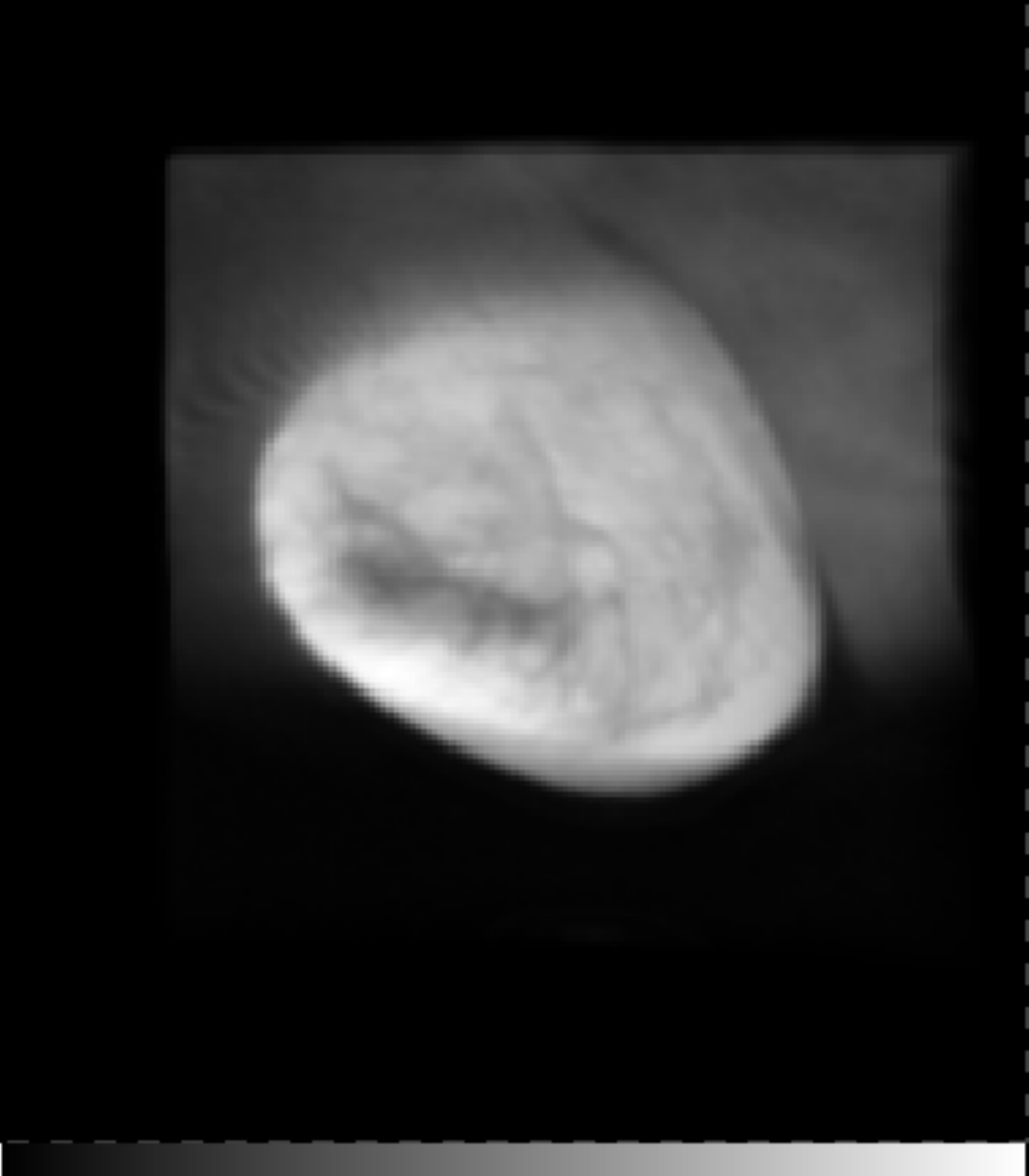}}
        \subfigure[]{
    \includegraphics[width=0.15\textwidth]{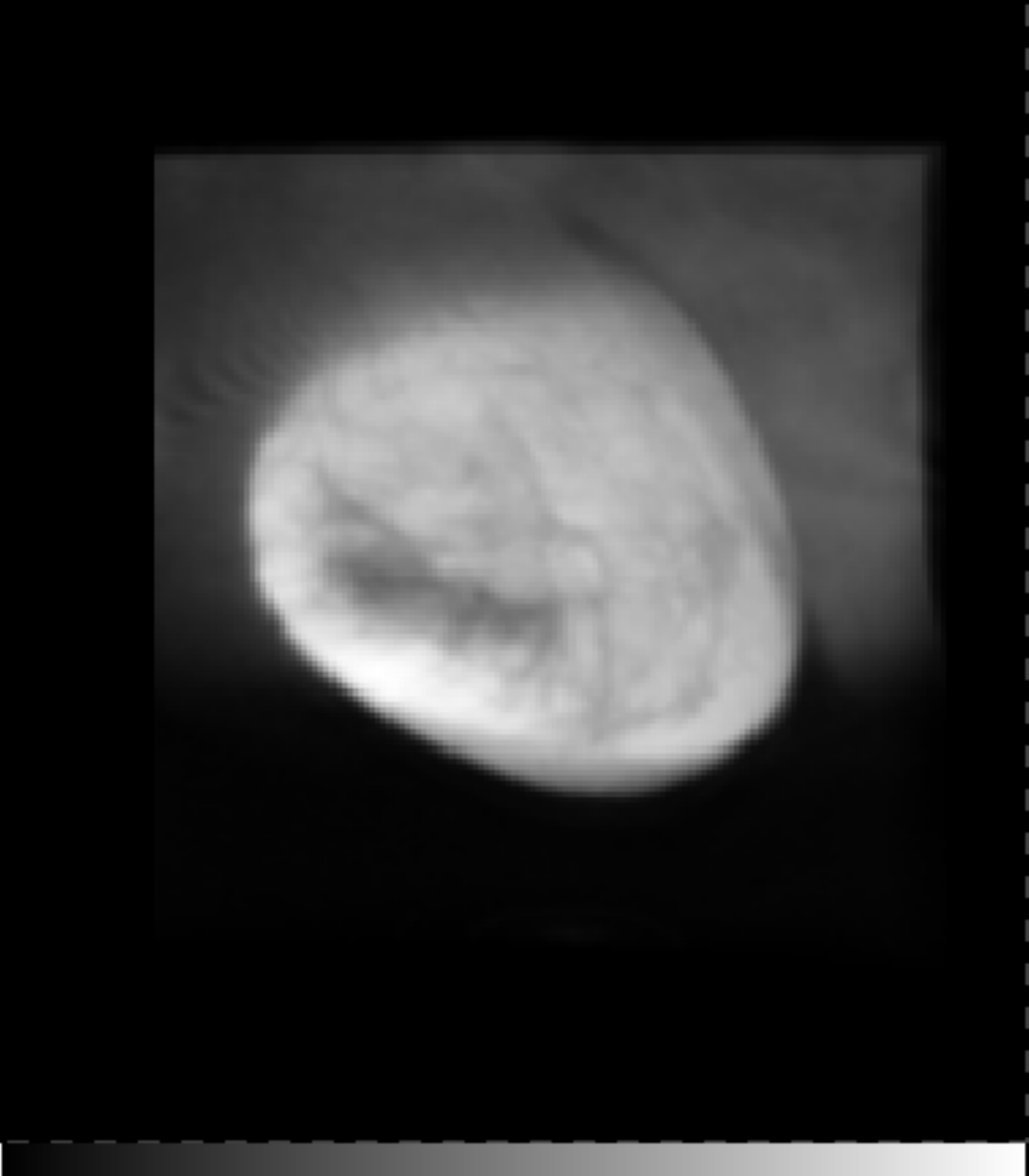}}
        \subfigure[]{
    \includegraphics[width=0.15\textwidth]{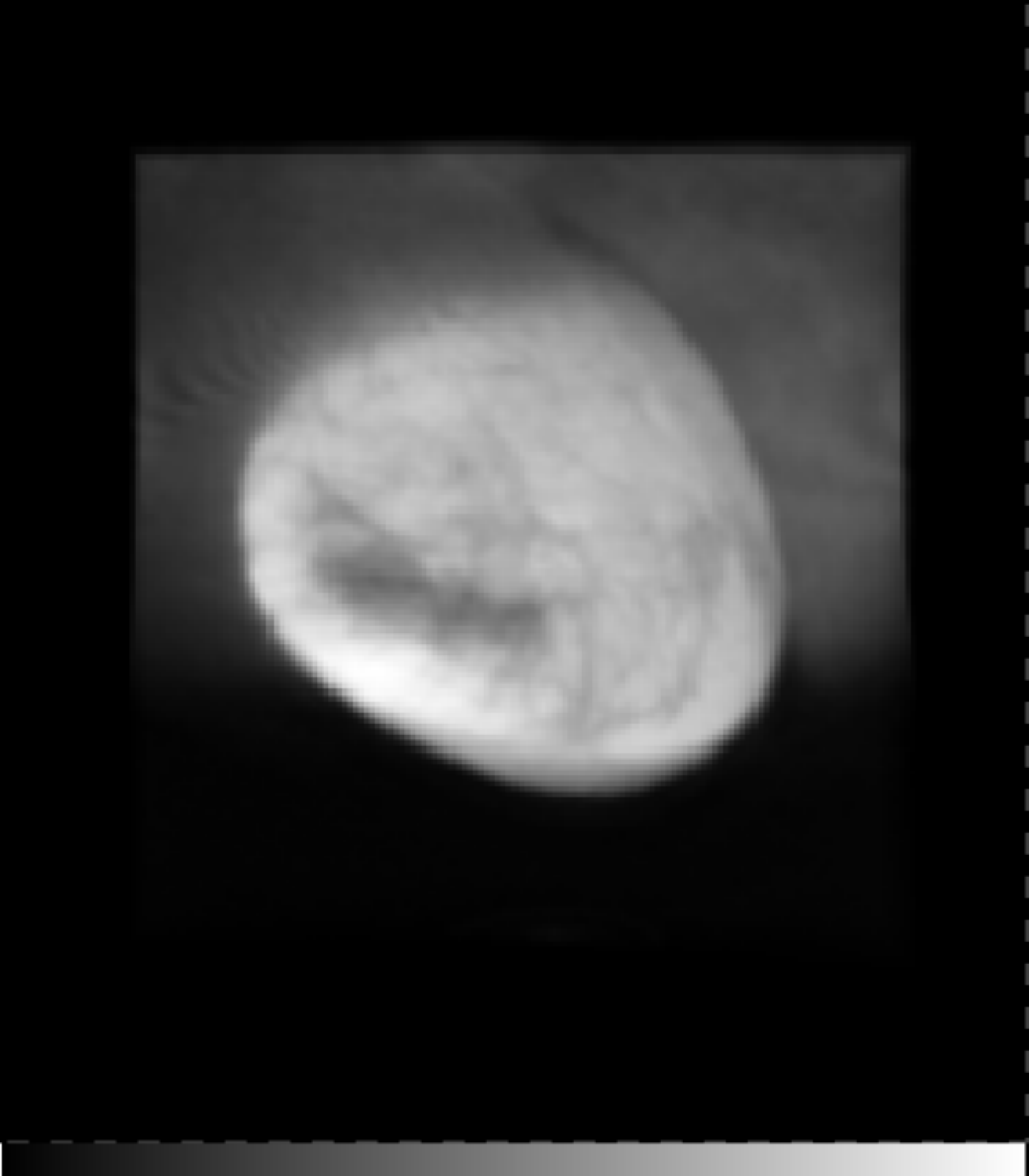}}
        \subfigure[]{
    \includegraphics[width=0.15\textwidth]{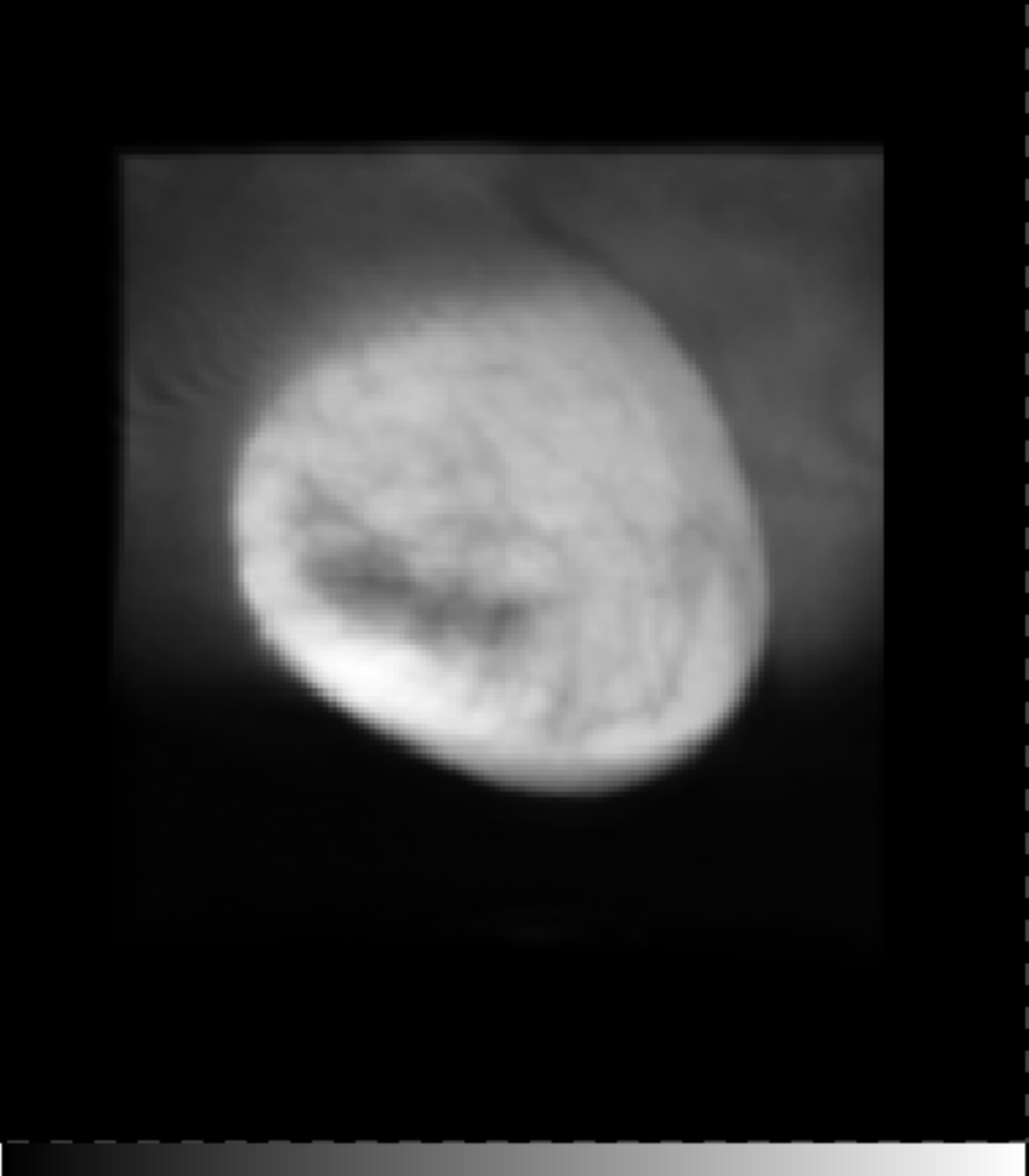}}
        \subfigure[]{
    \includegraphics[width=0.15\textwidth]{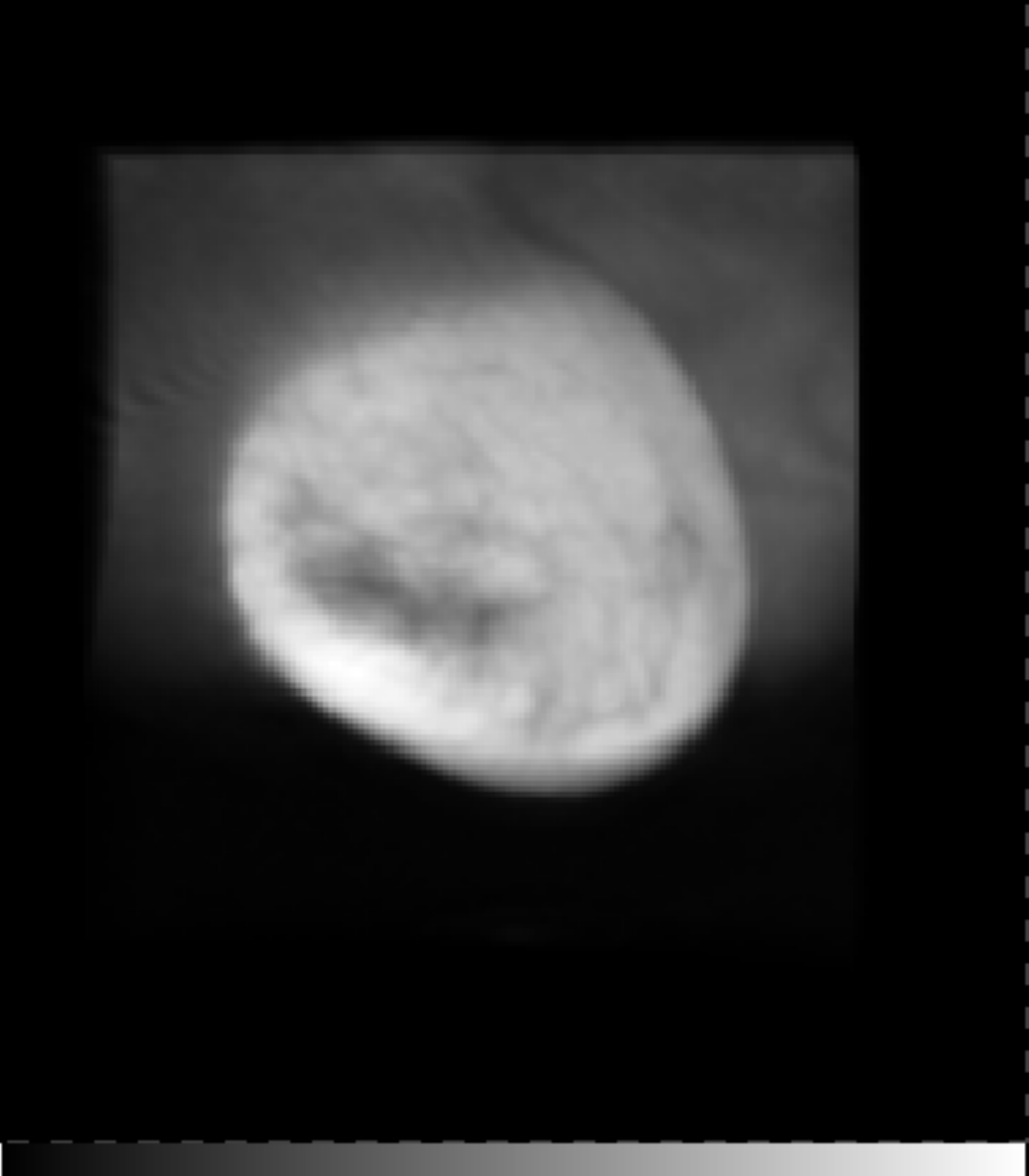}}
        \subfigure[]{
    \includegraphics[width=0.15\textwidth]{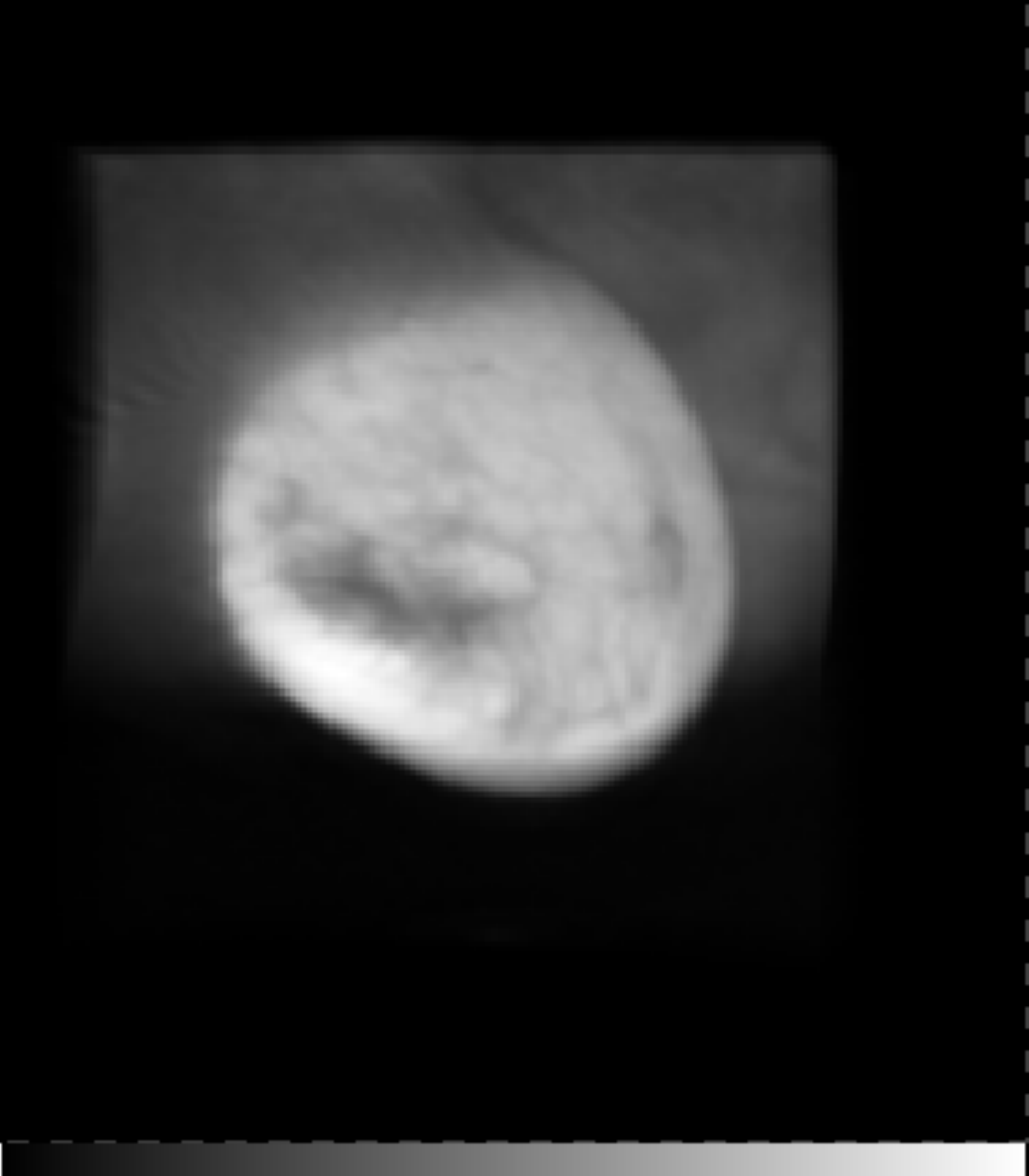}}
        \subfigure[]{
    \includegraphics[width=0.15\textwidth]{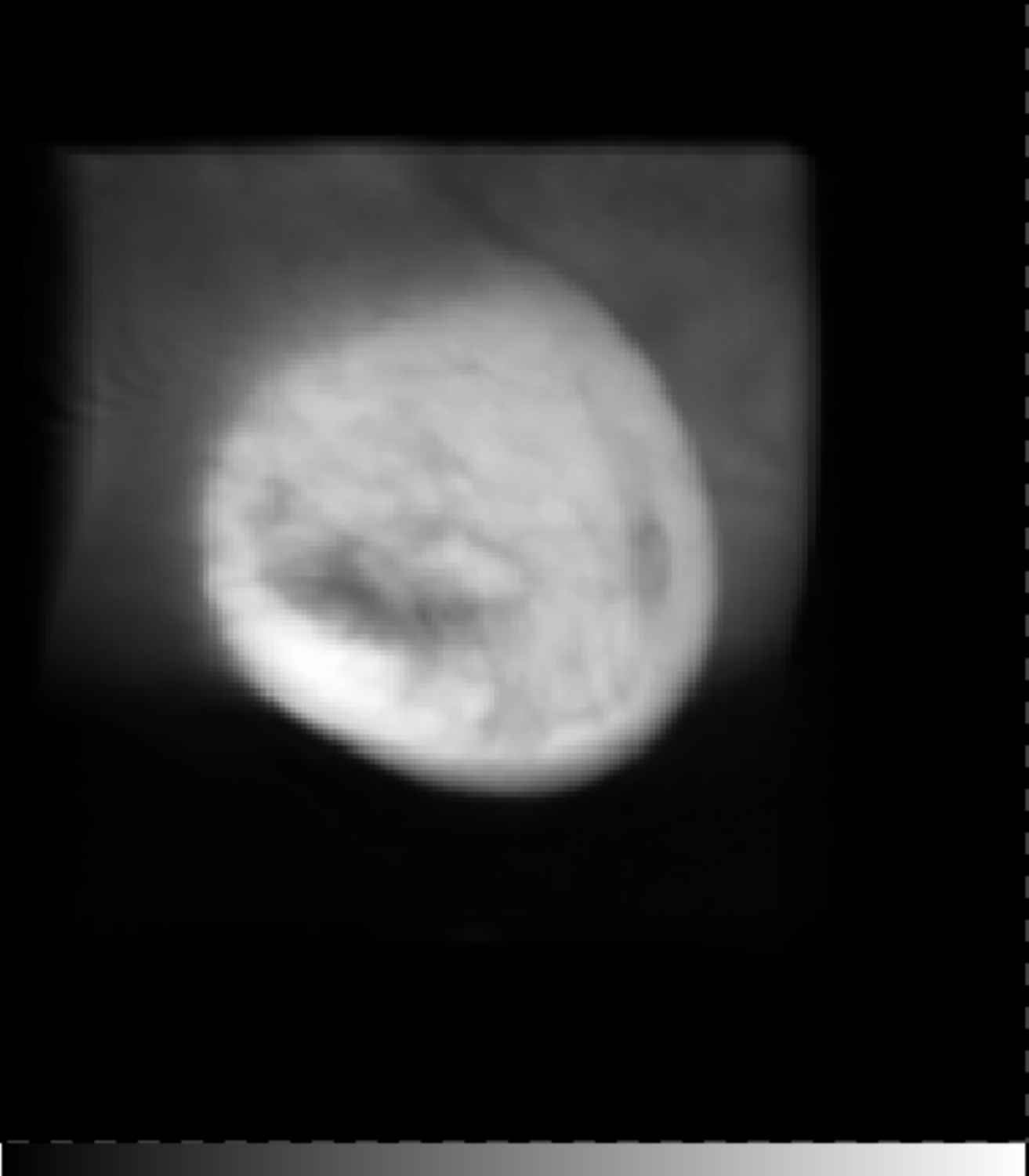}}
        \subfigure[]{\label{fig:subfig:projMoving11}
    \includegraphics[width=0.15\textwidth]{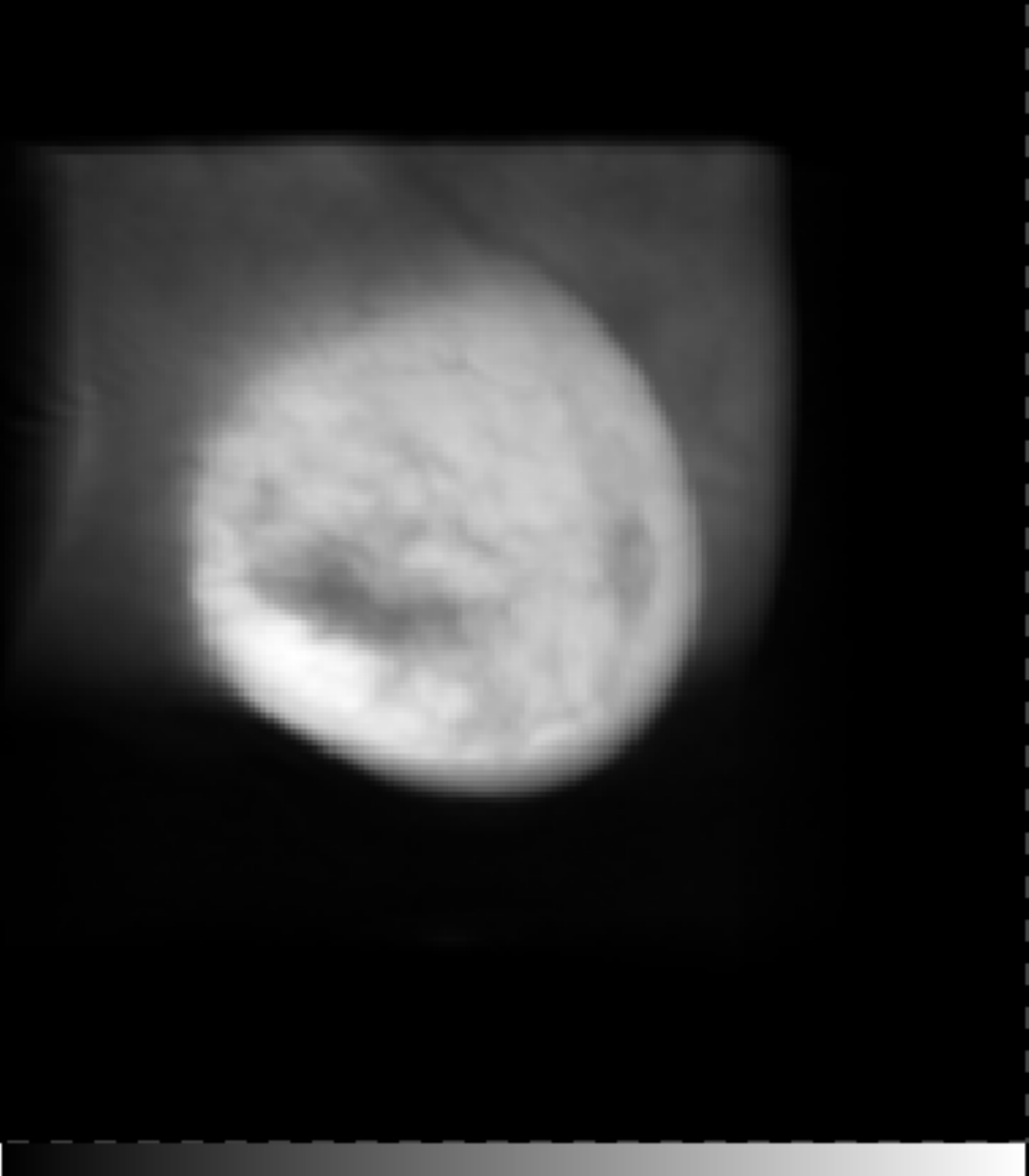}}
  % \figcaption{\small\bf\it The $11$ forward projections covering $\pm25^{\mathrm{o}}$ for both the fixed and moving images, {\it i.e.,} $\mathrm{p}_1$ (Figure\ref{fig:subfig:projFixed1}-\ref{fig:subfig:projFixed11}) and $\mathrm{p}_2$ (Figure \ref{fig:subfig:projMoving1}-\ref{fig:subfig:projMoving11}).}
	\figcaption{\small\bf\it The $11$ forward projections covering $\pm25^{\mathrm{o}}$ for both the fixed and moving breast MRI images, {\it i.e.,} $\mathrm{p}_1$ (a)-(k) and $\mathrm{p}_2$ (l)-(v).}
  \label{fig:P1FwdProj}
  %\end{center}
  \end{figure*}

A 3D image, ${\mathrm{f}}^{\mathrm{g}} \in\mathbb{R}^{\mathrm{D}_3}$, two sets of temporal data, $\mathrm{p}_1,~\mathrm{p}_2\in\mathbb{R}^{p_{\mathrm{num}} \times \mathrm{D}_2}$, the parametric transformations, $\mathcal{T}_{{\upzeta}}^{\mathrm{g}}$, and the system matrix, $A\in\mathbb{R}^{p_{\mathrm{num}} \times \mathrm{D}_2\times \mathrm{D}_3}:\mathbb{R}^{\mathrm{D}_3}\mapsto\mathbb{R}^{\mathrm{D}_2}$, can be related via
\begin{align}
  \mathrm{p}_1 &= A{\mathrm{f}}^{\mathrm{g}} = A\mathrm{R}(\mathrm{x}); \label{Reconstruction_System1} \\
  \mathrm{p}_2 &= A\mathcal{T}_{{\upzeta}}^{\mathrm{g}} {\mathrm{f}}^{\mathrm{g}} = A\mathrm{T}[\mathcal{T}_{{\upzeta}}(\mathrm{x})], \label{Reconstruction_System2}
\end{align}
where $\mathrm{D}_2$ and $\mathrm{D}_3$ denote the dimensions of 2D projection space and 3D volume space, respectively. In addition, ${\mathrm{f}}^{\mathrm{g}}$ and $\mathcal{T}_{{\upzeta}}^{\mathrm{g}}$ are the ground truth of the reconstruction and the parametric transformations respectively, whilst $\mathrm{R}$ and $\mathrm{T}$ represent the interpolations at original coordinates $x$ and transformed coordinates $\mathcal{T}_{{\upzeta}}(\mathrm{x})$. Forward projections, {\it i.e.,} $\mathrm{p}_1,~\mathrm{p}_2$, are acquired using a limited angle DBT geometry with $p_{\mathrm{num}}=11$ projections covering $\pm25^\circ$ (Figure \ref{fig:P1FwdProj}).

\subsection {Conventional Sequential Method}

In the conventional sequential method, the reconstruction of Equations \ref{Reconstruction_System1} and \ref{Reconstruction_System2} can be solved by minimising
\begin{align}
  \mathrm{f}_1^\star = \arg\min_{\mathrm{f}_1}~\Big(~f(\mathrm{f}_1) &= \frac{1}{2}\big\|A\mathrm{f}_1-\mathrm{p}_1\big\|^2~\Big) \label{Complete_Reconstruction_1}; \\
  \mathrm{f}_2^\star = \arg\min_{\mathrm{f}_2}~\Big(~f(\mathrm{f}_2) &= \frac{1}{2}\big\|A\mathrm{f}_2-\mathrm{p}_2\big\|^2~\Big), \label{Complete_Reconstruction_2}
\end{align}
where $\mathrm{f}_1 = \mathrm{R}(\mathrm{x})$ and $\mathrm{f}_2 = \mathrm{T}[\mathcal{T}_{{\upzeta}}(\mathrm{x})]$.

Following reconstruction, volumes $\mathrm{f}_1^\star$ and $\mathrm{f}_2^\star$, {\it i.e.,} the fixed and moving images, are registered with respect to the registration parameters ${\upzeta}$:
\begin{align} \label{Complete_Registration}
{\upzeta}^\star & = \arg\min_{\upzeta}~\Big(~f({\upzeta}) = \frac{1}{2}\big\|\mathcal{T}_{{\upzeta}}(\mathrm{f}_2^\star)-\mathrm{f}_1^\star\big\|^2~\Big)  \\
&= \arg\min_{\upzeta}~\Big(~f({\upzeta}) = \frac{1}{2}\big\| \mathrm{T}^\star[\mathcal{T}_{{\upzeta}}(\mathrm{x})] - \mathrm{R}^\star(\mathrm{x}) \big\|^2~\Big), \notag
\end{align}
in which $\mathrm{T}^\star$ and $\mathrm{R}^\star$ denote the interpolations using reconstructed intensities, and the similarity measurement is described by a sum of squared difference.

%______________________________________________________________________________________________________________________
% Section 4: Iterative Method
\section{Iterative Method}
\label{sec:iterative_method}

In our novel iterative reconstruction and registration method \citep{Yang2010a, Yang2010b}, we solve Equations \ref{Reconstruction_System1} and \ref{Reconstruction_System2} with respect to estimates $\mathrm{f}_1$ and $\mathrm{f}_2$ of $\mathrm{f}$ and the registration parameters ${\upzeta}$, by alternating an incomplete optimisation ({\it i.e.,} $j$ iterations) of the reconstructed volumes $\hat{\mathrm{f}}_1$ and $\hat{\mathrm{f}}_2$:
  \begin{align}
    \hat{\mathrm{f}}_1 &= j \mathrm{~STEPS~of~} \arg\min_{\mathrm{f}_1}~\Big(~f({\mathrm{\mathrm{f}_1}}) ~\Big)  \label{Reconstruction_1} \\
    \hat{\mathrm{f}}_2 &= j \mathrm{~STEPS~of~} \arg\min_{\mathrm{f}_2}~\Big(~f({\mathrm{\mathrm{f}_2}}) ~\Big) \label{Reconstruction_2}
  \end{align}
with the registration of the current estimates $\hat{\mathrm{f}}_1$ and $\hat{\mathrm{f}}_2$ with respect to the registration parameters ${\upzeta}$:
  \begin{align}
    \hat{\upzeta} &= \arg\min_{\upzeta}~\Big(~f({\upzeta}) = \frac{1}{2}\big\|\mathcal{T}_{{\upzeta}}(\hat{\mathrm{f}}_2)-\hat{\mathrm{f}}_1 \big\|^2~\Big) \\
	&= \arg\min_{\upzeta}~\Big(~f({\upzeta}) = \frac{1}{2}\big\| \hat{\mathrm{T}}[\mathcal{T}_{{\upzeta}}(\mathrm{x})] - \hat{\mathrm{R}}(\mathrm{x}) \big\|^2~\Big).
	\label{Registration}
  \end{align}

  %\begin{center}
  %\centering
  %\scalebox{0.888}{
  \begin{algorithm}
  \label{alg:Iterative_Method}
  %\dontprintsemicolon
  \caption{Iterative Method}
  ~ \\[0.1cm]
  \KwIn{$\mathrm{p}_1,~\mathrm{p}_2.$}
  \KwOut{$\mathrm{f}_1^\star$,~$\mathrm{f}_2^\star$,~$\hat{\mathcal{T}}_{{\upzeta}} \hat{\mathrm{f}}_2.$}
  ~ \\
  \Begin{
  \% Initialise $\mathrm{f}_1$ and $\mathrm{f}_2$ to zero vectors; \rm \\
  \% Initialise $\upzeta$ to a vector of identity matrix when we use affine transformation model. \rm \\
  $\mathrm{f}_1 := 0;$ $\mathrm{f}_2 := 0;$ $\upzeta := I;$ \\
  ~ \\[-0.05cm]
      \% Outer loop for the registration runs $k$ times \rm \\
	  \For{$k~\mathrm{iterations}$} {
          \% Inner loop for the reconstruction runs $j$ times \rm \\
          \For{$j~\mathrm{iterations}$} {
                $\hat{\mathrm{f}}_1 = j \mathrm{~STEPS~of~} \arg\min_{\mathrm{f}_1}~\big(~f({\mathrm{\mathrm{f}_1}})~\big)$; \\
			    $\hat{\mathrm{f}}_2 = j \mathrm{~STEPS~of~} \arg\min_{\mathrm{f}_2}~\big(~f({\mathrm{\mathrm{f}_2}})~\big)$; \\
          }
		$\hat{\upzeta} = \arg\min_{\upzeta}~\big(~f({\upzeta}) = \frac{1}{2}\big\|\mathcal{T}_{{\upzeta}}\hat{\mathrm{f}}_2-\hat{\mathrm{f}}_1 \big\|^2~\big)$; \\
		$\mathrm{f}_1 = \hat{\mathcal{T}}_{{\upzeta}} \hat{\mathrm{f}}_2$;
        $\mathrm{f}_2 = \hat{\mathrm{f}}_2$; \\
		}		
      ~ \\[0.2cm]
      \% Output $\mathrm{f}_1^\star$, $\mathrm{f}_2^\star$, and $\hat{\mathcal{T}}_{{\upzeta}} \hat{\mathrm{f}}_2$  \rm \\
	  $\mathrm{f}_1^\star = \hat{\mathrm{f}}_1$; \\
	  $\mathrm{f}_2^\star = \hat{\mathrm{f}}_2$; \\
	  $\hat{\mathcal{T}}_{{\upzeta}} \hat{\mathrm{f}}_2$.
      }
      ~ \\[0.05cm]
  \end{algorithm}
  %}
  %\end{center}

After each registration iteration (Equation \ref{Registration}), and prior to the next iteration of the reconstructions (Equations \ref{Reconstruction_1} and \ref{Reconstruction_2}), the reconstruction estimates are updated as follows (Equations \ref{Update1} and \ref{Update2}).
  \begin{align}
    \mathrm{f}_1 &= \hat{\mathcal{T}}_{{\upzeta}} (\hat{\mathrm{f}}_2) = \hat{\mathrm{T}}[\hat{\mathcal{T}}_{{\upzeta}}(\mathrm{x})] \label{Update1} \\
    \mathrm{f}_2 &= \hat{\mathrm{f}}_2. \label{Update2}
    % \mathrm{f}_2 &= \hat{\mathrm{f}}_2 = \hat{\mathrm{T}}(\mathrm{x})
    \end{align}
  This ``outer loop'' of reconstruction followed by registration is repeated $k$ times. The last iteration outputs $\mathrm{f}_1^\star = \hat{\mathrm{f}}_1$, $\mathrm{f}_2^\star = \hat{\mathrm{f}}_2$ and $\hat{\mathcal{T}}_{{\upzeta}} \hat{\mathrm{f}}_2$.

In addition, the following analytical gradients are used to calculate $\hat{\mathrm{f}}_1$ and $\hat{\mathrm{f}}_2$ for the reconstruction
  \begin{align}
  g(\mathrm{f}_1) &= A^T(A\mathrm{f}_1-\mathrm{p}_1) \\
  g(\mathrm{f}_2) &= A^T(A\mathrm{f}_2-\mathrm{p}_2).
  \end{align}

Similarly, by the chain rule, the analytical gradient for the registration is
  \begin{align}
  g(\upzeta)
  &= \Big( \hat{\mathrm{f}}_1 - \mathcal{T}_{{\upzeta}} (\hat{\mathrm{f}}_2) \Big) \frac{ \partial \mathcal{T}_{{\upzeta}} (\hat{\mathrm{f}}_2) }{ \partial \upzeta } \\
  &= \Big( \hat{\mathrm{R}}(\mathrm{x}) - \hat{\mathrm{T}}[\mathcal{T}_{{\upzeta}}(\mathrm{x})] \Big) \frac{ \partial \hat{\mathrm{T}}[\mathcal{T}_{{\upzeta}}(\mathrm{x})] }{ \partial \upzeta } \\
  &= \Big( \hat{\mathrm{R}}(\mathrm{x}) - \hat{\mathrm{T}}[\mathcal{T}_{{\upzeta}}(\mathrm{x})] \Big) \frac{ \partial \hat{\mathrm{T}}[\mathcal{T}_{{\upzeta}}(\mathrm{x})] }{ \partial \mathcal{T}_{{\upzeta}}(\mathrm{x}) } \frac{ \partial \mathcal{T}_{{\upzeta}}(\mathrm{x}) }{ \partial \upzeta }. \label{Registration_Gradient_Factorise}
  \end{align}

It consists of three parts, {\it i.e.,} the image difference $\big( \hat{\mathrm{R}}(\mathrm{x}) - \hat{\mathrm{T}}[\mathcal{T}_{{\upzeta}}(\mathrm{x})] \big)$, the partial derivative of the moving image $\frac{ \partial \hat{\mathrm{T}}[\mathcal{T}_{{\upzeta}}(\mathrm{x})] }{ \partial \mathcal{T}_{{\upzeta}}(\mathrm{x}) }$ evaluated at location $\mathcal{T}_{{\upzeta}}(\mathrm{x})$, and the partial derivative of the transformation $\frac{ \partial \mathcal{T}_{{\upzeta}}(\mathrm{x}) }{ \partial \upzeta }$. As we employed the gradient information to get the updated parameters $\upzeta$, the moving image used to calculate the partial derivative is the \emph{original} moving image (not the updated or transformed moving image). In addition, this partial derivative, {\it i.e.,} spatial derivative, of the original moving image is calculated using the image gradient defined as
  \begin{equation}
  \nabla \hat{\mathrm{T}} (\mathrm{y}_{{\upzeta}}) = \frac{ \partial \hat{\mathrm{T}}[\mathcal{T}_{{\upzeta}}(\mathrm{x})] }{ \partial \mathcal{T}_{{\upzeta}}( \mathrm{x}) } = \Big( \frac{\partial \hat{\mathrm{T}}}{\partial x}\mathrm{y}_{{\upzeta}}, \frac{\partial \hat{\mathrm{T}}}{\partial y}\mathrm{y}_{{\upzeta}}, \frac{\partial \hat{\mathrm{T}}}{\partial z}\mathrm{y}_{{\upzeta}} \Big)^T,
  \end{equation}
in which $\mathrm{y}_{{\upzeta}} = \mathcal{T}_{{\upzeta}}(\mathrm{x}) = (x, y, z)^T$.

The preceding iterative reconstruction and registration method is summarised in Algorithm \ref{alg:Iterative_Method}, and in our implementation we can use a non-linear conjugate gradient or Limited Memory BFGS (L-BFGS) optimiser to solve the steps in Equations \ref{Reconstruction_1} and \ref{Reconstruction_2}.

At each update of the volumes after the outer loop registration, we use the transformation of $\mathrm{f}_2$ to correct $\mathrm{f}_1$ that is
  \begin{equation}
    \mathrm{f}_1 = \hat{\mathcal{T}}_{{\upzeta}} \hat{\mathrm{f}}_2;
  \end{equation}
however, an alternative method is updating $\mathrm{f}_1$ using the average of the transformed $\hat{\mathrm{f}}_2$ and reconstructed $\hat{\mathrm{f}}_1$ that is
  \begin{equation}
    \mathrm{f}_1 = \frac{1}{2}(\hat{\mathcal{T}}_{{\upzeta}} \hat{\mathrm{f}}_2 + \hat{\mathrm{f}}_1),  \label{Update_Alter_1}
  \end{equation}
in which we gather information of both $\hat{\mathrm{f}}_1$ and $\hat{\mathrm{f}}_2$. Furthermore, we can also incorporate the inverse transformation of $\mathrm{f}_1$ into the correction of $\mathrm{f}_2$. However, not all transformations have an analytical inverse.

To sum up, our iterative method alternately performs incomplete reconstructions for two temporal data sets, followed by a registration.

%______________________________________________________________________________________________________________________
% Section 5: Simultaneous Method
\section{Simultaneous Method}
\label{sec:simultaneous_method}

\subsection{Formulation of the Simultaneous Method}

Whilst combined reconstruction and registration algorithms have been applied to other modalities ({\it e.g.,} PET, SPECT and MRI), little has been published on applying these techniques to DBT. We have proposed an iterative method, which {\it partially coupled} the two tasks by alternating between optimising image intensities and parametric transformations to obtain a reduced objective functional \citep{Yang2010a, Yang2010b}. An alternative to registering the images after reconstruction or partially coupling them, is to perform the two tasks simultaneously ({\it fully coupled}). This avoids the assumptions of missing data being equal to zero (implicit in algorithms such as FBP). The work in this section hypothesises that the two tasks are not independent but reciprocal, and that combining them will enhance the performance of both \cite{Yang2011}.

Using this hypothesis, we have developed an algorithm, which outputs one unified result for the reconstruction and registration (Algorithm \ref{alg:Simultaneous_Method}). However, the introduction of the  nonlinear parametric transformation renders the solution of the inverse problem more complex.

We solve the inverse problem by forming the objective function given by
\begin{align}
  \{ \mathrm{f}^{\star},\upzeta^{\star} \} &= \arg\min_{\mathrm{f},\upzeta}~\Big(~f({\mathrm{f},\upzeta})~\Big), \\
  f({\mathrm{f},\upzeta}) &= \frac{1}{2} \Big(\big\| A\mathrm{f}-\mathrm{p}_1\big\|^2 + \big\| A\mathcal{T}_{{\upzeta}}\mathrm{f}-\mathrm{p}_2\big\|^2 \Big),
  \label{Unconstrained_optimisation_Cost_Function}
\end{align}
in which $\mathrm{f}$ denotes the estimation of the unknown volume, and $\upzeta$ is the estimation of the unknown parametric transformations \citep{Yang2012a,Yang2012b,Yang2012c}.

%\begin{center}
%\centering
%\scalebox{0.888}{
    \begin{algorithm}
    \label{alg:Simultaneous_Method}
    %\dontprintsemicolon
    \caption{Simultaneous Method}
      ~ \\[0.1cm]
      \KwIn{$\mathrm{p}_1,~\mathrm{p}_2.$}
      \KwOut{$\mathrm{f}^\star,~\upzeta^\star.$}
      ~ \\
      \Begin{
		\% Initialise $\mathrm{f}$ to a vector with all zero entries; \rm \\
  		\% Initialise $\upzeta$ to a vector of identity matrix when we use affine transformation model; \rm \\
		\% (Initialise $\upzeta$ to a vector with all zero entries when we use B-spline model). \rm \\
		$\mathrm{f} := 0;$ \\
		$\upzeta := I;$ \\
  		~ \\[-0.05cm]
        \% Simultaneous reconstruction and registration loop \rm \\
        \For{$k~\mathrm{iterations}$} {
          ~ \\[0.01cm]
          $\{\mathrm{f}^{\star},\upzeta^{\star}\} = \arg\min_{\mathrm{f},\upzeta}~\Big(~f({\mathrm{f},\upzeta})~\Big)$; \\
          }
        ~ \\[0.2cm]
        \% Output $\mathrm{f}^\star$ and $\upzeta^\star$ \rm \\
        $\mathrm{f}^\star$; \\
        $\upzeta^\star$. \\
        }
      ~ \\[0.05cm] \label{algr:Simultaneous_Method}
    \end{algorithm}
%}
%\end{center} ~\\[-0.898cm]

A minimiser $\{ \mathrm{f}^{\star},\upzeta^{\star} \}$ of $f({\mathrm{f},\upzeta})$ is characterised by the necessary condition that the partial derivative with respect to  $\mathrm{f}$ and $\upzeta$ equals zero. The partial derivative with respect to  $\mathrm{f}$ is straightforward, and is given by
\begin{equation}
\begin{split}
  g(\mathrm{f}) &= \frac{\partial f({\mathrm{f},\upzeta})}{\partial \mathrm{f}} \\
  &= A^T(A\mathrm{f}-\mathrm{p}_1)+\mathcal{T}_{{\upzeta}}^\ast A^T(A\mathcal{T}_{{\upzeta}} \mathrm{f}-\mathrm{p}_2),
\end{split}
\end{equation}
in which $g(\mathrm{f})$ is the gradient with respect to $\mathrm{f}$, and $\mathcal{T}_{{\upzeta}}^\ast$ is the adjoint operator of $\mathcal{T}_{{\upzeta}}$.

To derive the partial derivative with respect to $\upzeta$, we apply a small perturbation to the objective function.  The linearisation via the norm then yields,
\begin{align}
  f&\Big( \mathrm{f},\upzeta + \Delta \upzeta \Big) \notag \\
  & = \frac{1}{2} \Big(\big\| A\mathrm{f}-\mathrm{p}_1\big\|^2+\big\| A\mathcal{T}_{{\upzeta}+\Delta \upzeta}\mathrm{f}-\mathrm{p}_2\big\|^2 \Big) \\
  & \approx \frac{1}{2} \Big(\big\| A\mathrm{f}-\mathrm{p}_1\big\|^2 + \big\| A\mathcal{T}_{{\upzeta}}\mathrm{f} + A \frac{\partial \mathcal{T}_{{\upzeta}}}{\partial\upzeta} \mathrm{f} \Delta \upzeta - \mathrm{p}_2\big\|^2 \Big). \notag
\end{align}

By taking the derivative with respect to  $\Delta \upzeta$, and equating the result to zero, we obtain:
\begin{equation}
  \Big( A \frac{\partial \mathcal{T}_{{\upzeta}}}{\partial\upzeta} \mathrm{f} \Big)^T \Big( A\mathcal{T}_{{\upzeta}}\mathrm{f} + A \frac{\partial \mathcal{T}_{{\upzeta}}}{\partial\upzeta} \mathrm{f} \Delta \upzeta - \mathrm{p}_2 \Big) = 0;
\end{equation}
If $g(\upzeta)$ denotes the gradient then we have,
\begin{equation}
\begin{split}
  g(\upzeta) = \frac{\partial f({\mathrm{f},\upzeta})}{\partial \upzeta} &= \Big( A \frac{\partial \mathcal{T}_{{\upzeta}}}{\partial\upzeta} \mathrm{f} \Big)^T \Big( A\mathcal{T}_{{\upzeta}}\mathrm{f} - \mathrm{p}_2 \Big) \\ &= \Big( A \mathcal{T}_{{\upzeta}}^{'} \mathrm{f} \Big)^T \Big( A\mathcal{T}_{{\upzeta}} \mathrm{f}-\mathrm{p}_2 \Big).
\end{split}
\end{equation}

\subsection{Decoupled Solver for the Fully-coupled System}

Our survey of published simultaneous methods indicates that none of these studies solved the simultaneous reconstruction and registration problem directly (Table \ref{table:LiteratureCompare}). Similarly, we adopt a decoupled approach to solve the combined problem  because the objective function in Equation \ref{Unconstrained_optimisation_Cost_Function} is a nonconvex function of the transformation parameters $\upzeta$ and therefore very challenging to minimise. As in previous studies we simplify the simultaneous optimisation using the decoupled alternating minimisation technique, where we update $\mathrm{f}$ holding $\upzeta$ fixed and vice versa, {\it i.e.,} the $n+1$-th estimate is computed from  the $n$-th estimate as follows,
  \begin{align}
  \mathrm{f}^{n+1} &= \arg\min_{\mathrm{f}}~f\Big(\mathrm{f},~\upzeta^n\Big), \label{ReconDecoupled} \\
  \upzeta^{n+1} &= \arg\min_{\upzeta}~f\Big(\mathrm{f}^{n+1},~\upzeta~\Big), \label{RegnDecoupled}
  \end{align}
  and the gradients are given by
  \begin{align}
  g(\mathrm{f})^{n+1} &= A^T \Big(A\mathrm{f}-\mathrm{p}_1\Big) + \mathcal{T}_{{\upzeta}^n}^\ast A^T \Big(A\mathcal{T}_{{\upzeta^n}} \mathrm{f}-\mathrm{p}_2\Big), \label{ReconDecoupledG} \\
  g(\upzeta)^{n+1} &= \Big( A \mathcal{T}_{{\upzeta}}^{'} \mathrm{f}^{n+1} \Big)^T \Big( A\mathcal{T}_{{\upzeta}} \mathrm{f}^{n+1} - \mathrm{p}_2 \Big). \label{RegnDecoupledG}
  \end{align}

Solving the simultaneous inverse problem using the decoupled optimisation is totally different from the sequential method or the iterative method described in \cite{Yang2010a} and \cite{Yang2010b}. First, the gradient with respect to the image intensities is not a simple addition of the derivative in equations \ref{Complete_Reconstruction_1} and \ref{Complete_Reconstruction_2} because here we estimate a single volume rather than two as in the sequential or iterative methods. Accordingly, we have one unified result instead of two reconstructions, which need to be registered in a further step. More significantly, because of the presence of the system matrix $A$ in the gradient formulation in equation \ref{RegnDecoupledG}, the simultaneous concept is more challenging than a typical image registration problem described in equation \ref{Complete_Registration}, {\it i.e.,} registering two complete reconstructions in the sequential method or registering the two current estimates of the incomplete reconstructions in the iterative method.

  \begin{table*}[!htbp]
  \caption{\small\emph{Comparison of the gradient information used in the iterative and simultaneous methods.}}
  \centering
  \scalebox{0.758}{
  \setlength{\floatsep}{10pt plus 3pt minus 2pt} % Set the length between the table and the text body
  \begin{tabular}{p{4.8cm}p{7.18cm}^p{9.8cm}}
  \addlinespace
  \toprule\rowstyle{\bfseries}

  & \multicolumn{1}{c}{Reconstruction Part} & \multicolumn{1}{c}{Registration Part} \\
  \midrule
  \multicolumn{1}{l}{Iterative Method} & $g(\mathrm{f}_1) = A^T(A\mathrm{f}_1-\mathrm{p}_1)$ & \multicolumn{1}{l}{$g(\upzeta) = \big( \hat{\mathrm{R}}(\mathrm{x}) - \hat{\mathrm{T}}[\mathcal{T}_{{\upzeta}}(\mathrm{x})] \big) \frac{ \partial \hat{\mathrm{T}}[\mathcal{T}_{{\upzeta}}(\mathrm{x})] }{ \partial \mathcal{T}_{{\upzeta}}(\mathrm{x}) } \frac{ \partial \mathcal{T}_{{\upzeta}}(\mathrm{x}) }{ \partial \upzeta }$} \\ \cline{2-2} \\[-4mm]
  \multicolumn{1}{l}{} & $g(\mathrm{f}_2) = A^T(A\mathrm{f}_2-\mathrm{p}_2)$ & \multicolumn{1}{l}{} \\
  \bottomrule
  \toprule
  & \multicolumn{1}{c}{Intensity Part} & \multicolumn{1}{c}{Transformation Part} \\
  \midrule
  Simultaneous Method & \multicolumn{1}{l}{$g(\mathrm{f}) = A^T(A\mathrm{f}-\mathrm{p}_1)+\mathcal{T}_{{\upzeta}}^\ast A^T(A\mathcal{T}_{{\upzeta}} \mathrm{f}-\mathrm{p}_2)$} & \multicolumn{1}{l}{$g(\upzeta) = \Big( \frac{ \partial \mathrm{T}[\mathcal{T}_{{\upzeta}}(\mathrm{x})] }{ \partial \mathcal{T}_{{\upzeta}}(\mathrm{x}) }^{\ast} \Big) \frac{ \partial \mathcal{T}_{{\upzeta}}(\mathrm{x}) }{ \partial \upzeta } A^T \big( A\mathcal{T}_{{\upzeta}} \mathrm{f}-\mathrm{p}_2 \big)$}\\

  \bottomrule
  \end{tabular}
  }
  \label{table:Gradient_Information_Summary}
  \end{table*}

\subsection{Derivative Operator of the Transformations}

The derivative of the transformation operation is a key component of the algorithm and has great impact on the result of the optimisation. Deriving an analytical derivative of the transformation is desirable because it would be fast to compute but is complicated by the need to formulate the derivative of the underlying interpolation. In addition, some interpolation schemes have no analytical derivative. For this reason therefore, we use the Finite Difference Method (FDM) to approximate the derivative operation:
  \begin{equation}
    \mathcal{T}_{{\upzeta}}^{'} \approx \frac{ \mathcal{T}_{{\upzeta+\epsilon}} + \mathcal{T}_{{\upzeta-\epsilon}} }{2\epsilon}
  \end{equation}
where $\epsilon$ is a small number.

\subsection{Optimisation}

The optimisation is performed using a quasi-Newton based L-BFGS method, which is described as a generic form in Algorithm \ref{algr:BFGS}. This approximates the inverse of the Hessian matrix whilst avoiding the considerable memory overhead (for large DBT data sets) associated with computing 2nd order derivatives or their fully dense approximations directly.

%\begin{center}
%  \centering
%  \scalebox{0.858}{
  \begin{algorithm}
  \caption{quasi-Newton Method}

  \mbox{} \\
  
  \KwIn{$k,~\mathrm{f}_{\mathrm{initial}}.$}
  \KwOut{$\mathrm{f}_{\mathrm{optimised}}.$}

  \mbox{} \\
  $H_{\mathrm{initial}} := I;$ \% Initialise the inverse Hessian matrix as identity matrix \\
  $\mathrm{d}_{\mathrm{initial}} := -H_{\mathrm{initial}}g(\mathrm{f}_{\mathrm{initial}});$ ~~~~\bf \it \% Initial search direction is the negative gradient \rm \\

  \Begin{ 
  		\While{stopping criterion unfulfilled} {
  		
			$\tau_k := \arg\min_{\tau>0} f(\mathrm{f}_k + \tau \mathrm{d}_k);$  ~~~~\bf \it \% Line Search \rm \\
			$\mathrm{f}_{k+1} := \mathrm{f}_k + \tau_k \mathrm{d}_k;$ ~~~~\bf \it \% Update the $\mathrm{f}$ \rm \\

            $\mathrm{s}_k := \mathrm{f}_{k+1}-\mathrm{f}_k$; \\
            $\mathrm{z}_k := g(\mathrm{f}_{k+1})-g(\mathrm{f}_k);$ \\
            
            \mbox{} \\

            $H_{k+1} := H_k + \frac{1}{\mathrm{s}_k^T \mathrm{z}_k} \Bigg[
            \Big( 1 + \frac{\mathrm{z}_k^T H_k \mathrm{z}_k}{\mathrm{s}_k^T \mathrm{z}_k} \Big) \mathrm{s}_k \mathrm{s}_k^T
            -H_k \mathrm{z}_k \mathrm{s}_k^T - \mathrm{s}_k \mathrm{z}_k^T H_k \Bigg];$ ~~~~\bf \it \% BFGS \rm  \\

			\mbox{} \\

            $d_{k+1} := -H_{k+1} g(\mathrm{f}_{k+1});$ ~~~~\bf \it \% Update the search direction \rm \\

			\mbox{} \\

            $k := k+1$;
            
            \mbox{} \\
      }
    $\mathrm{f}_{\mathrm{optimised}} := \mathrm{f}_{k+1};$ }
    \label{algr:BFGS}
  \end{algorithm}
%  }
%  \end{center}

%______________________________________________________________________________________________________________________
% Section 6: Results and Discussion
\section{Results}
\label{sec:results}

In this section, we demonstrate the performance of our combined reconstruction and registration framework using both iterative and simultaneous methods. Both affine and B-spline transformation models have been considered. First, we combine optimisation of the two temporal reconstructions with the $12$ degrees of freedom, of an affine transformation, which globally describes the translation, scaling, rotation and shearing between the two time points. Second, we can also substitute non-rigid B-spline deformations for the affine transformation in this framework. We begin in Section \ref{sec:Seqen_vs_Iter} using an affine transformation model, and test this using a software synthetic toroidal phantom image. We can compare the final transformed moving image $\hat{\mathcal{T}}_{{\upzeta}} \hat{\mathrm{f}}_2$ with the original fixed image ${\mathrm{f}}_1^{\mathrm{g}}$, which is the ground truth of the reconstruction, to analyse the accuracy of our iterative method. The difference image, between $\hat{\mathcal{T}}_{{\upzeta}} \hat{\mathrm{f}}_2$ and ${\mathrm{f}}_1^{\mathrm{g}}$, is compared with the difference image of the conventional sequential method, {\it i.e.,} differences between $\mathcal{T}^\star_{{\upzeta}} \mathrm{f}_2^\star$ and ${\mathrm{f}}_1^{\mathrm{g}}$, in which $\mathcal{T}^\star$ is calculated using ${\upzeta}^\star$ from Equation \ref{Complete_Registration}, and $\mathrm{f}_2^\star$ is obtained from Equation \ref{Complete_Reconstruction_2}. In Section \ref{sec:Seqen_vs_Simult}, we qualitatively and quantitatively assess the performance of our simultaneous method using the affine transformation model, and test on various simulated data sets. In addition, both our iterative and simultaneous methods are compared with the conventional sequential method. Finally, we analyse the efficacy of incorporating a non-rigid B-spline transformation model into our simultaneous method in Section \ref{sec:Simult_Bspline}.

% Finally, Section \ref{sec:Simult_Detect_Changes} addresses the problem of detecting changes using our combined framework.

\subsection{Sequential Method {\it vs.} Iterative Method} \label{sec:Seqen_vs_Iter}

\begin{figure}[!htb]
  \begin{center}
  \centering % \setlength{\floatsep}{10pt plus 3pt minus 2pt}
        \subfigure[]{
    \includegraphics[width=0.15\textwidth]{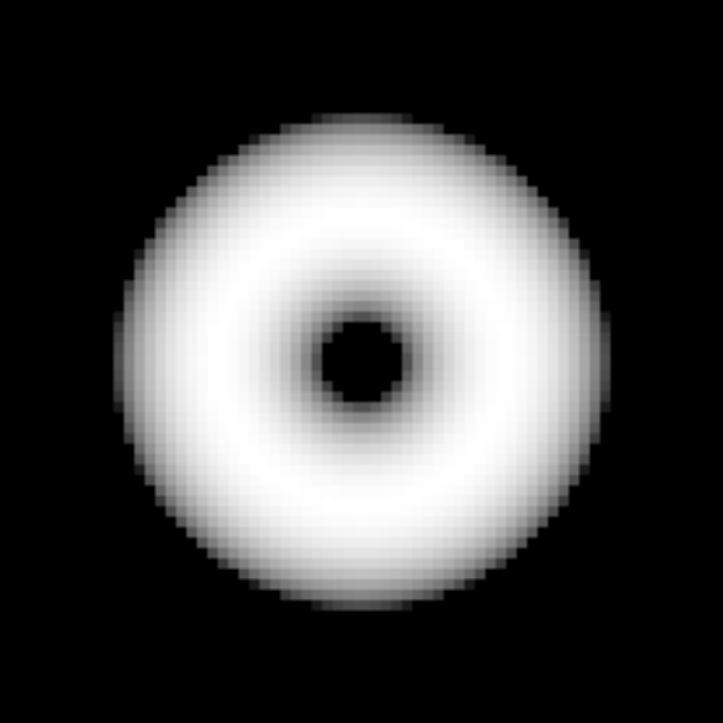}}
        \subfigure[]{
    \includegraphics[width=0.15\textwidth]{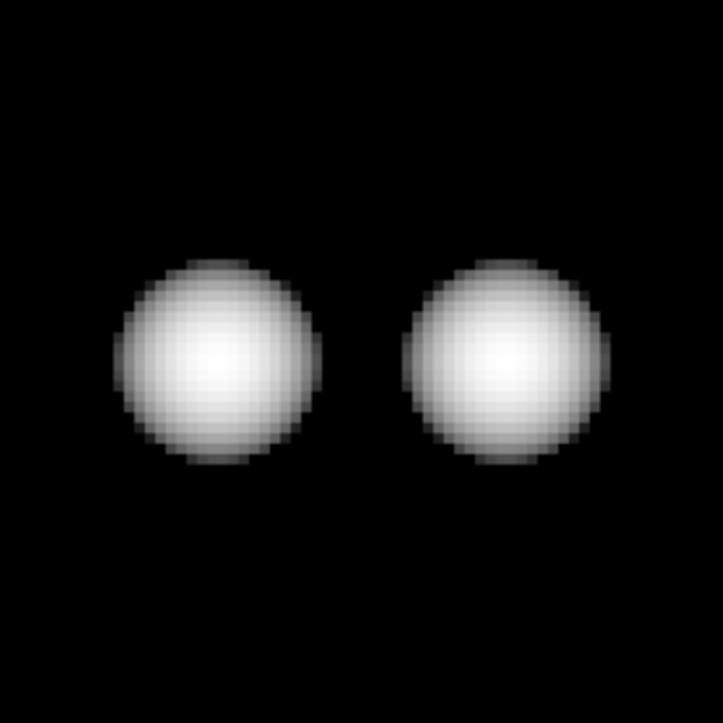}}
        \subfigure[]{
    \includegraphics[width=0.15\textwidth]{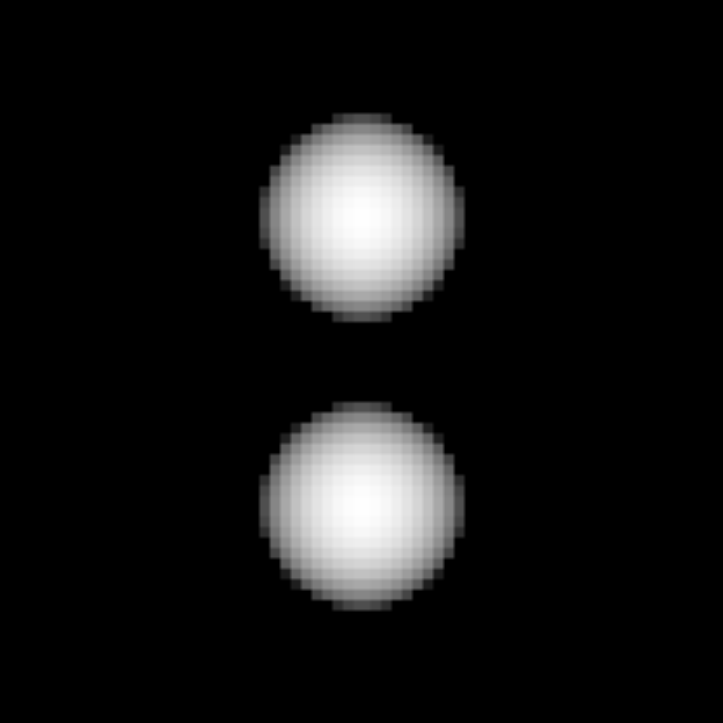}}  \\
        \subfigure[]{
    \includegraphics[width=0.15\textwidth]{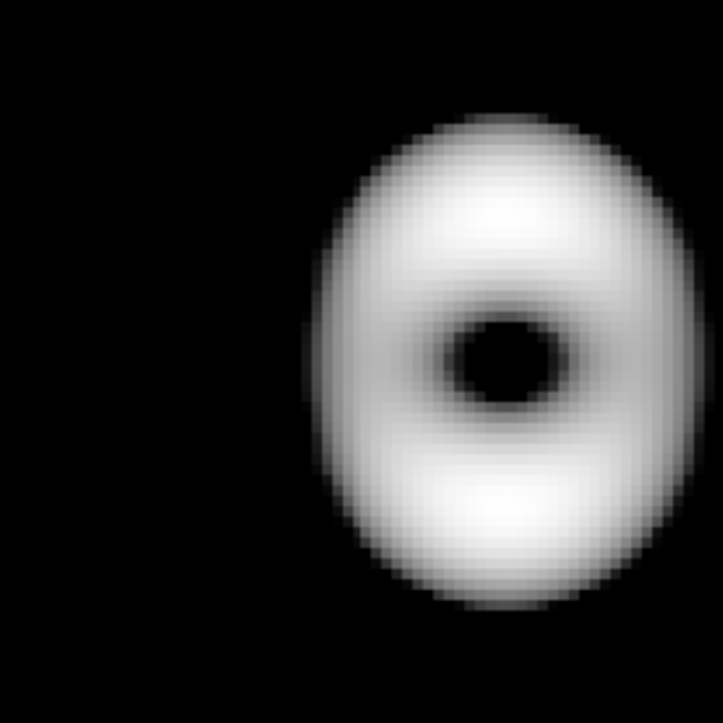}}
        \subfigure[]{
    \includegraphics[width=0.15\textwidth]{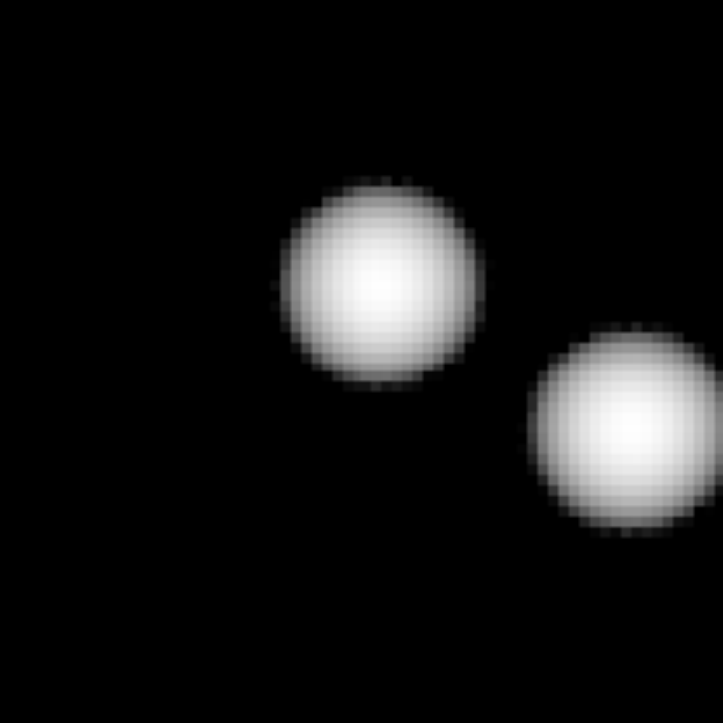}}
        \subfigure[]{
    \includegraphics[width=0.15\textwidth]{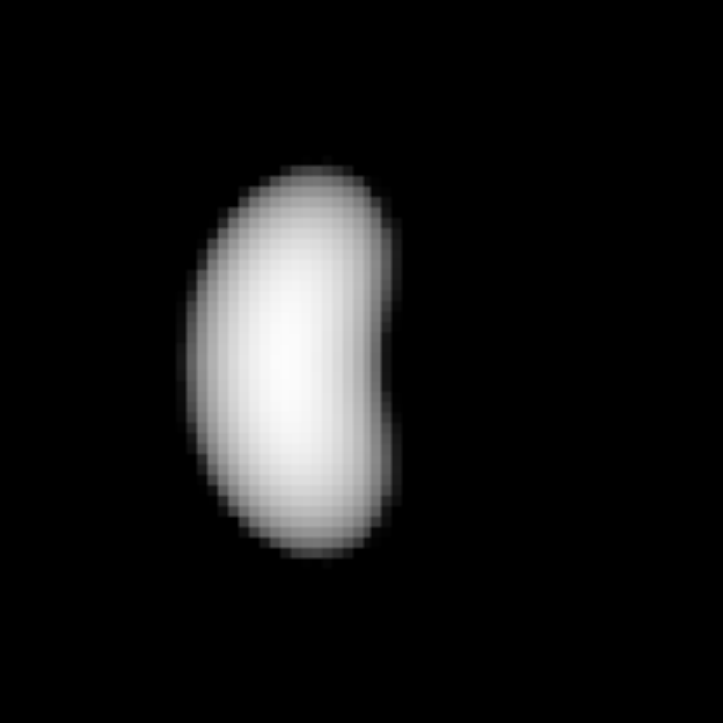}} \\
        \subfigure{
    \includegraphics[width=0.45\textwidth]{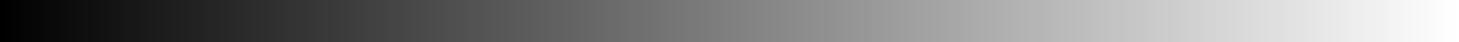}}
	\figcaption{\small\bf\it Toroid phantom. (a)-(c): Fixed image; (d)-(f): Moving image. (Left: Coronal view; Middle: Transverse view; Right: Sagittal view.)}
  \label{fig:Toroid3D_Iterative_Method}
  \end{center}
  \end{figure}

\begin{figure}[!htb]
  \begin{center}
  \centering % \setlength{\floatsep}{10pt plus 3pt minus 2pt}
        \subfigure[]{
    \includegraphics[width=0.15\textwidth]{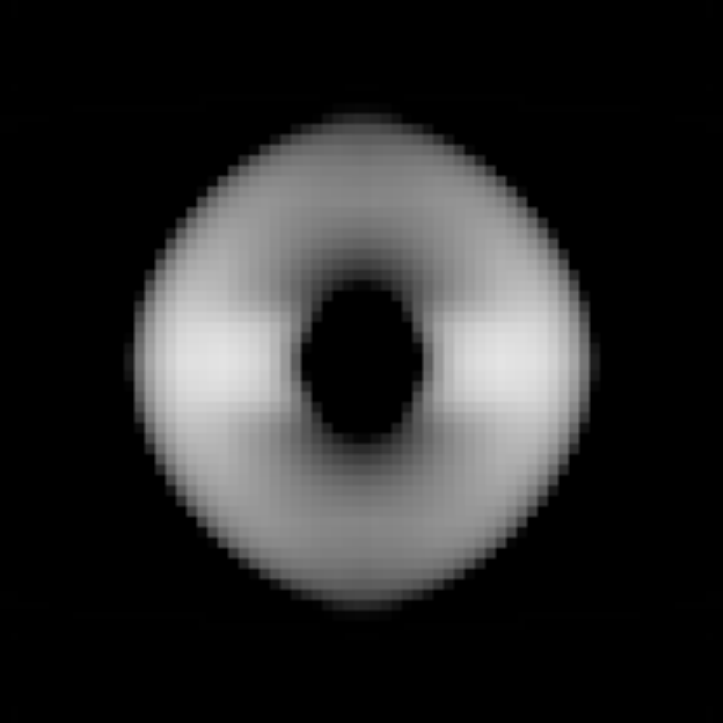}}
        \subfigure[]{
    \includegraphics[width=0.15\textwidth]{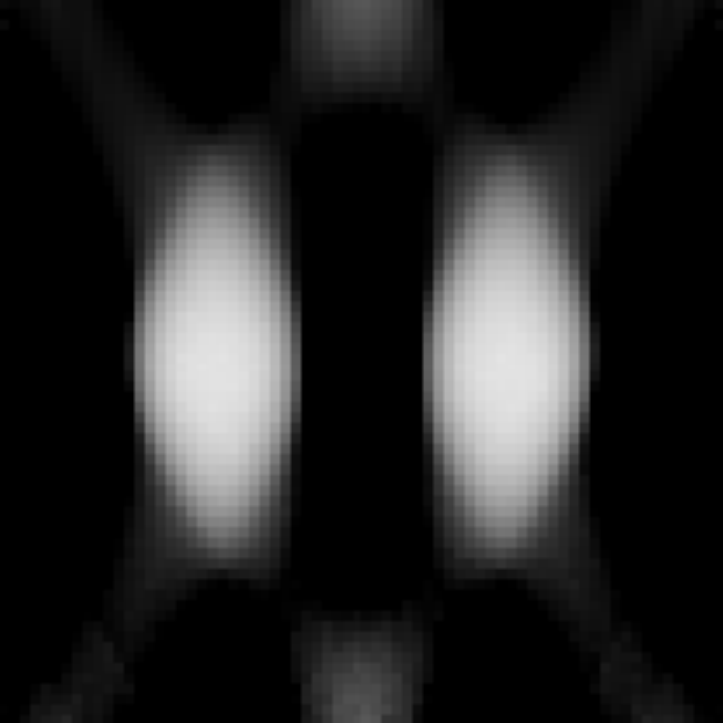}}
        \subfigure[]{
    \includegraphics[width=0.15\textwidth]{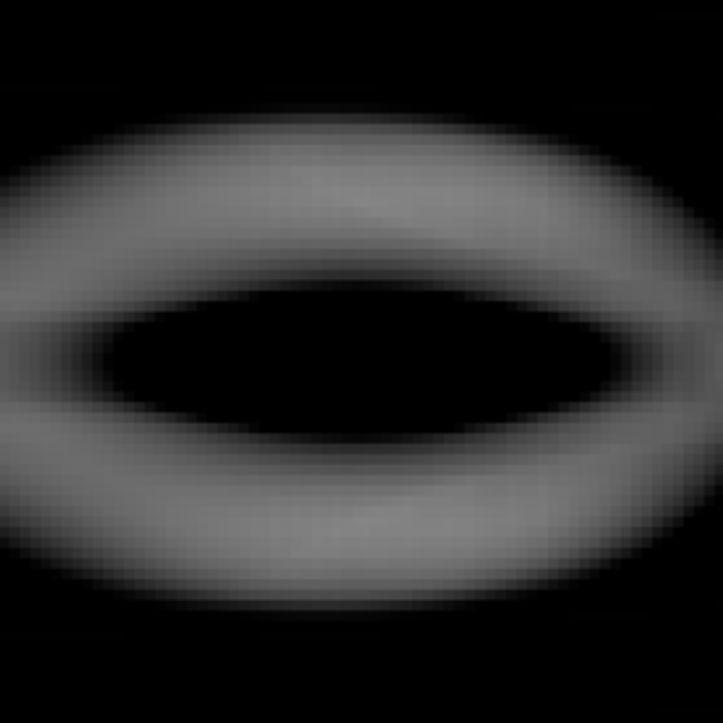}}  \\
        \subfigure[]{
    \includegraphics[width=0.15\textwidth]{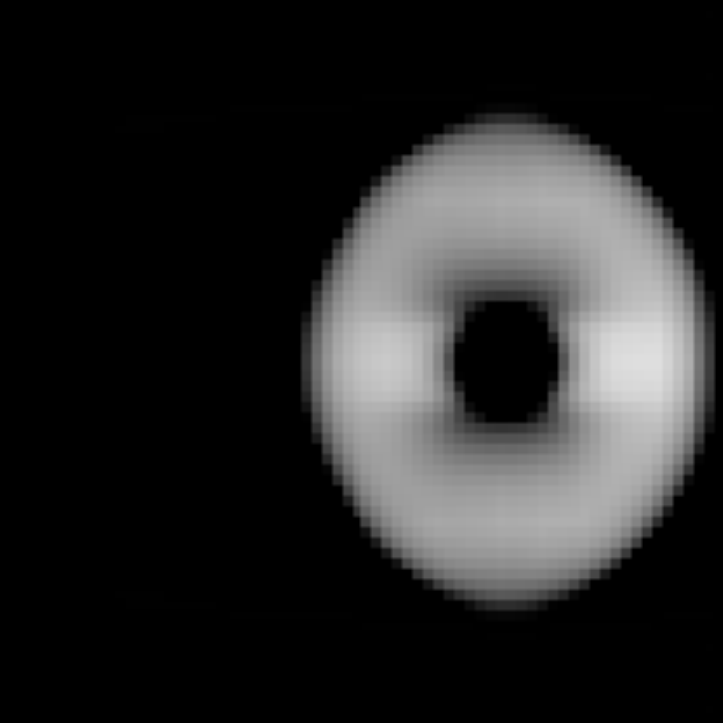}}
        \subfigure[]{
    \includegraphics[width=0.15\textwidth]{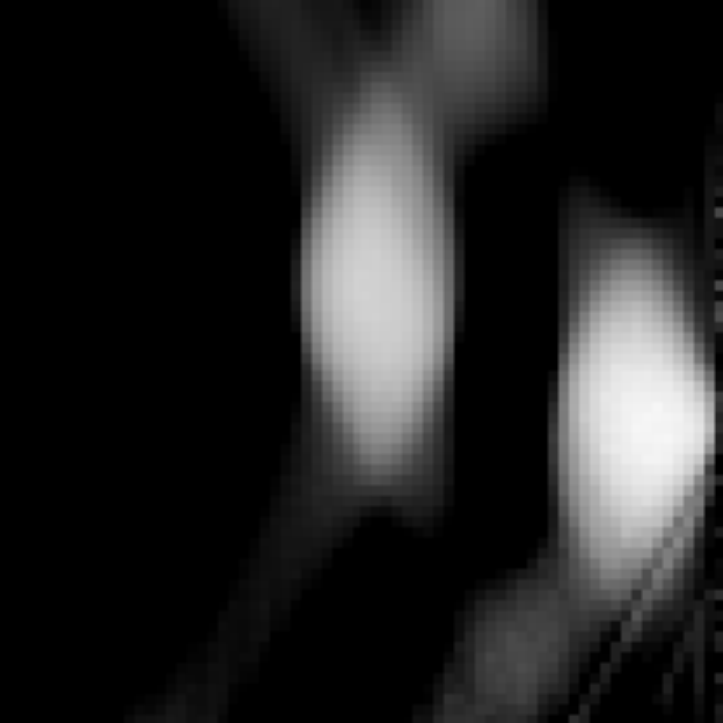}}
        \subfigure[]{
    \includegraphics[width=0.15\textwidth]{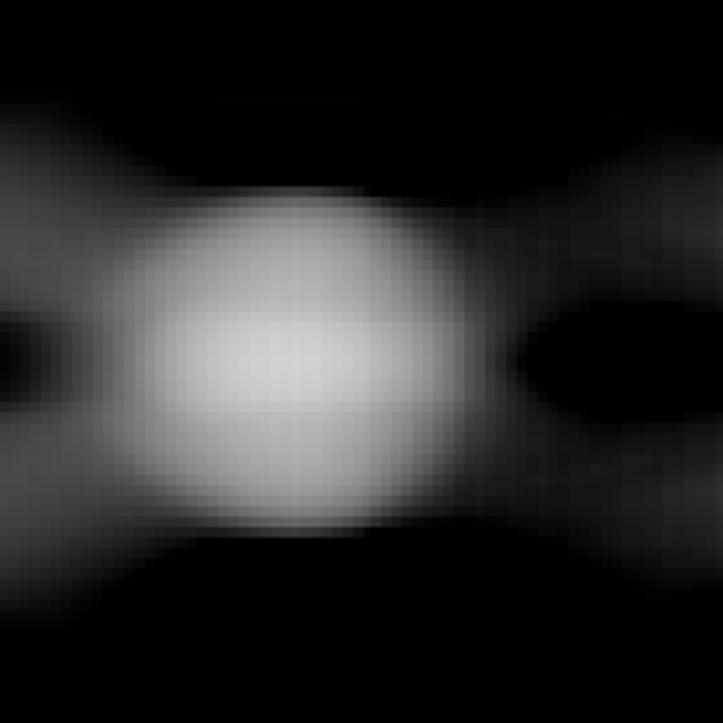}} \\
        \subfigure{
    \includegraphics[width=0.45\textwidth]{PhantomImage_70_11_1_4.pdf}}
	\figcaption{\small\bf\it Reconstruction using two acquisitions $\mathrm{p}_1$ and $\mathrm{p}_2$ respectively without registration. (a)-(c): Fixed image reconstruction; (d)-(f): Moving image reconstruction.}
  \label{fig:ReconNoRegnToroid3D}
  \end{center}
  \end{figure}

\begin{figure}[!htb]
  \begin{center}
  \centering % \setlength{\floatsep}{10pt plus 3pt minus 2pt}
        \subfigure[]{
    \includegraphics[width=0.15\textwidth]{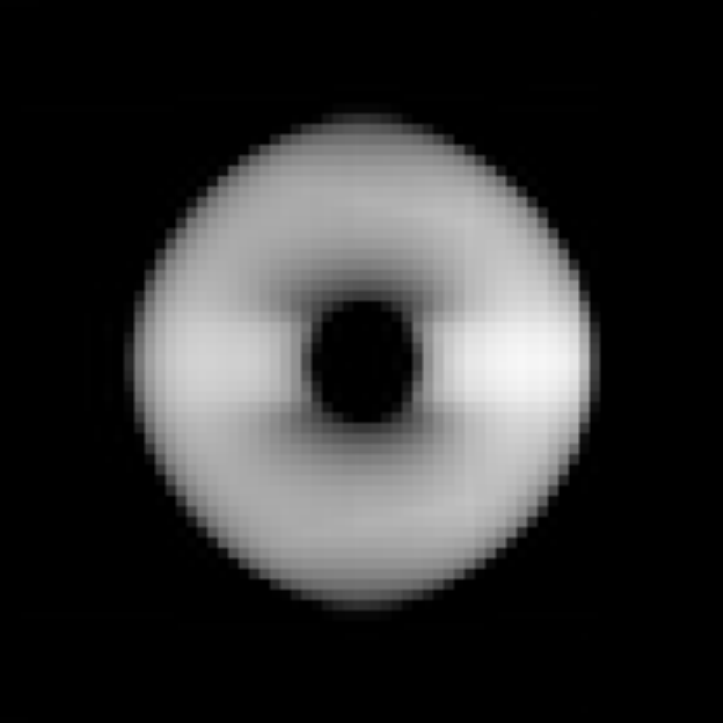}}
        \subfigure[]{
    \includegraphics[width=0.15\textwidth]{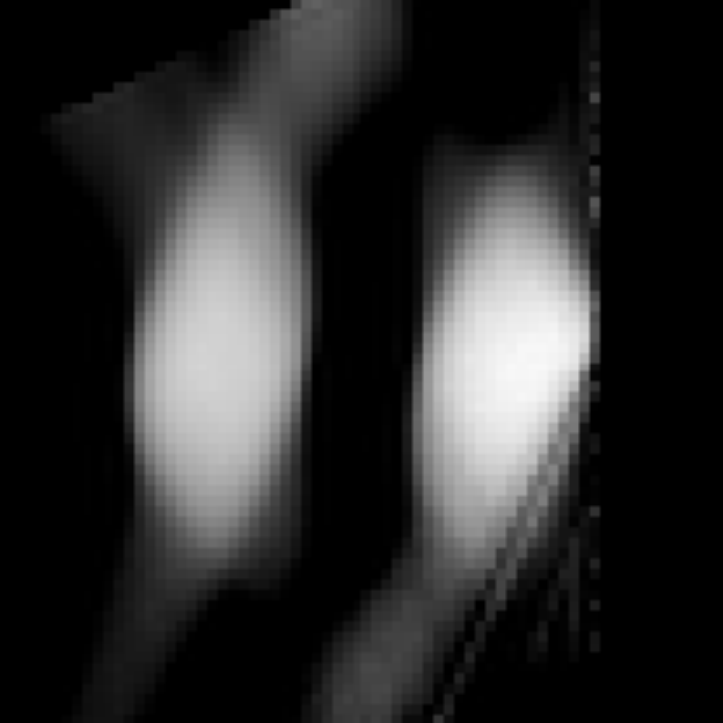}}
        \subfigure[]{
    \includegraphics[width=0.15\textwidth]{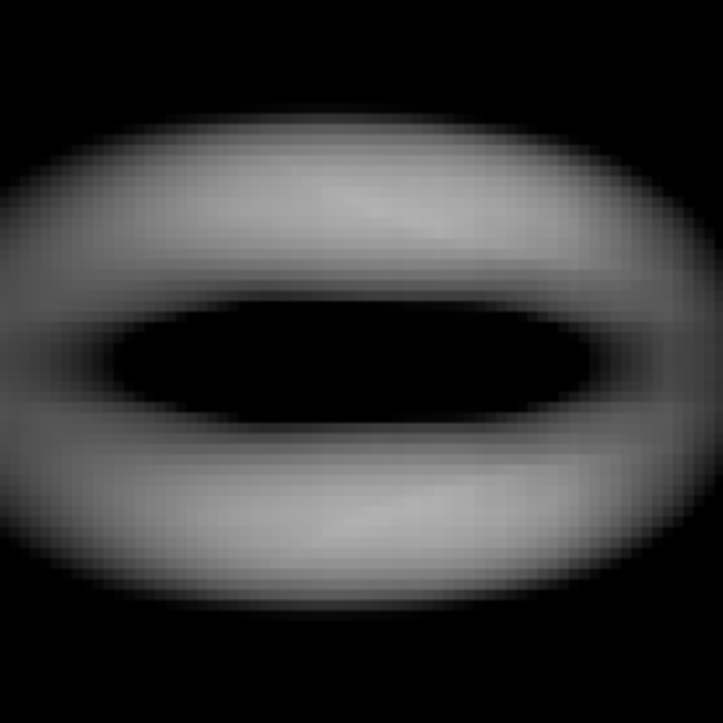}}  \\
        \subfigure[]{
    \includegraphics[width=0.15\textwidth]{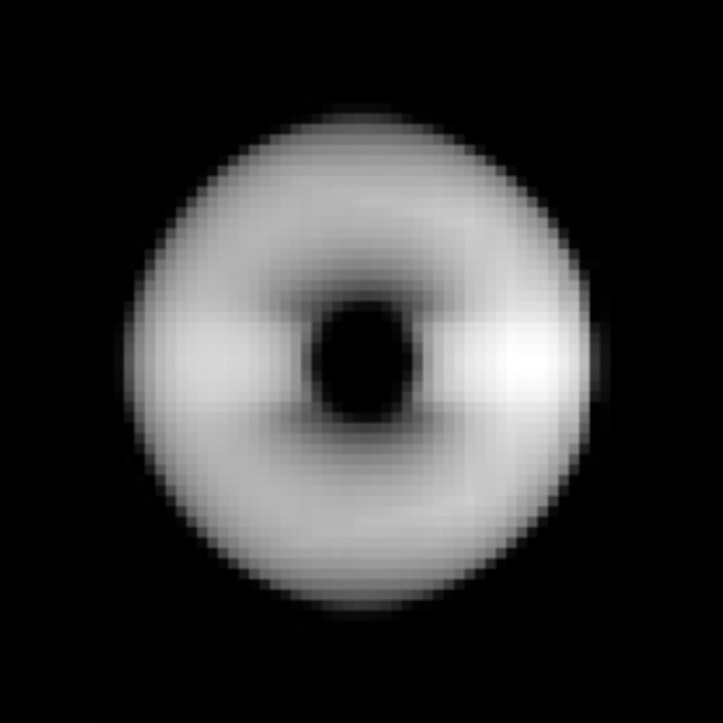}}
        \subfigure[]{
    \includegraphics[width=0.15\textwidth]{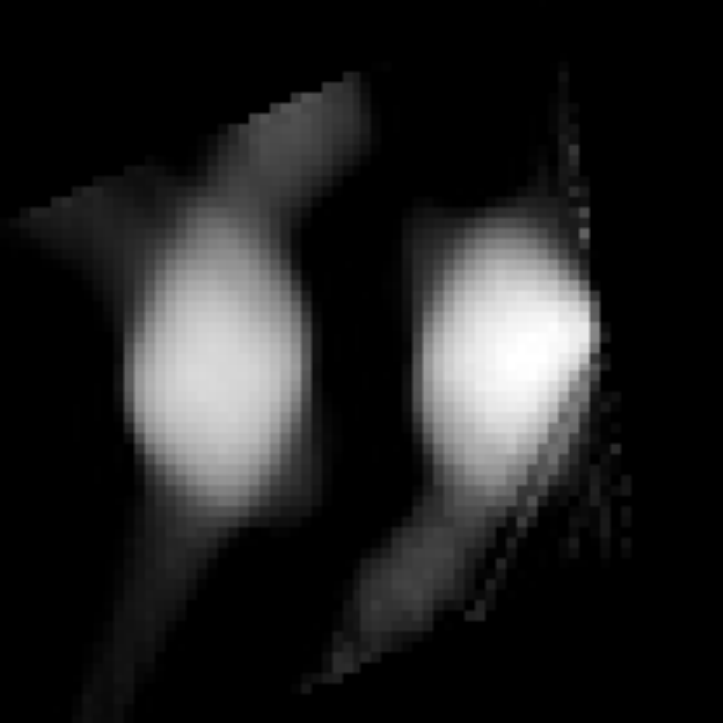}}
        \subfigure[]{
    \includegraphics[width=0.15\textwidth]{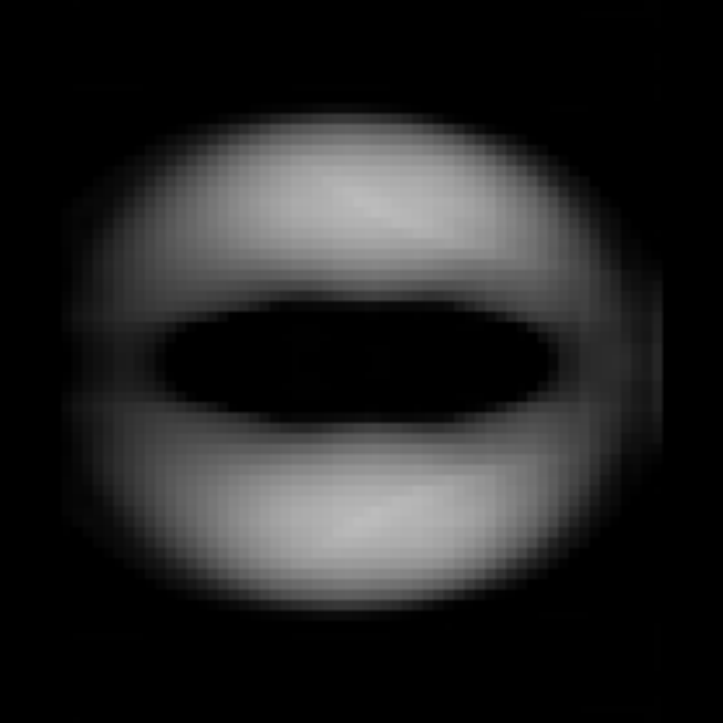}} \\
        \subfigure[]{
    \includegraphics[width=0.15\textwidth]{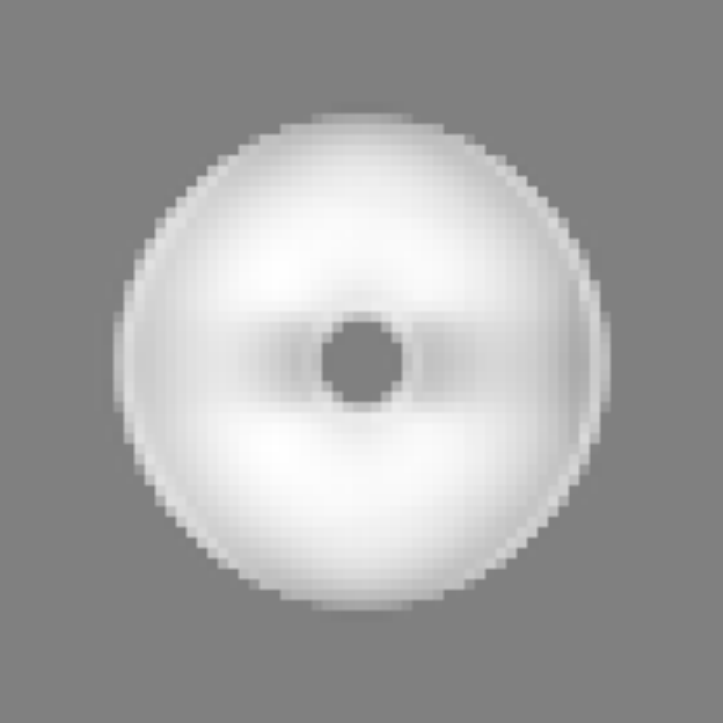}}
        \subfigure[]{
    \includegraphics[width=0.15\textwidth]{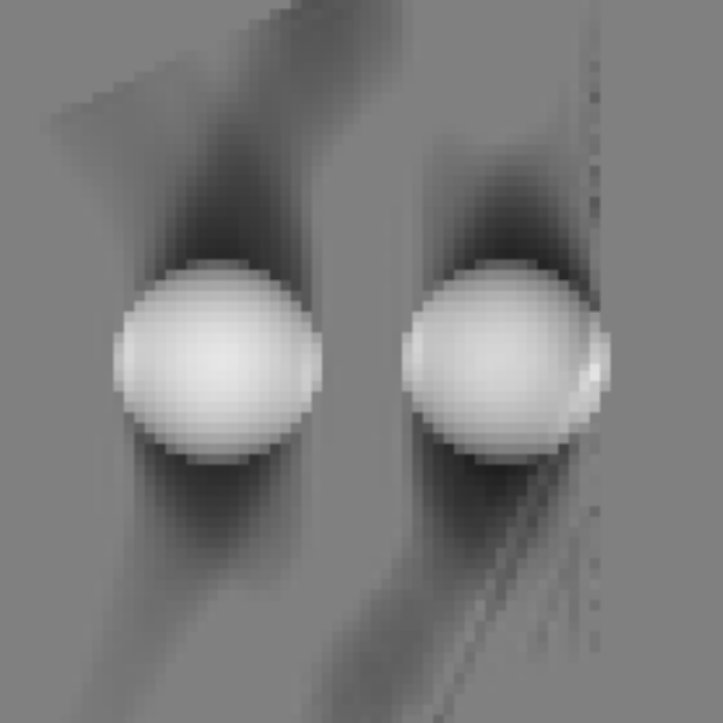}}
        \subfigure[]{
    \includegraphics[width=0.15\textwidth]{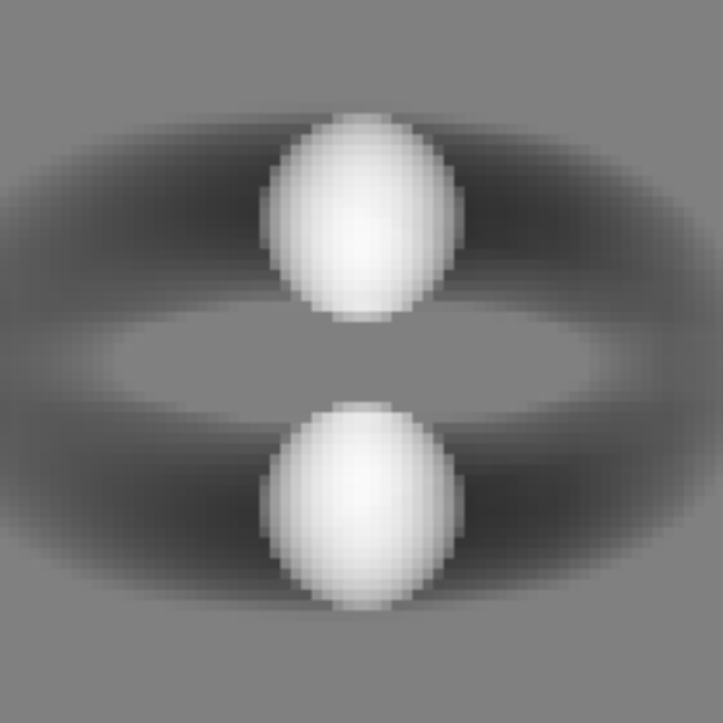}} \\
        \subfigure[]{
    \includegraphics[width=0.15\textwidth]{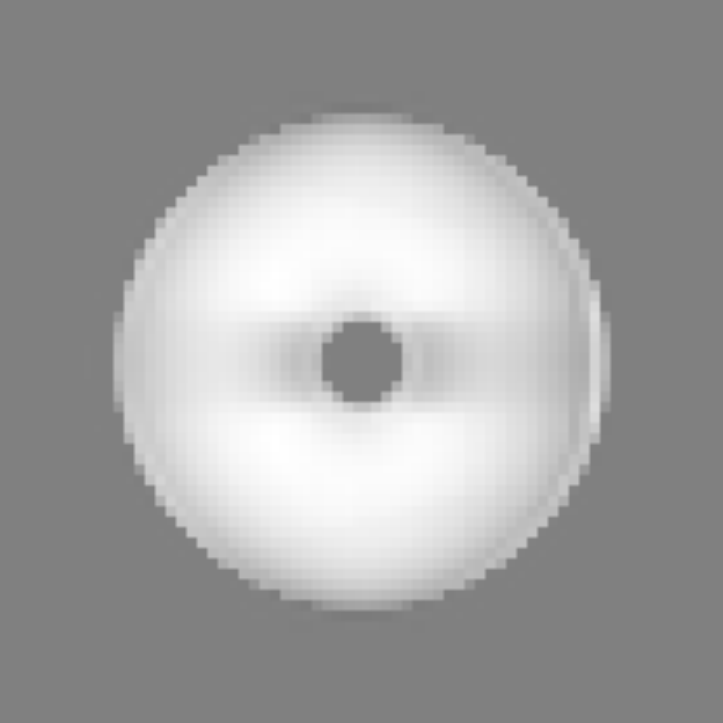}}
        \subfigure[]{
    \includegraphics[width=0.15\textwidth]{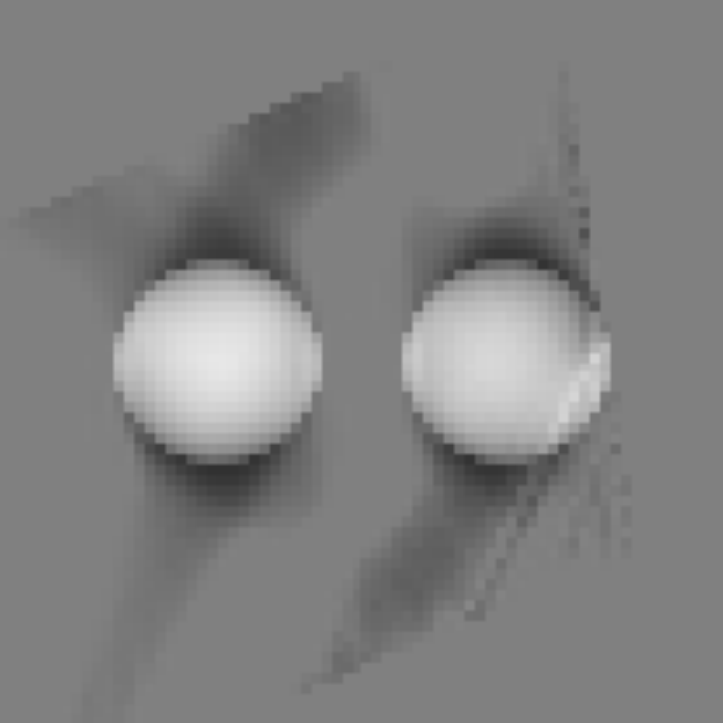}}
        \subfigure[]{
    \includegraphics[width=0.15\textwidth]{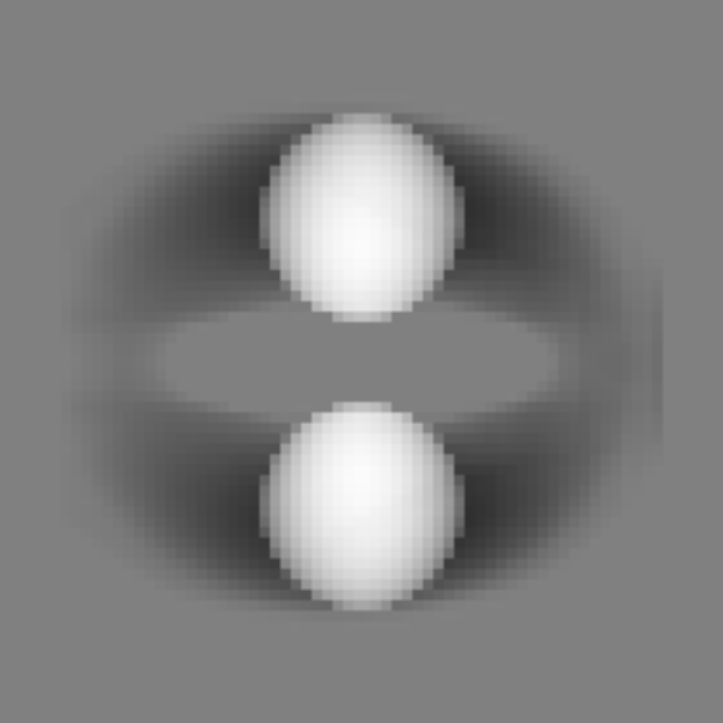}} \\
        \subfigure{
    \includegraphics[width=0.45\textwidth]{PhantomImage_70_11_1_4.pdf}}
	\figcaption{\small\bf\it Sequential results {\it vs.} iterative results. (a)-(c): Sequential result; (d)-(f): Iterative result; (g)-(i): Difference image between the sequential result and the fixed image (Figure \ref{fig:Toroid3D_Iterative_Method} (a)-(c)); (j)-(l): Difference image between the iterative result and the fixed image (Figure \ref{fig:Toroid3D_Iterative_Method} (a)-(c)).}
  \label{fig:Toroid__Iterative_Results}
  \end{center}
  \end{figure}

\begin{figure}[!htb]
  \begin{center}
  \centering % \setlength{\floatsep}{10pt plus 3pt minus 2pt}
  \includegraphics[width=0.8\textwidth]{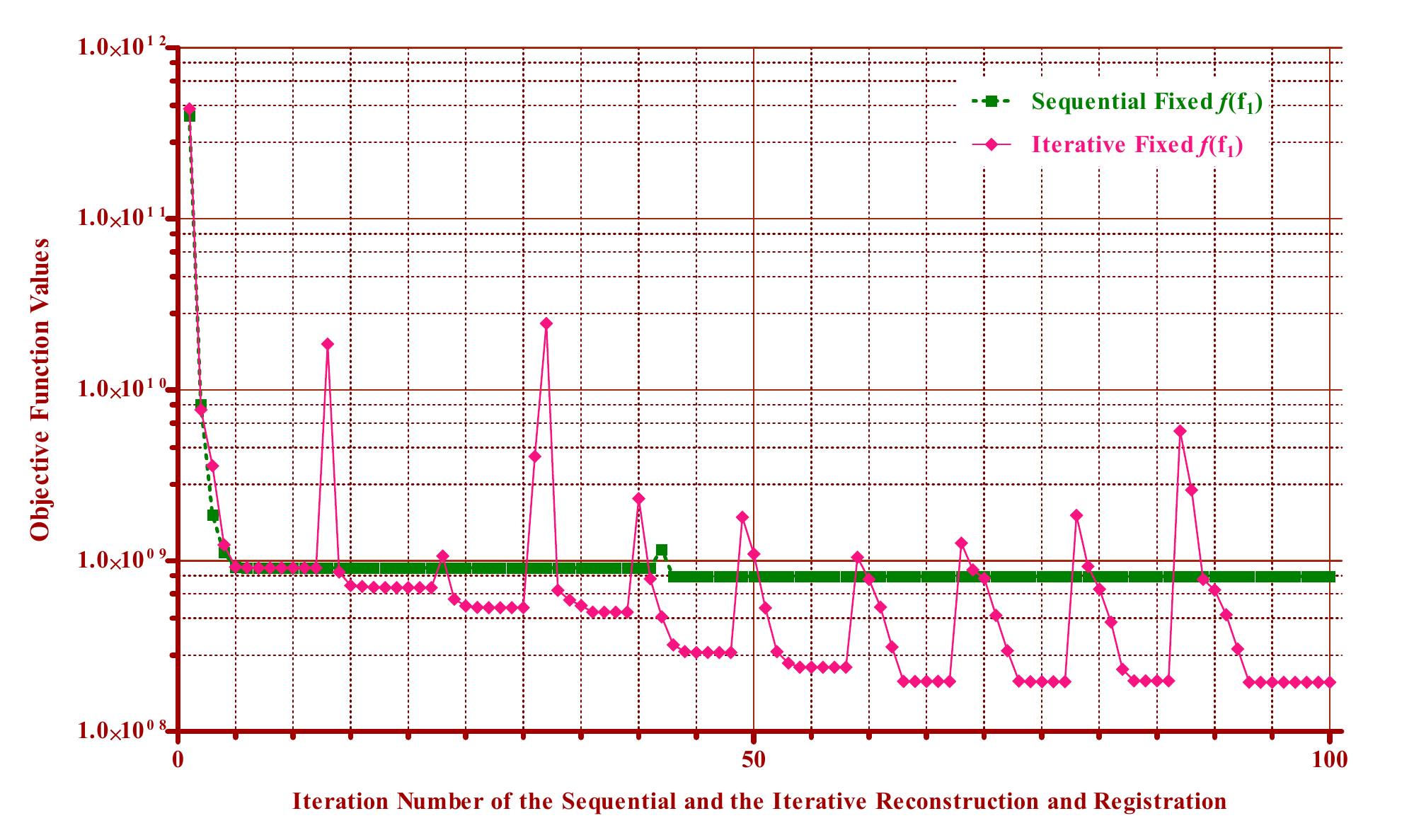}
  \caption{\small\emph{Plot of the objective function $f(\mathrm{f}_1) = \frac{1}{2}\big\|A\mathrm{f}_1-\mathrm{p}_1\big\|^2$ for the fixed image using both sequential and iterative methods.}}
  \label{fig:Toroid_Cost_Function}
  \end{center}
  \end{figure}

In this experiment, we created a software synthetic toroidal phantom, which is embedded in a 3D volume of $70\times70\times70$mm$^3$ with $1$mm resolution in each direction (Figure \ref{fig:Toroid3D_Iterative_Method} (a)-(c)). The ground truth affine transformation is a translation of $[10, 0, -20]$ mm and a rotation about the $y$ axis of $-30^{\mathrm{o}}$ (Figure \ref{fig:Toroid3D_Iterative_Method} (d)-(f)). The reconstructions of the fixed and moving images without registration are shown in Figure \ref{fig:ReconNoRegnToroid3D} (a)-(c) and (d)-(f) respectively. As seen in Figure \ref{fig:Toroid__Iterative_Results} (a)-(c), the {\it iterative} results are more compact and accurate than the {\it sequential results}, {\it i.e.,} the transformed moving image reconstruction $\mathcal{T}^\star_{{\upzeta}} \mathrm{f}_2^\star$ (Figure \ref{fig:Toroid__Iterative_Results} (d)-(f)), and the out of plane blurring is reduced. The mean squared error (MSE $= \frac{1}{\mathrm{D}_3} \| {\mathrm{f}}_1^\star - {\mathrm{f}}^{\mathrm{g}} \|^2$) is decreased from $10^6$ to $10^4$ in order of magnitude; however, for the {\it iterative} method this value of $1.26\times10^4$ is superior to the {\it sequential} result of $2.01\times10^4$ ($\mathrm{D}_3$ is the total number of voxels). In addition, from lower value of the objective function, we can conclude that our iterative method outperformed the sequential method (Figure \ref{fig:Toroid_Cost_Function}).

% Results of the experiment 1

\subsection{Sequential Method {\it vs.} Simultaneous Method} \label{sec:Seqen_vs_Simult}

\subsubsection{Test on a Toroid Phantom Image}

\begin{figure}[!htb]
  \begin{center}
  \centering % \setlength{\floatsep}{10pt plus 3pt minus 2pt}
        \subfigure[]{
    \includegraphics[width=0.15\textwidth]{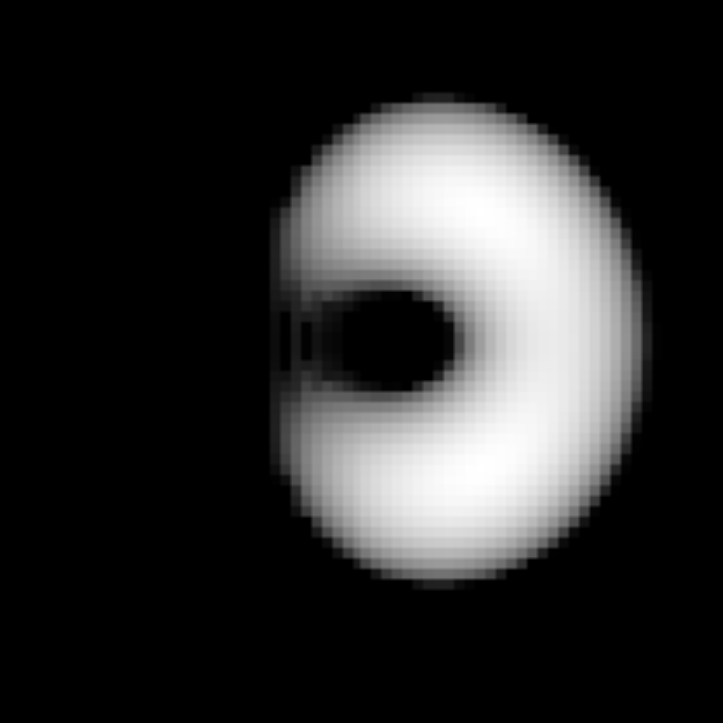}}
        \subfigure[]{
    \includegraphics[width=0.15\textwidth]{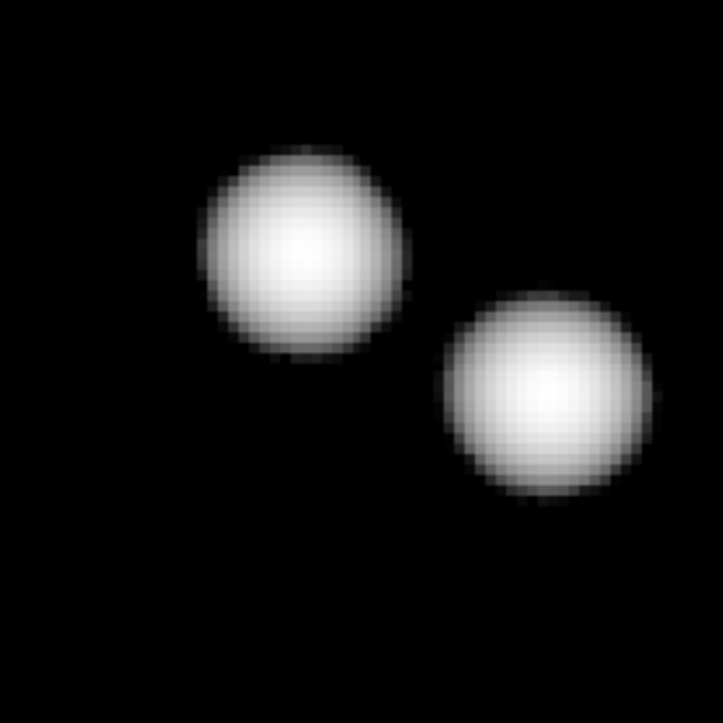}}
        \subfigure[]{
    \includegraphics[width=0.15\textwidth]{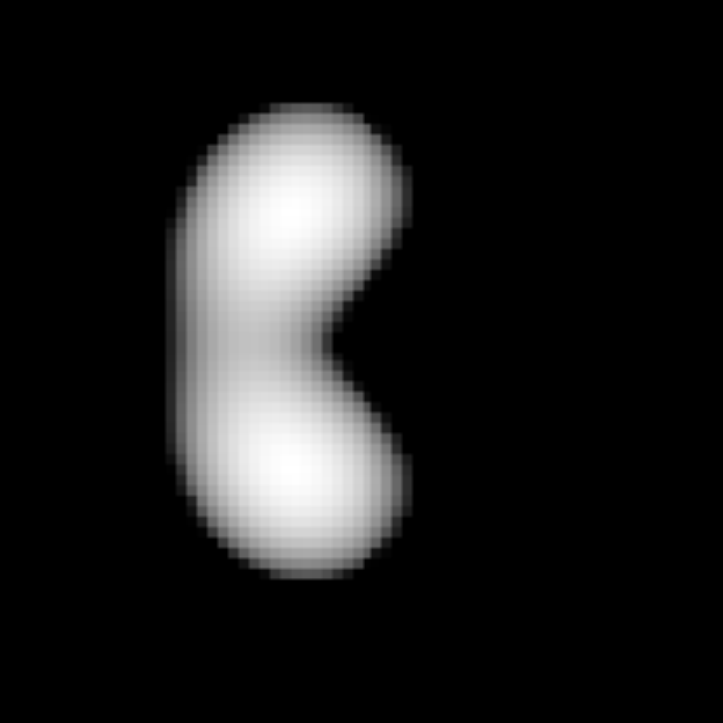}} \\
        \subfigure{
    \includegraphics[width=0.45\textwidth]{PhantomImage_70_11_1_4.pdf}}
	\figcaption{\small\bf\it Toroid phantom test case 1. (a)-(c): The moving image. (The fixed image is showed in Figure \ref{fig:Toroid3D_Iterative_Method} (a)-(c)).}
  \label{fig:Toroid3D_Simultaneous_Moving}
  \end{center}
  \end{figure}

\begin{figure}[!htb]
  \begin{center}
  \centering % \setlength{\floatsep}{10pt plus 3pt minus 2pt}
        \subfigure[]{
    \includegraphics[width=0.15\textwidth]{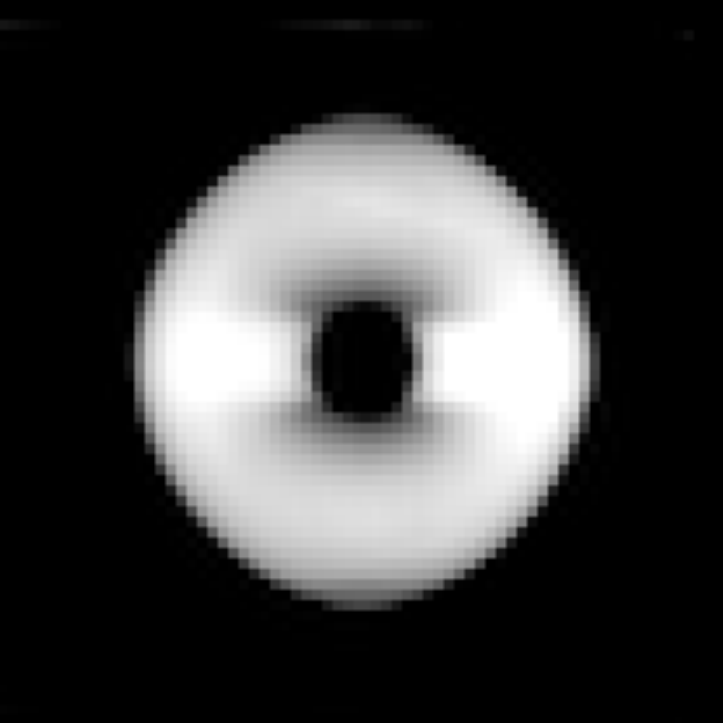}}
        \subfigure[]{
    \includegraphics[width=0.15\textwidth]{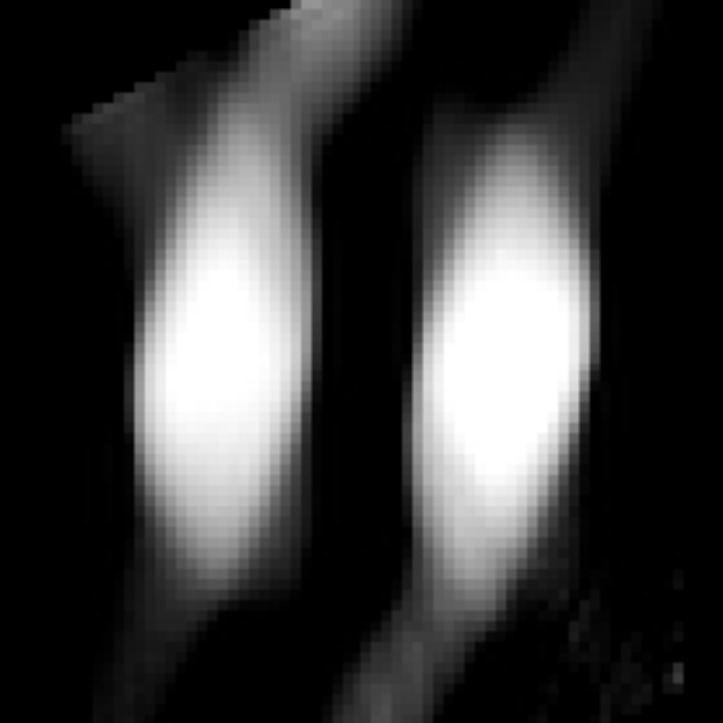}}
        \subfigure[]{
    \includegraphics[width=0.15\textwidth]{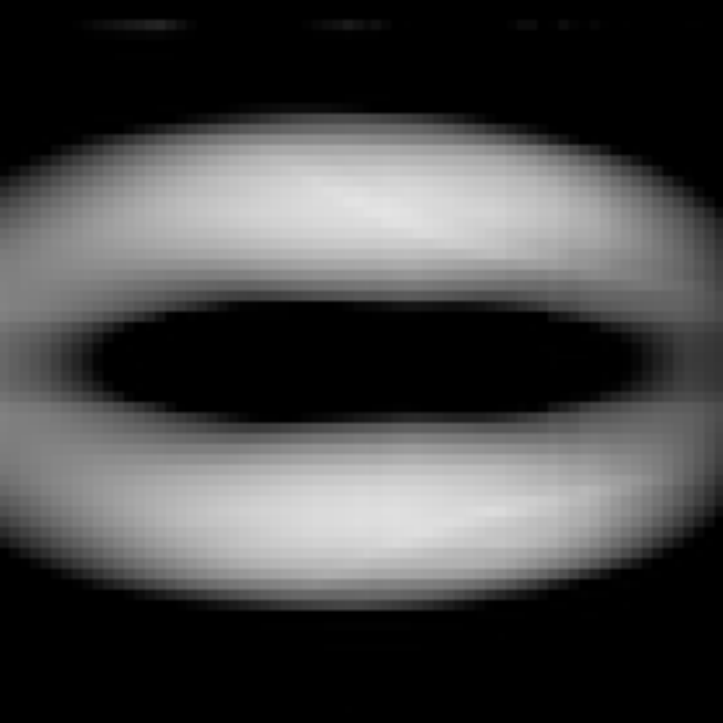}}  \\
        \subfigure[]{
    \includegraphics[width=0.15\textwidth]{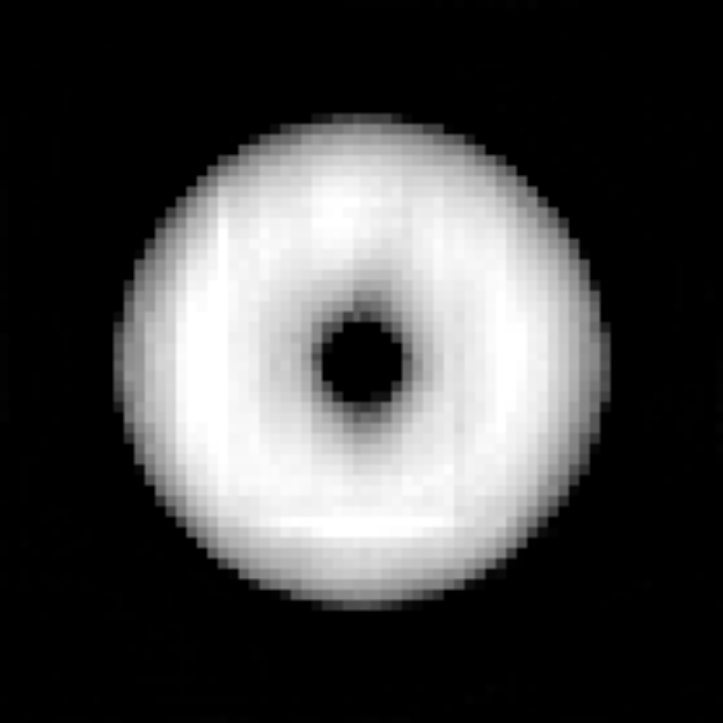}}
        \subfigure[]{
    \includegraphics[width=0.15\textwidth]{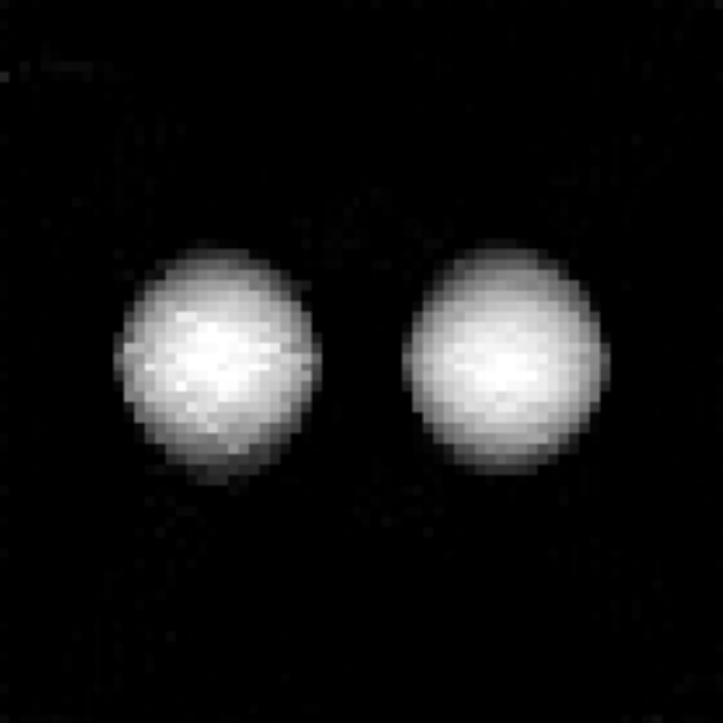}}
        \subfigure[]{
    \includegraphics[width=0.15\textwidth]{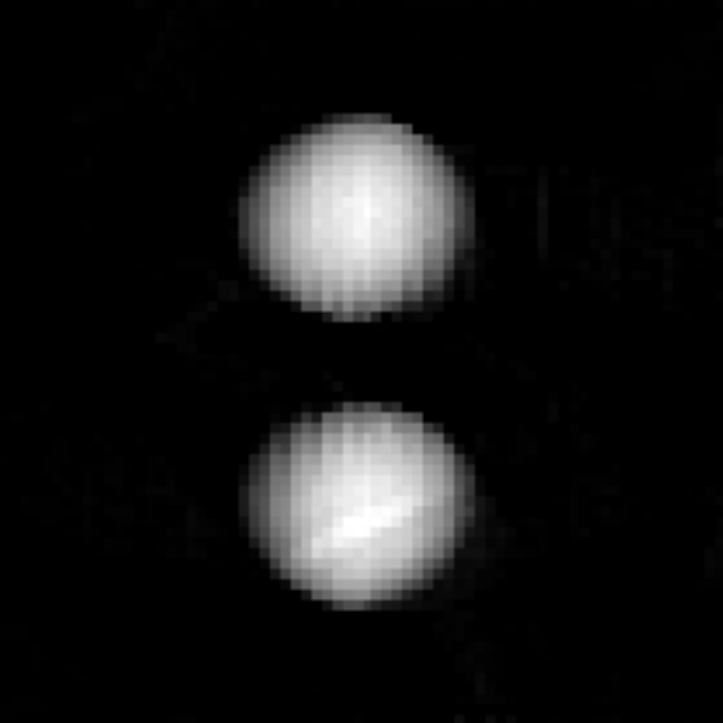}} \\
        \subfigure[]{
    \includegraphics[width=0.15\textwidth]{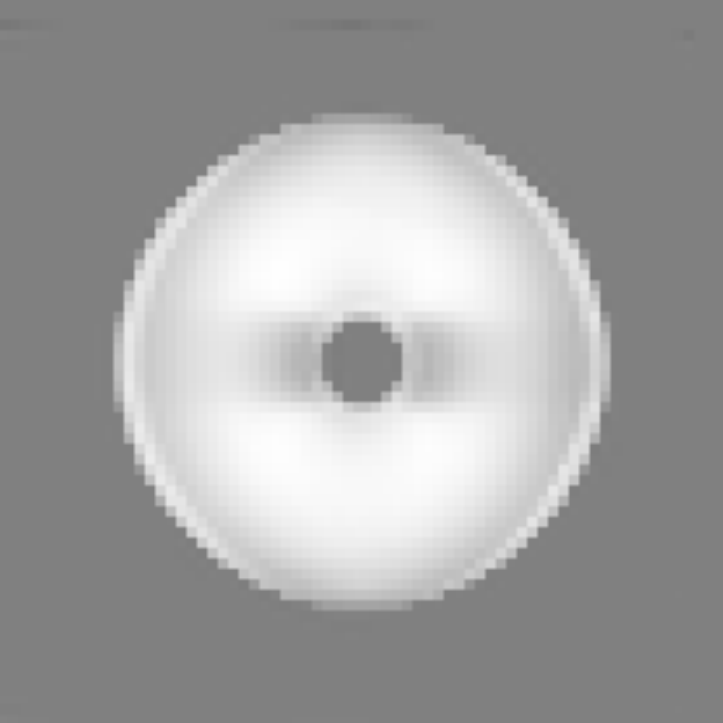}}
        \subfigure[]{
    \includegraphics[width=0.15\textwidth]{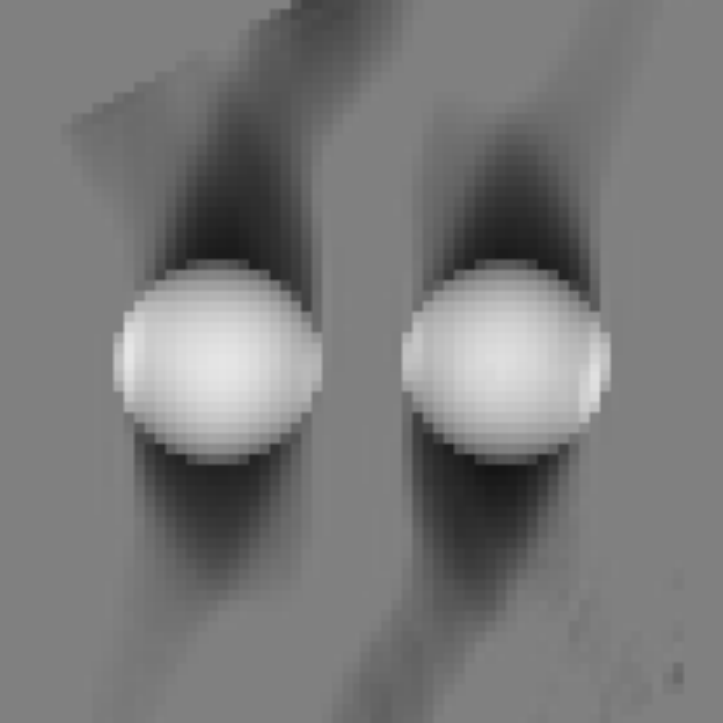}}
        \subfigure[]{
    \includegraphics[width=0.15\textwidth]{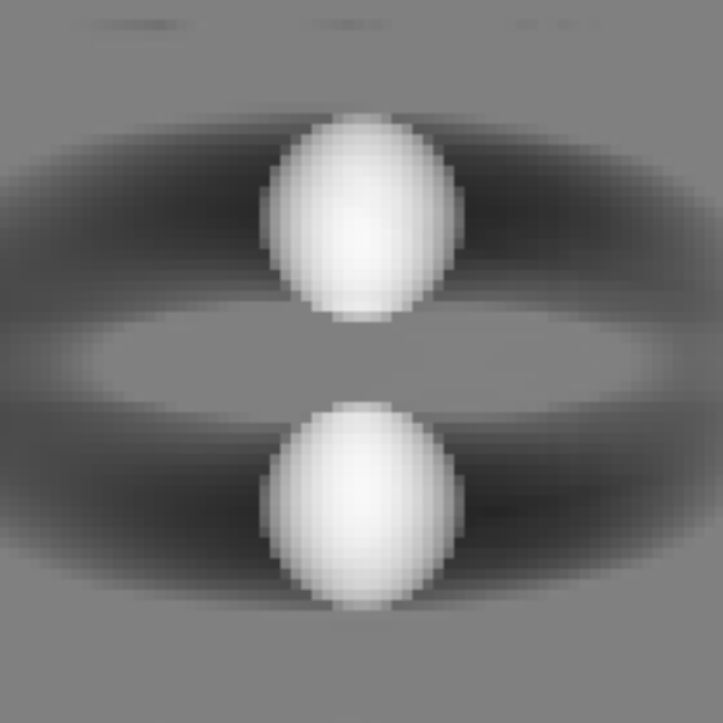}} \\
        \subfigure[]{
    \includegraphics[width=0.15\textwidth]{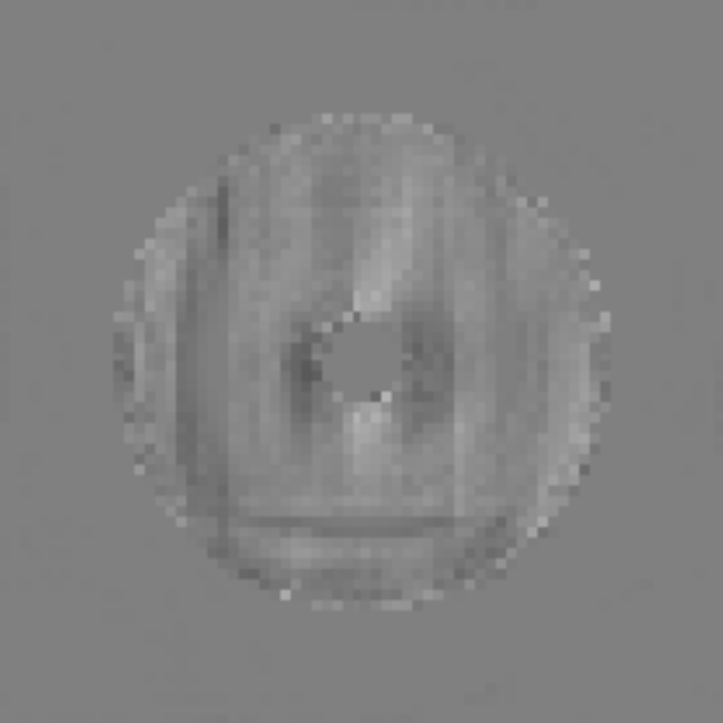}}
        \subfigure[]{
    \includegraphics[width=0.15\textwidth]{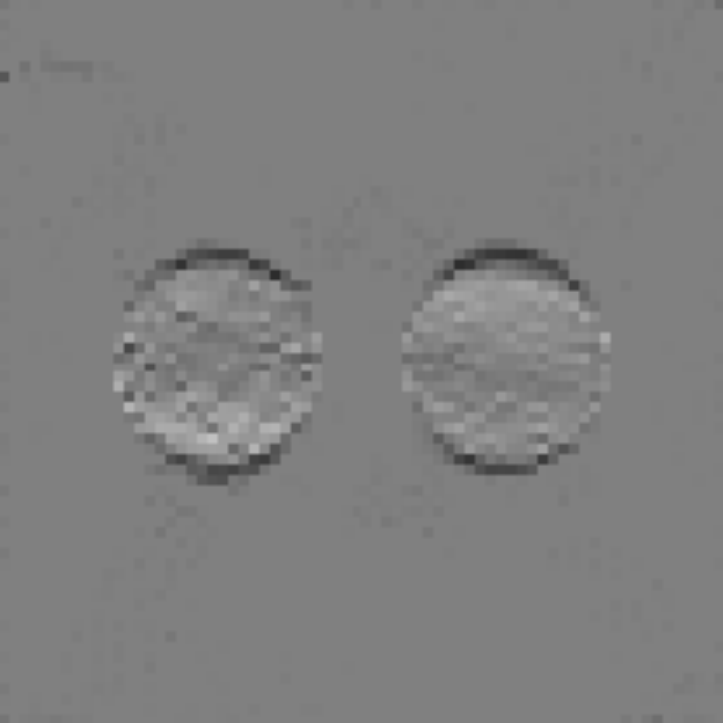}}
        \subfigure[]{
    \includegraphics[width=0.15\textwidth]{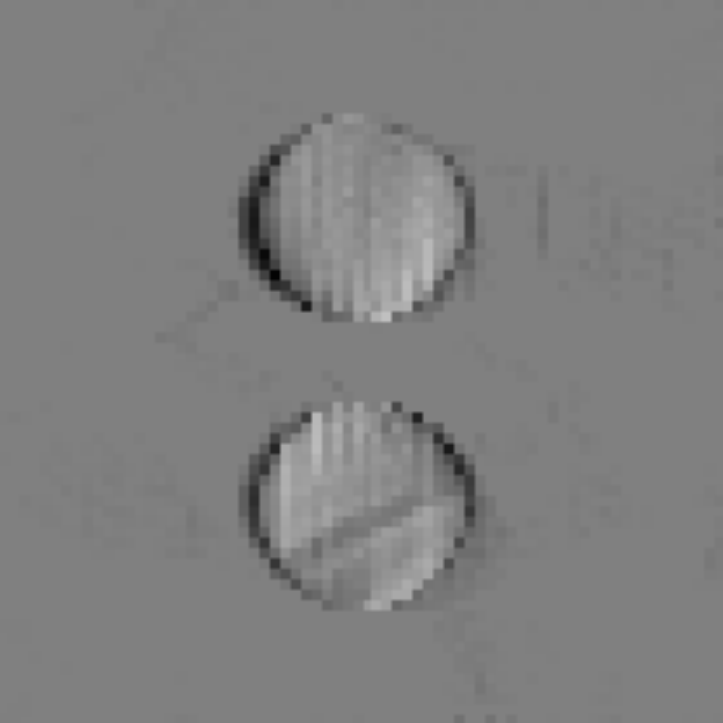}} \\
        \subfigure{
    \includegraphics[width=0.45\textwidth]{PhantomImage_70_11_1_4.pdf}}
	\figcaption{\small\bf\it Sequential results {\it vs.} simultaneous results. (a)-(c): Sequential result; (d)-(f): Simultaneous result; (g)-(i): Difference image between the sequential result and the fixed image (Figure \ref{fig:Toroid3D_Iterative_Method} (a)-(c)); (j)-(l): Difference image between the simultaneous result and the fixed image (Figure \ref{fig:Toroid3D_Iterative_Method} (a)-(c)).}
  \label{fig:Toroid3D_Simultaneous_Result}
  \end{center}
  \end{figure}

We performed $20$ different set of randomly simulated affine transformations to test the robustness of our simultaneous method. Affine test case 1 is presented here as an example (Figure \ref{fig:Toroid3D_Simultaneous_Moving} (a)-(c)). Results of the two different methods, {\it i.e.,} sequential {\it vs.} simultaneous, were compared. We found that there were fewer artefacts in the results of our simultaneous method compared to the sequential method (Figure \ref{fig:Toroid3D_Simultaneous_Result} (a)-(c) {\it vs.} (d)-(f)).

  \begin{figure}[!htb]
  \begin{center}
  \centering % \setlength{\floatsep}{10pt plus 3pt minus 2pt}
  \setlength{\abovecaptionskip}{0pt}
	\setlength{\belowcaptionskip}{0pt}
  \includegraphics[width=0.8\textwidth]{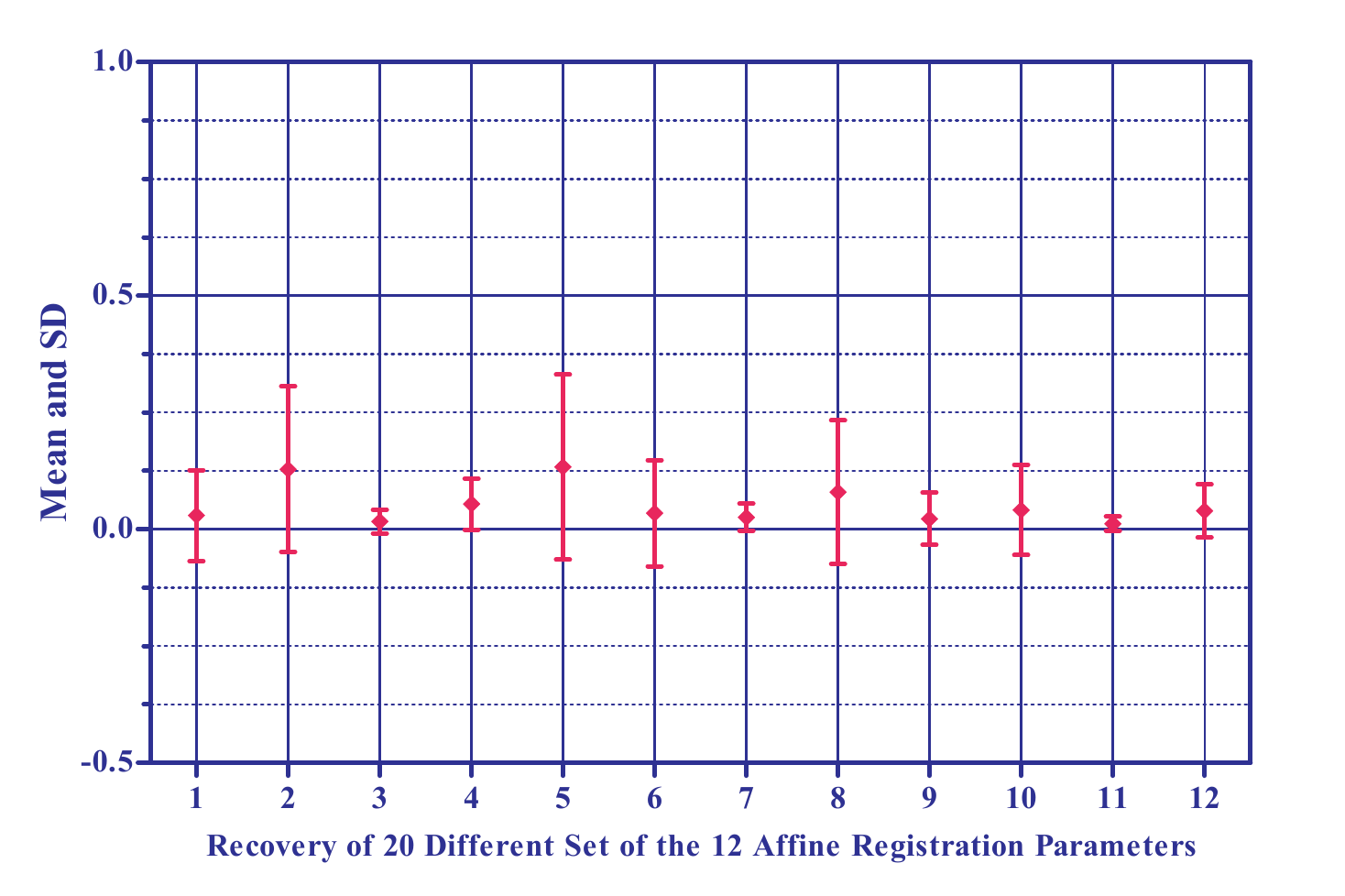}
  \figcaption{\small\bf\it Plot of the mean and standard deviation of the absolute error between the recovered and the ground truth transformation parameters for 20 different randomly generated affine transformations. The 12 parameters, which are unitless, are calculated combining rotation, scaling, shearing and translation. In other words, these 12 parameters are the 12 entries of the affine transformation matrix. In the plot, parameters number 4, 8, 12 are the translations along each direction, and other parameters are obtained from the multiplication of the rotation matrix and matrices of the scaling and shearing.}
  \label{fig:Toroid_70_20Sets}
  \end{center}
  \end{figure}

Similarly the difference images indicate that the simultaneous method is superior to the sequential method. The absolute errors between the recovered affine parameters and the ground truth of the transformations were also calculated. The results show that the recovery of the parameters were accurate and consistent for all the $20$ tests (Figure \ref{fig:Toroid_70_20Sets}).

\subsubsection{Test on a Uncompressed Breast MR Image} \label{sec:Uncompressed_Breast_MRI}

\begin{figure}[!htb]
  \begin{center}
  \centering % \setlength{\floatsep}{10pt plus 3pt minus 2pt}
        \subfigure[]{
    \includegraphics[width=0.18\textwidth]{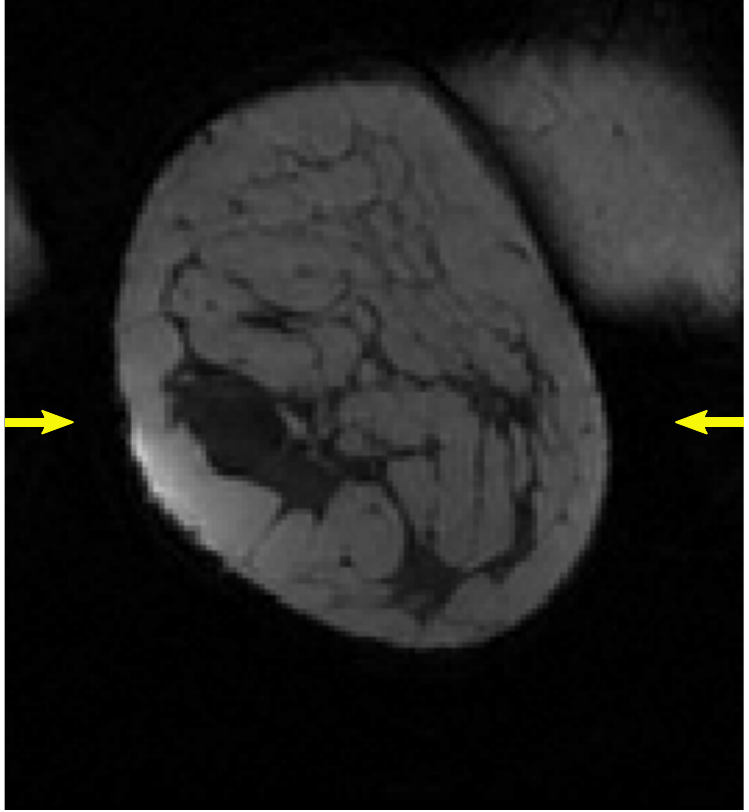}}
        \subfigure[]{
    \includegraphics[width=0.0938\textwidth]{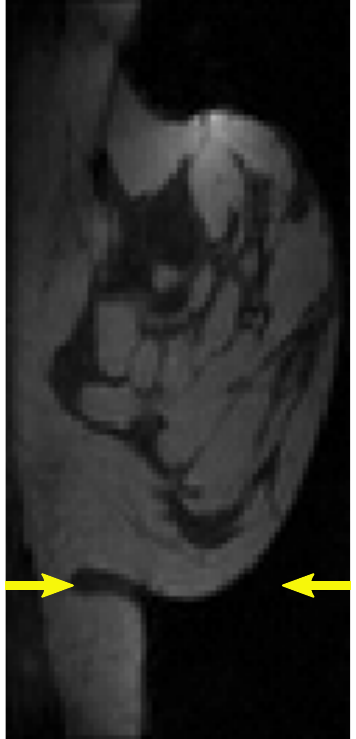}}
        \subfigure[]{
    \includegraphics[width=0.08218\textwidth]{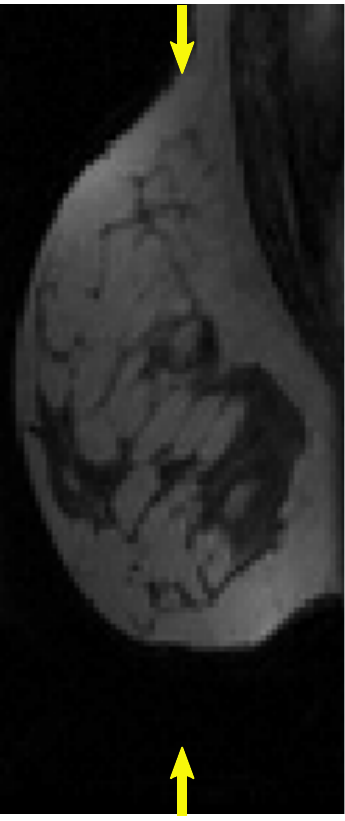}}  \\
        \subfigure[]{
    \includegraphics[width=0.18\textwidth]{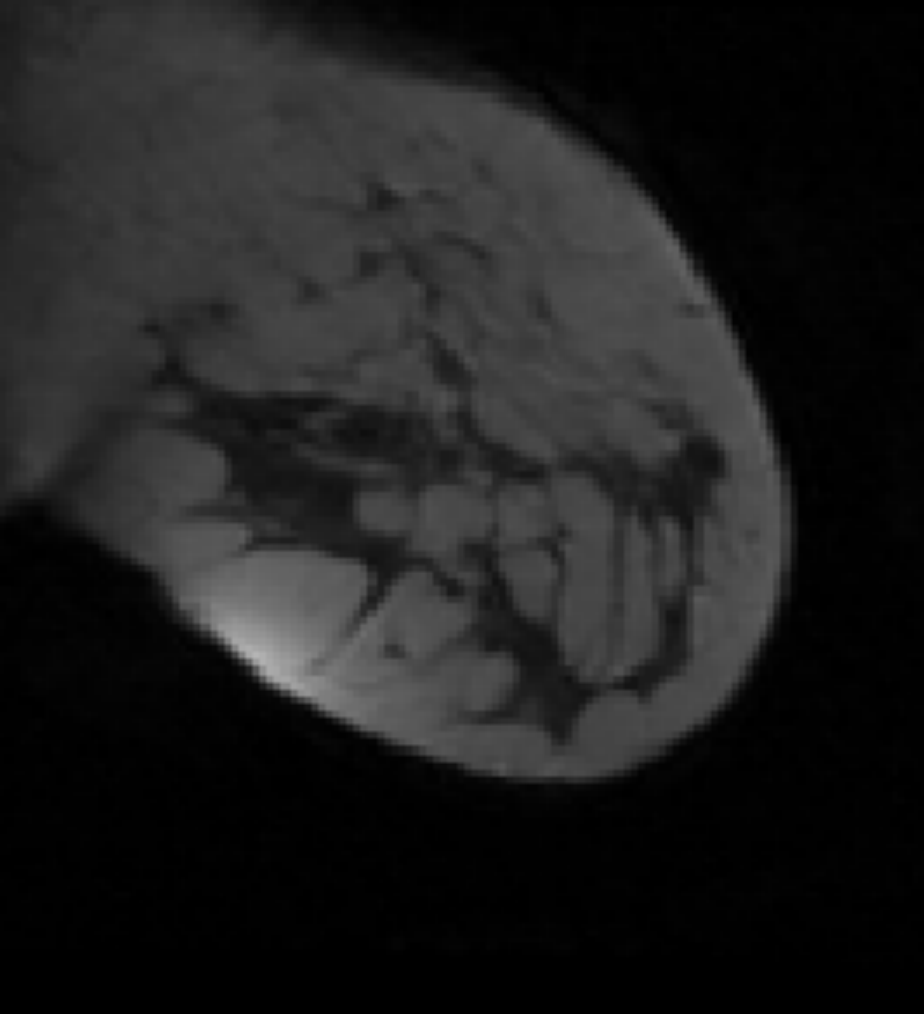}}
        \subfigure[]{
    \includegraphics[width=0.092\textwidth]{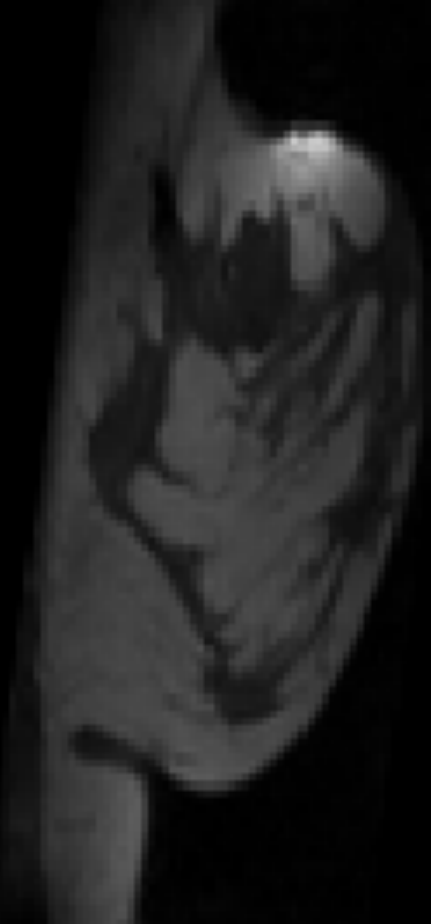}}
        \subfigure[]{
    \includegraphics[width=0.084\textwidth]{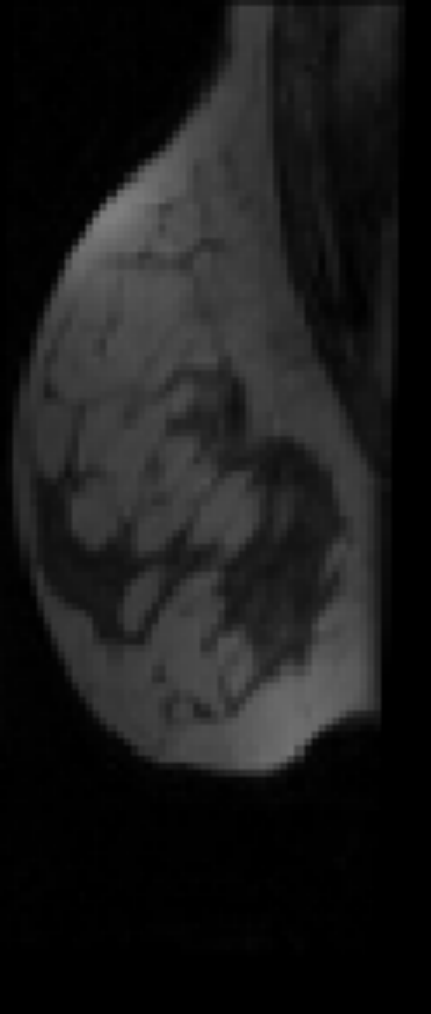}} \\
        \subfigure[]{
    \includegraphics[width=0.18\textwidth]{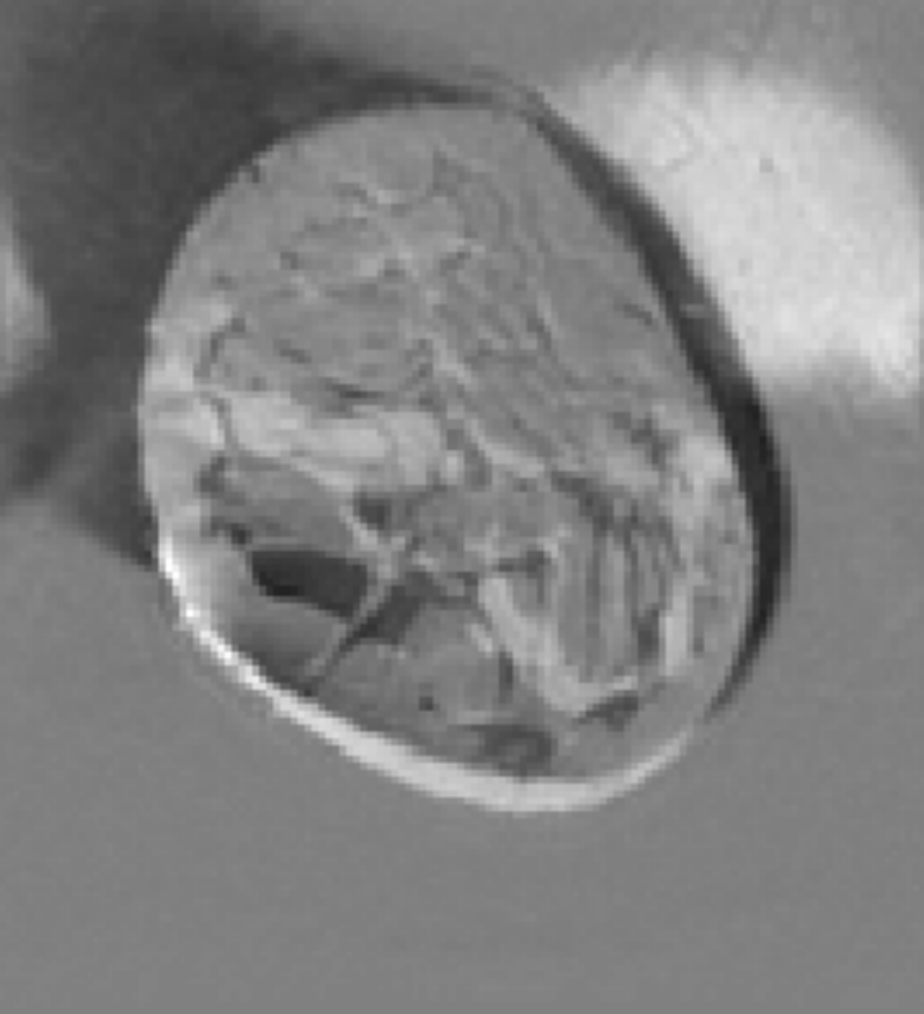}}
        \subfigure[]{
    \includegraphics[width=0.092\textwidth]{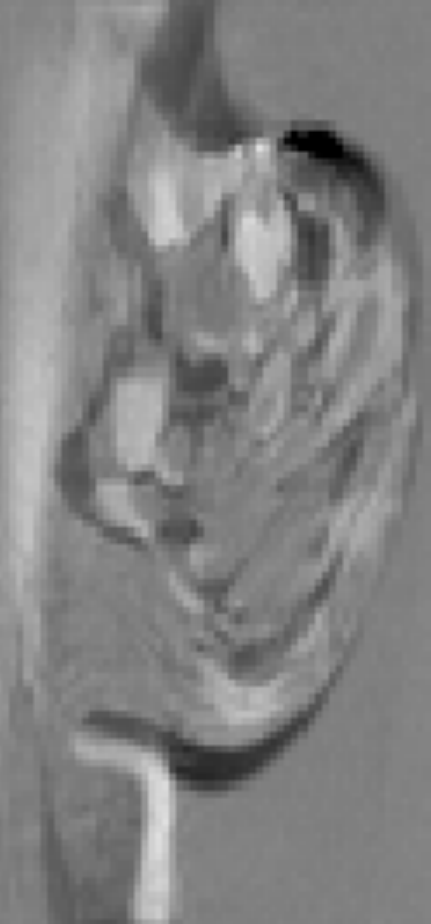}}
        \subfigure[]{
    \includegraphics[width=0.084\textwidth]{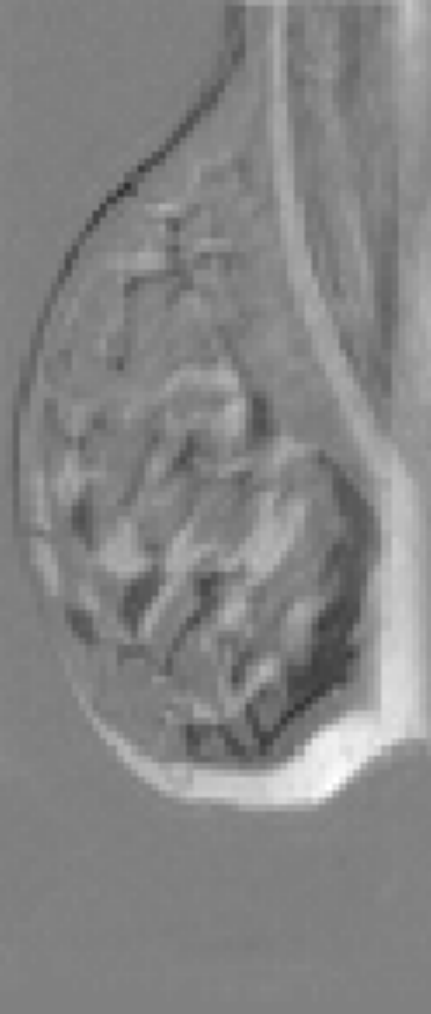}}  \\
        \subfigure{
    \includegraphics[width=0.45\textwidth]{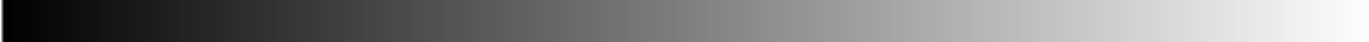}}
	\figcaption{\small\bf\it Breast MRI test case 8. (a)-(c): Fixed image; (d)-(f): Moving image; (g)-(i): Initial difference between the fixed and moving images. The transverse view has been rotated $90^{\mathrm{o}}$ clockwise for the purpose of better display. (Left: Coronal view; Middle: Transverse view; Right: Sagittal view.)}
  \label{fig:P1_128_8}
  \end{center}
  \end{figure}

\begin{figure}[!htb]
  \begin{center}
  \centering % \setlength{\floatsep}{10pt plus 3pt minus 2pt}
        \subfigure[]{
    \includegraphics[width=0.18\textwidth]{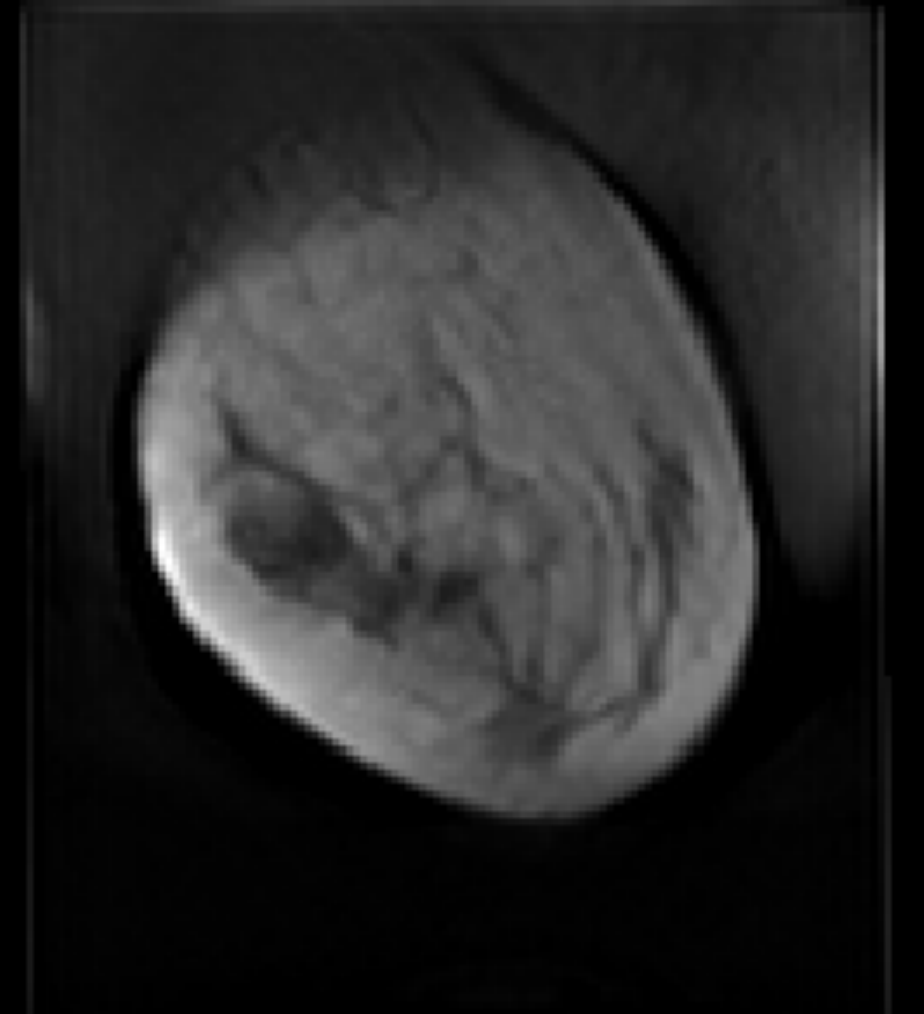}}
        \subfigure[]{
    \includegraphics[width=0.092\textwidth]{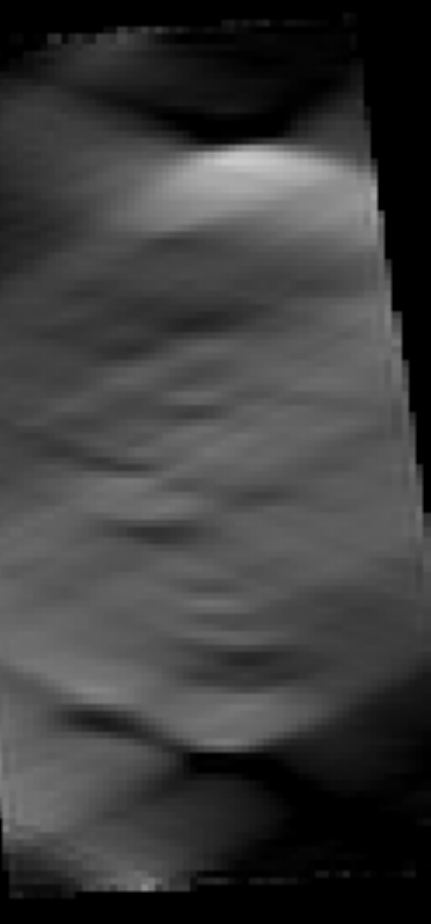}}
        \subfigure[]{
    \includegraphics[width=0.084\textwidth]{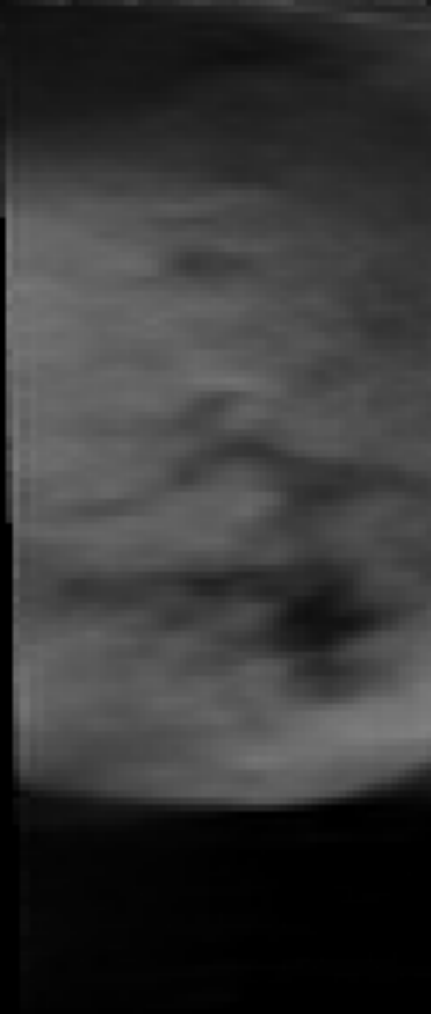}}  \\
        \subfigure[]{
    \includegraphics[width=0.18\textwidth]{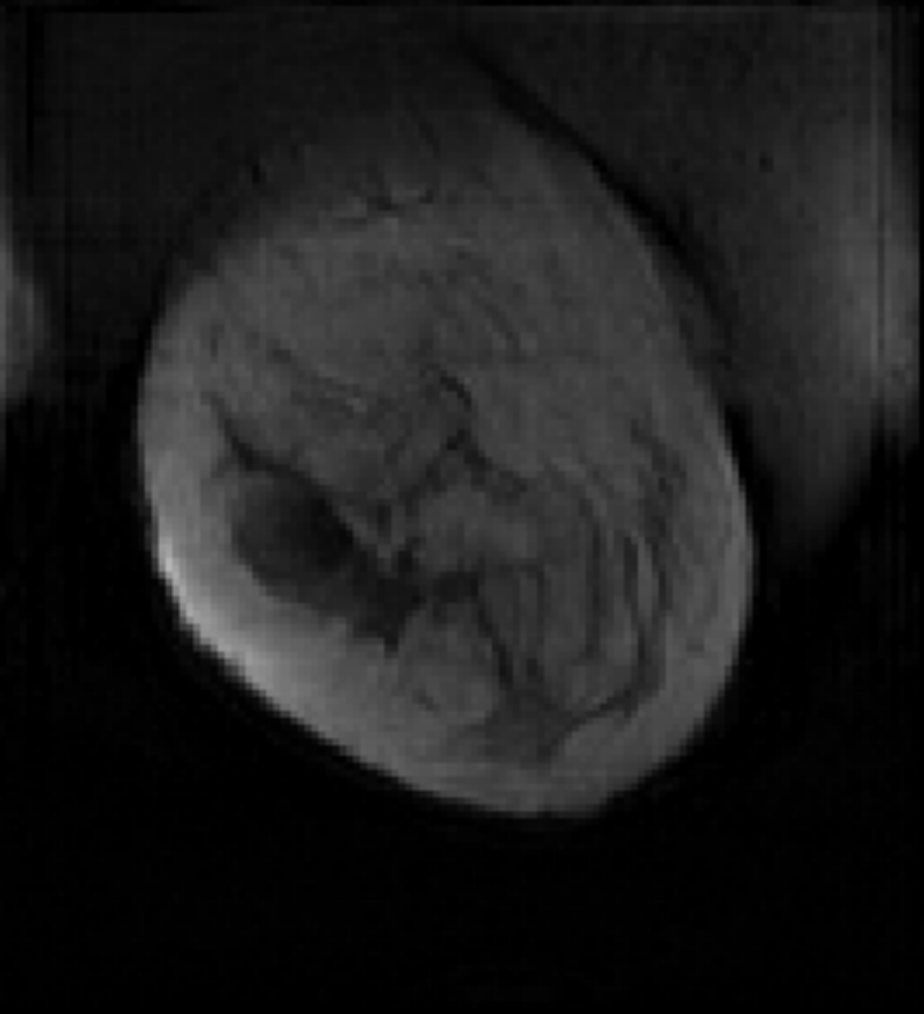}}
        \subfigure[]{
    \includegraphics[width=0.092\textwidth]{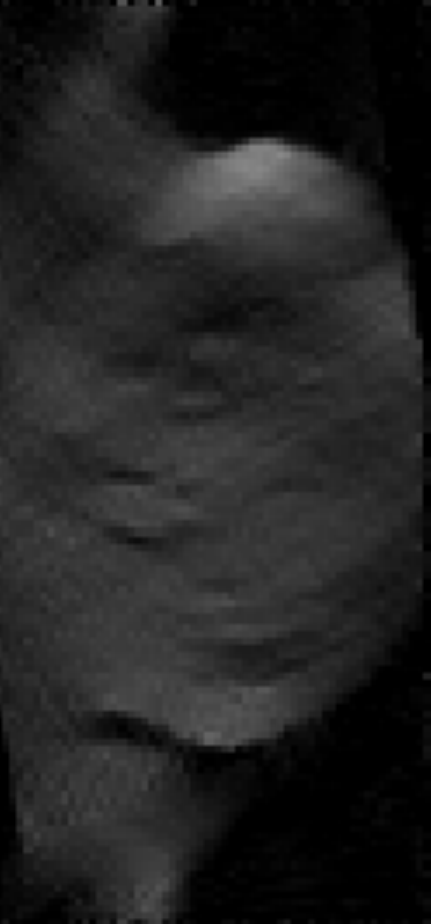}}
        \subfigure[]{
    \includegraphics[width=0.084\textwidth]{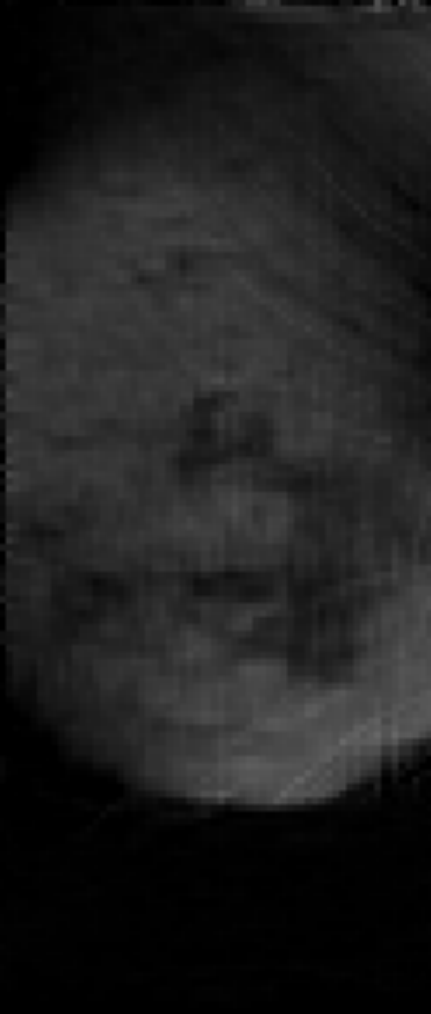}} \\
        \subfigure[]{
    \includegraphics[width=0.18\textwidth]{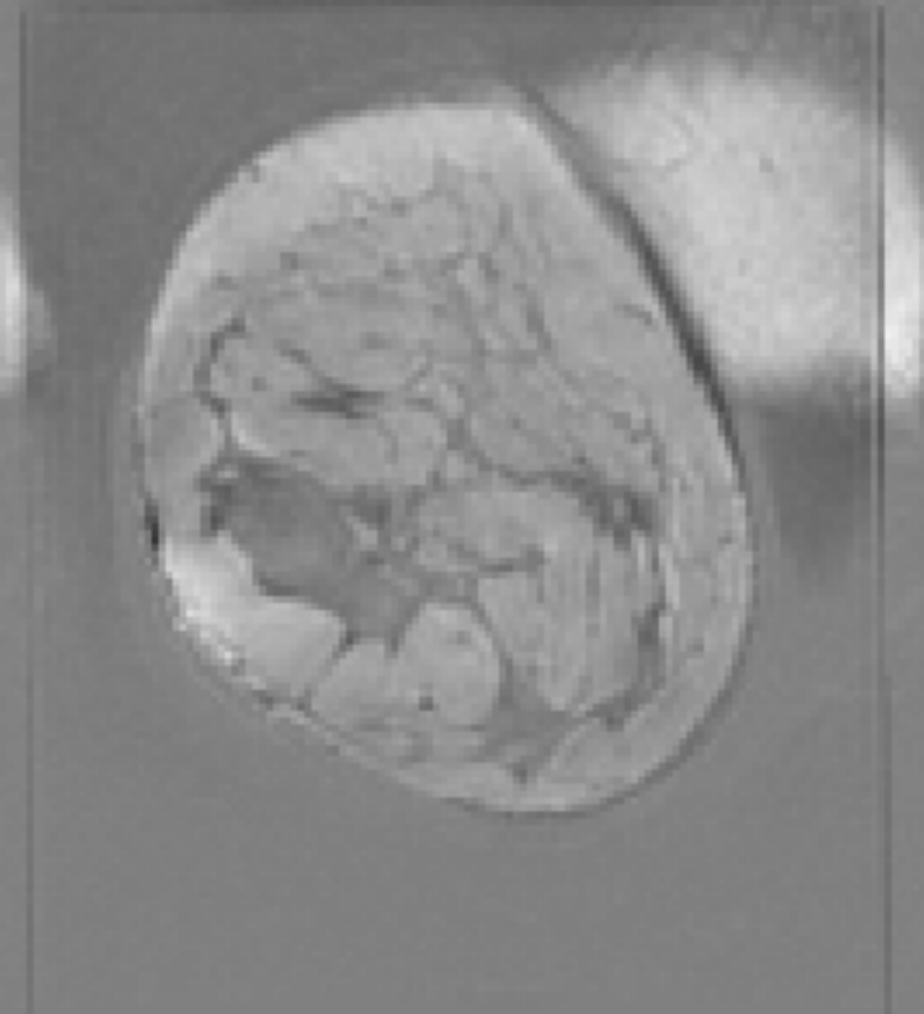}}
        \subfigure[]{
    \includegraphics[width=0.092\textwidth]{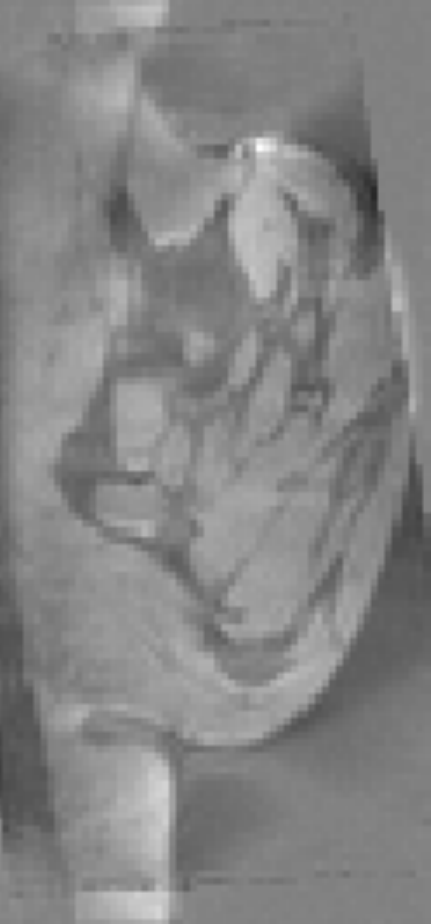}}
        \subfigure[]{
    \includegraphics[width=0.084\textwidth]{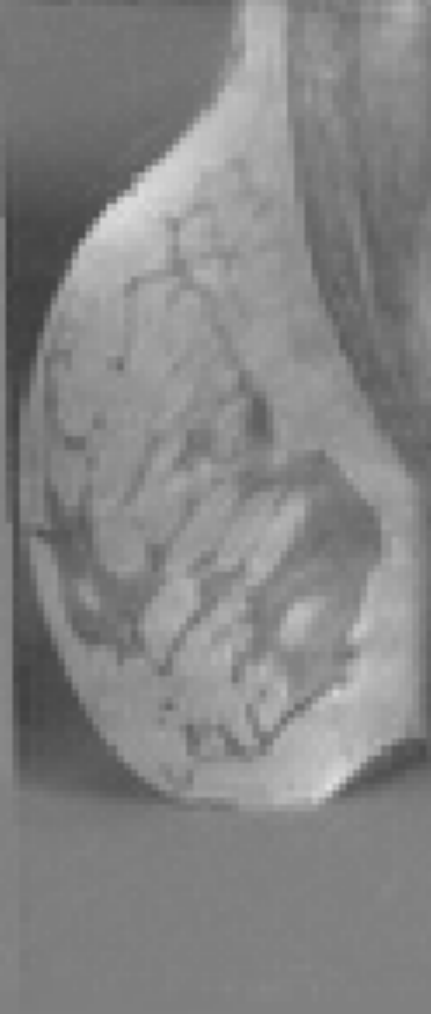}}  \\
        \subfigure[]{
    \includegraphics[width=0.18\textwidth]{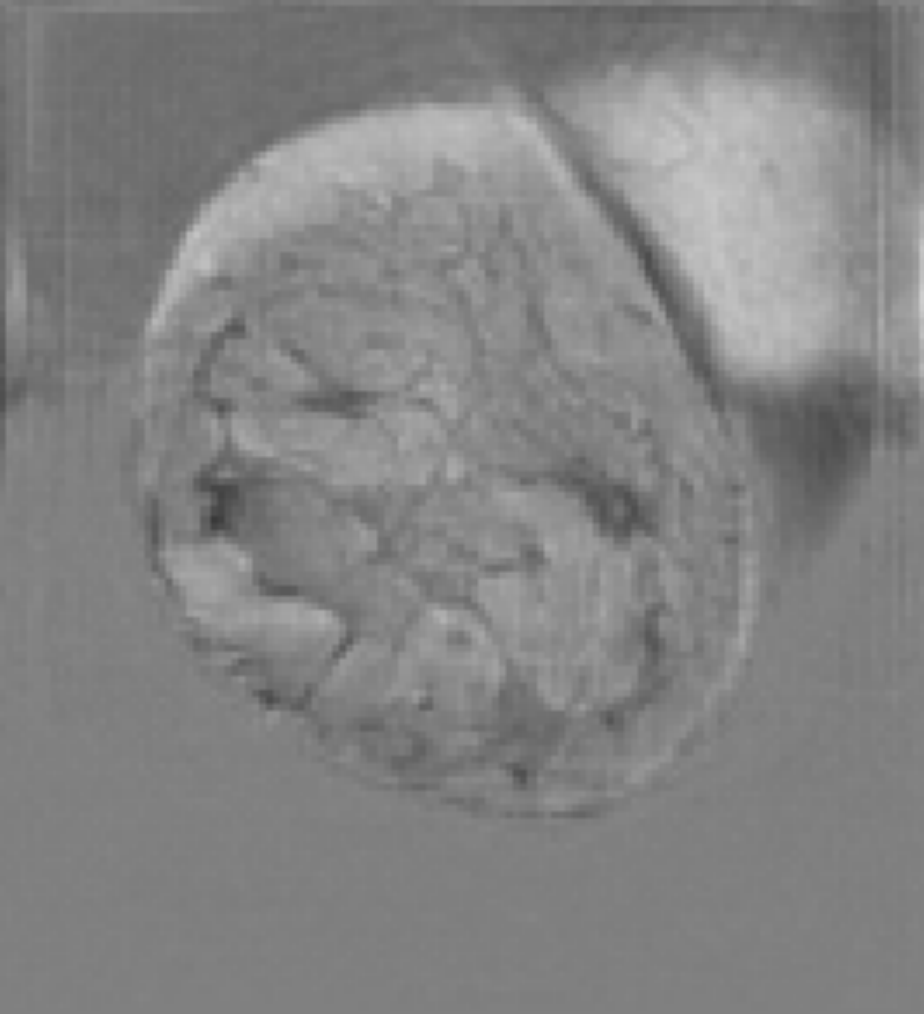}}
        \subfigure[]{
    \includegraphics[width=0.092\textwidth]{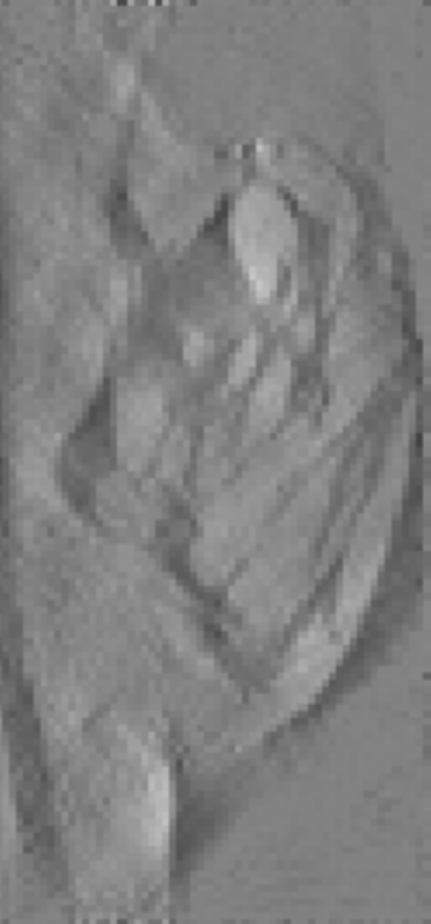}}
        \subfigure[]{
    \includegraphics[width=0.084\textwidth]{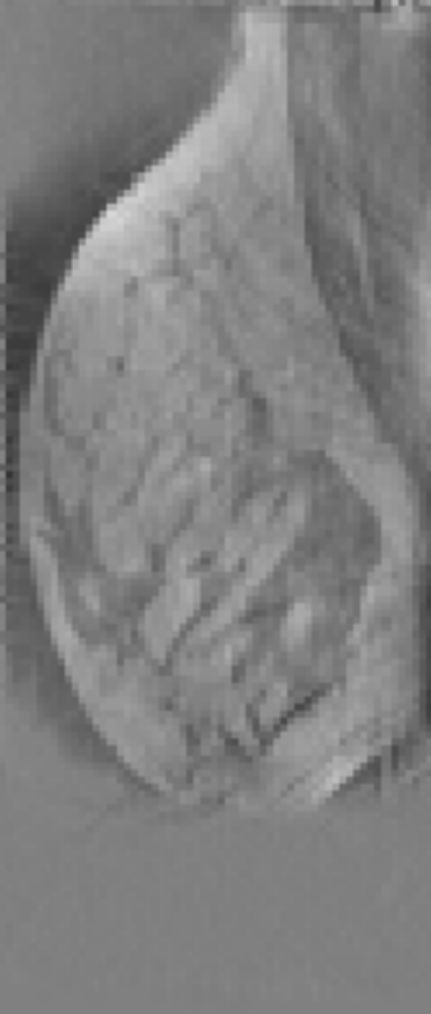}} \\
        \subfigure{
    \includegraphics[width=0.45\textwidth]{PhantomImage_128_11_8_4.pdf}}
	\figcaption{\small\bf\it Sequential results {\it vs.} simultaneous results. (a)-(c): Sequential result; (d)-(f): Simultaneous result; (g)-(i): Differences between the sequential result and the fixed image; (j)-(l): Differences between the simultaneous result and the fixed image.}
  \label{fig:P1_128_8_Results}
  \end{center}
  \end{figure}

\begin{figure}[!htb]
  \begin{center}
  \centering % \setlength{\floatsep}{10pt plus 3pt minus 2pt}
  \includegraphics[width=0.8\textwidth]{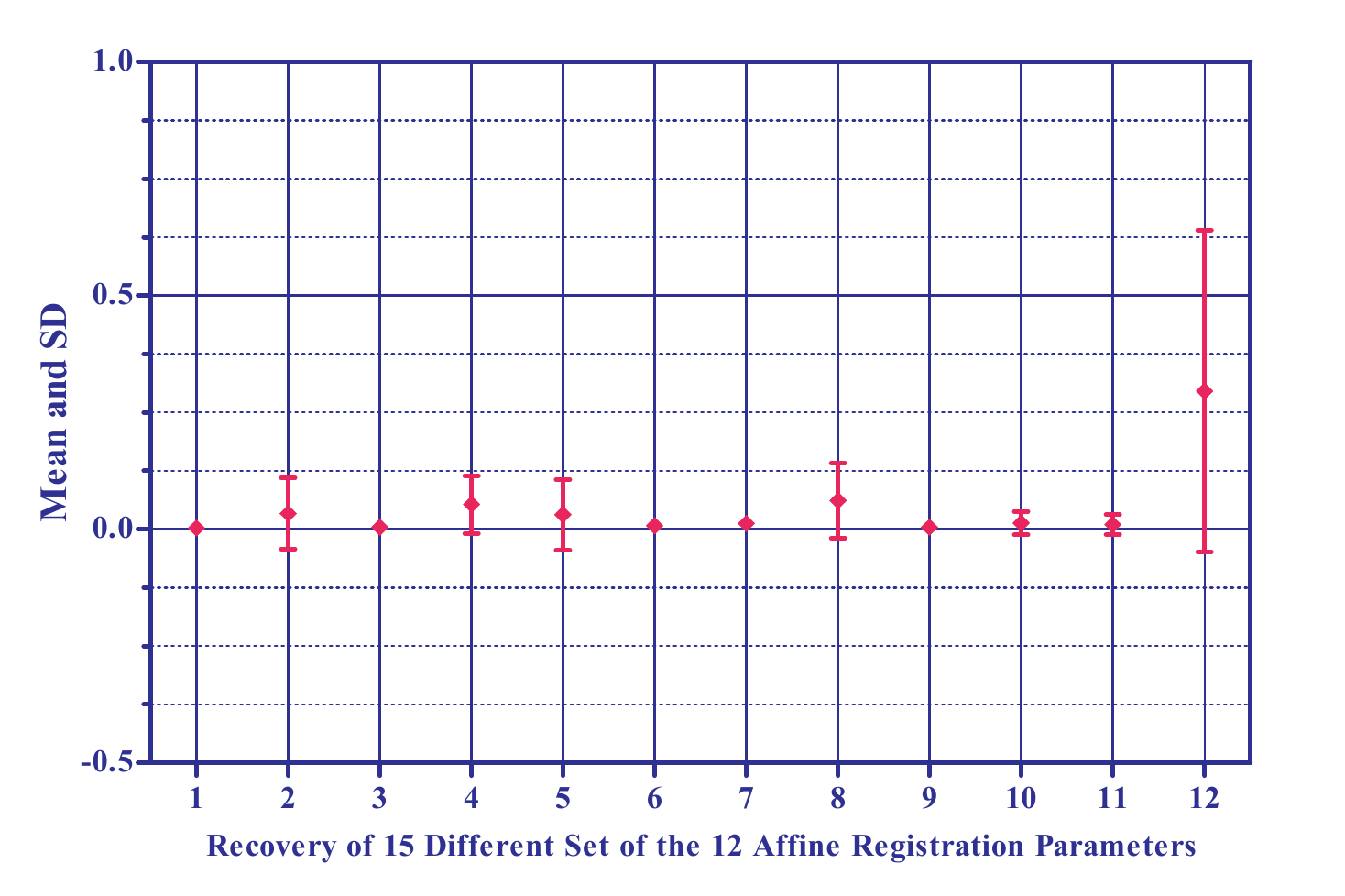}
  \figcaption{\small\bf\it Mean and standard deviation of the absolute error between the recovered and the ground truth of 15 different affine transformations.}
  \label{fig:P1_128_15Sets}
  \end{center}
  \end{figure}

The results of the experiment on the uncompressed breast MR image suggest that our simultaneous method has clear advantages over the sequential method. One $128\times140\times60$mm$^3$ breast MR image with $0.48\times0.48\times0.48$mm$^3$ resolution (Figure \ref{fig:P1_128_8} (a)-(c)) was used for all $15$ tests (Test case 8 in Figure \ref{fig:P1_128_8} (d)-(f) and the initial difference image in Figure \ref{fig:P1_128_8} (g)-(i)). Less artefacts were found in the results of our simultaneous method than the results of the sequential method (Figure \ref{fig:P1_128_8_Results} (a)-(c) {\it vs.} (d)-(f) and difference images in Figure \ref{fig:P1_128_8_Results} (g)-(i) {\it vs.} (j)-(l)). Figure \ref{fig:P1_128_15Sets} shows the recovery of the transformation parameters.

% Results of the experiment 2

\begin{figure}[!htb]
  \begin{center}
  \centering % \setlength{\floatsep}{10pt plus 3pt minus 2pt}
        \subfigure{
    \includegraphics[width=0.8\textwidth]{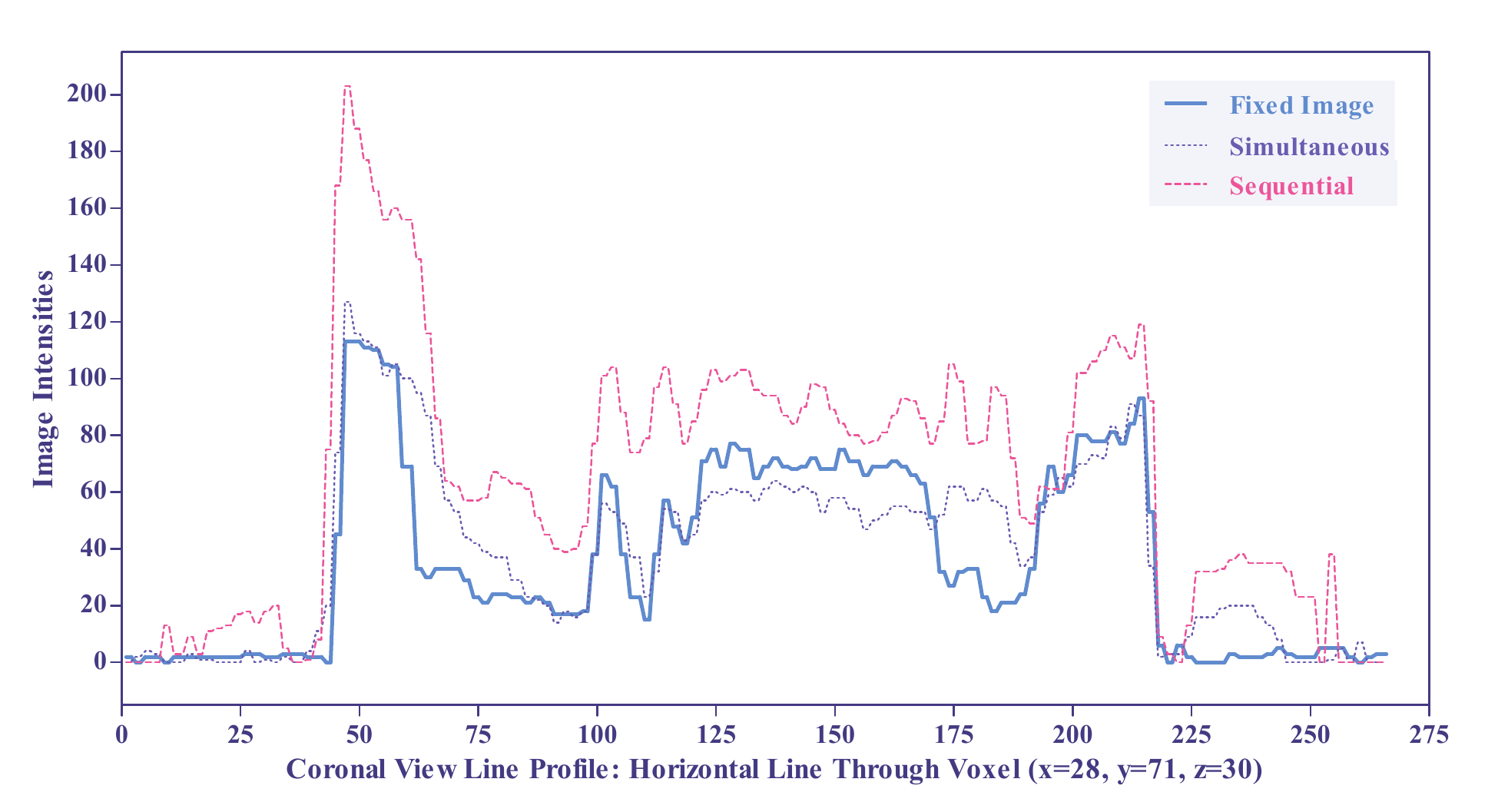}}  \\
        \subfigure{
    \includegraphics[width=0.8\textwidth]{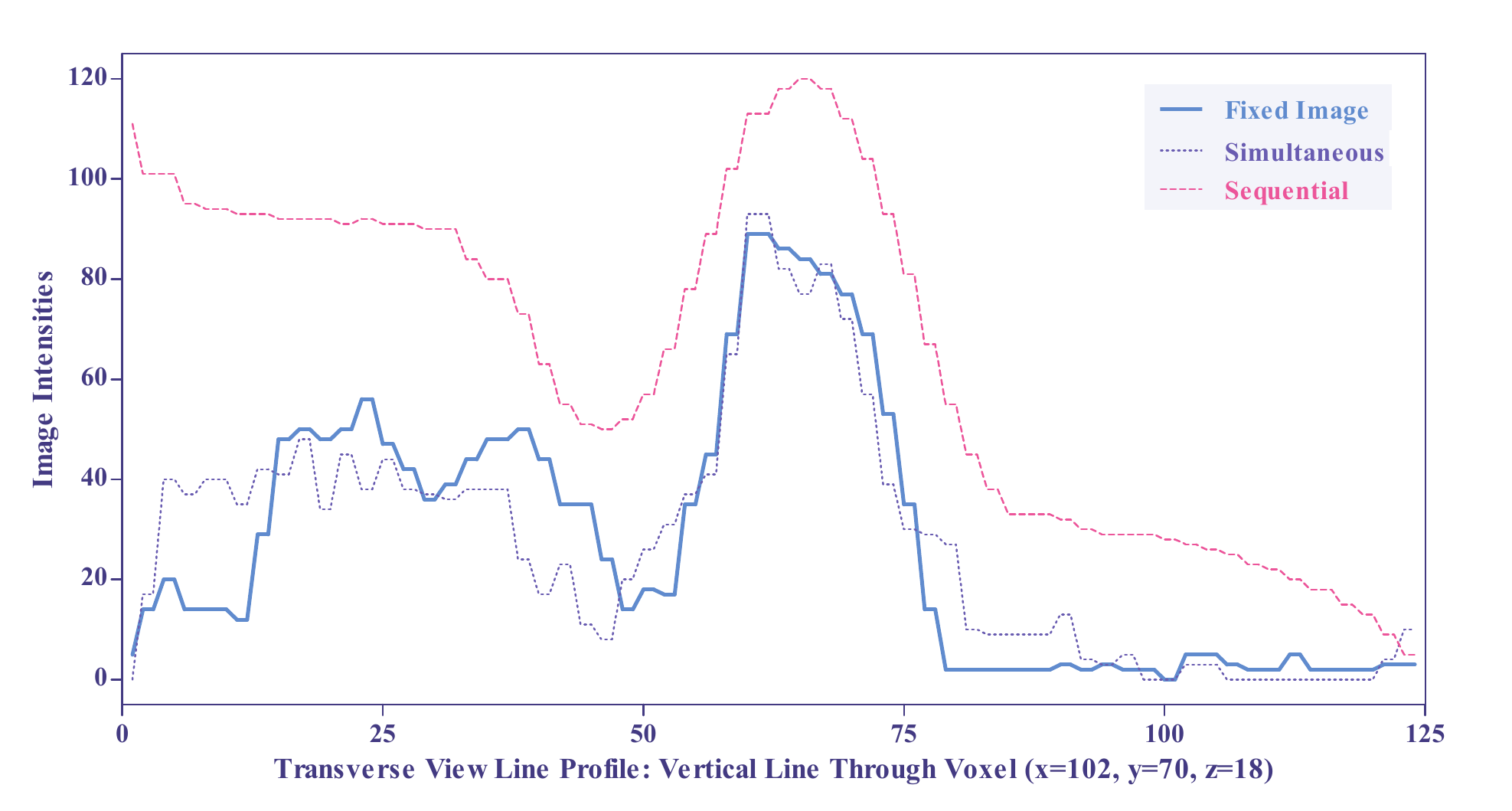}}  \\
        \subfigure{
    \includegraphics[width=0.8\textwidth]{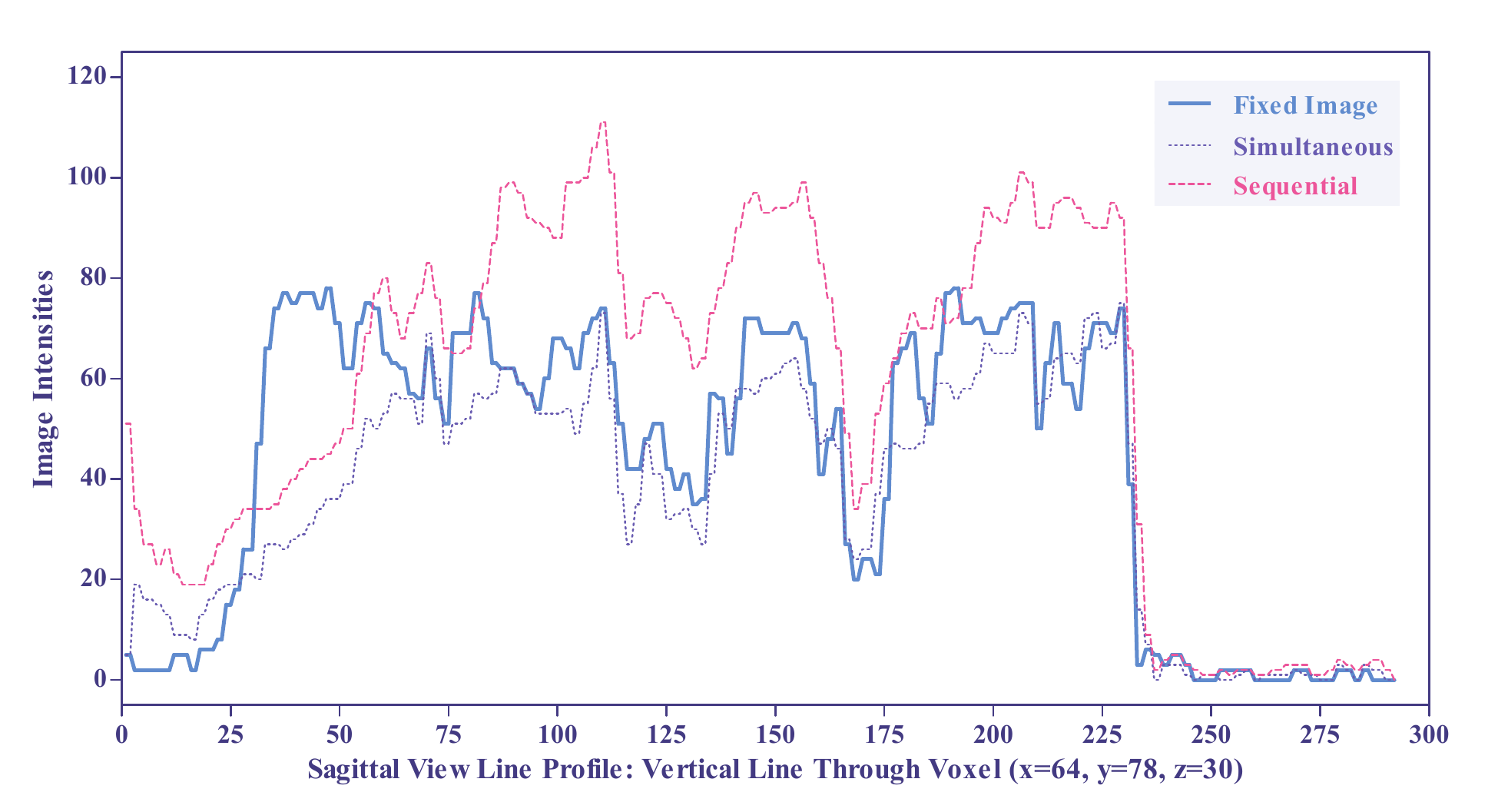}}
	\figcaption{\small\bf\it Line profiles of the three views of the breast MRI test case 8. (The line profiles were drawn between the two arrows of each view as seen in Figure \ref{fig:P1_128_8} (a)-(c) as an example, and they were at the same positions for other corresponding images.)}
  \label{fig:MRILineProfile}
  \end{center}
  \end{figure}

Cross-sectional line profiles in each view were plotted (Figure \ref{fig:MRILineProfile}). By comparing the reconstruction and registration results to the original fixed image, we found that our simultaneous method produced a more accurate intensities estimation than the sequential method.

\subsubsection{Test on a Simulated DBT}

\begin{figure}[!htb]
  \begin{center}
  \centering % \setlength{\floatsep}{10pt plus 3pt minus 2pt}
        \subfigure[]{
    \includegraphics[width=0.238\textwidth]{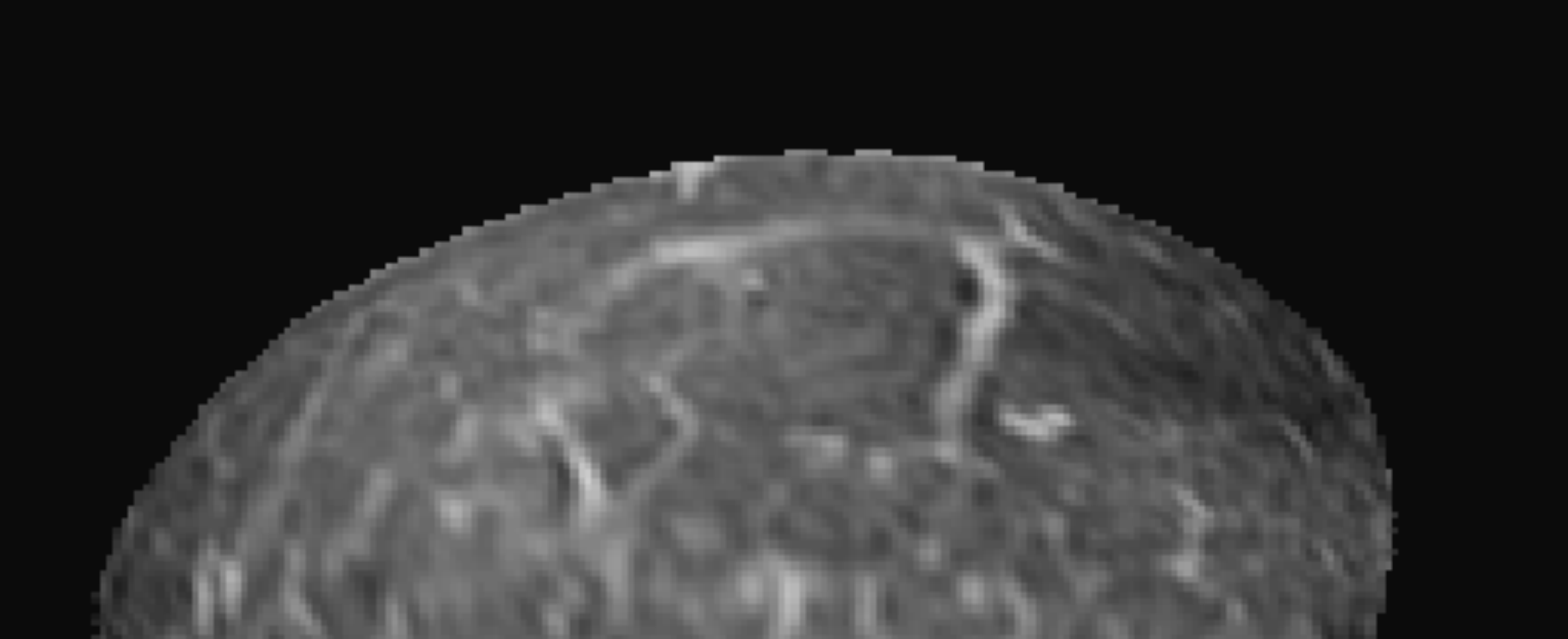}}
        \subfigure[]{
    \includegraphics[width=0.118\textwidth]{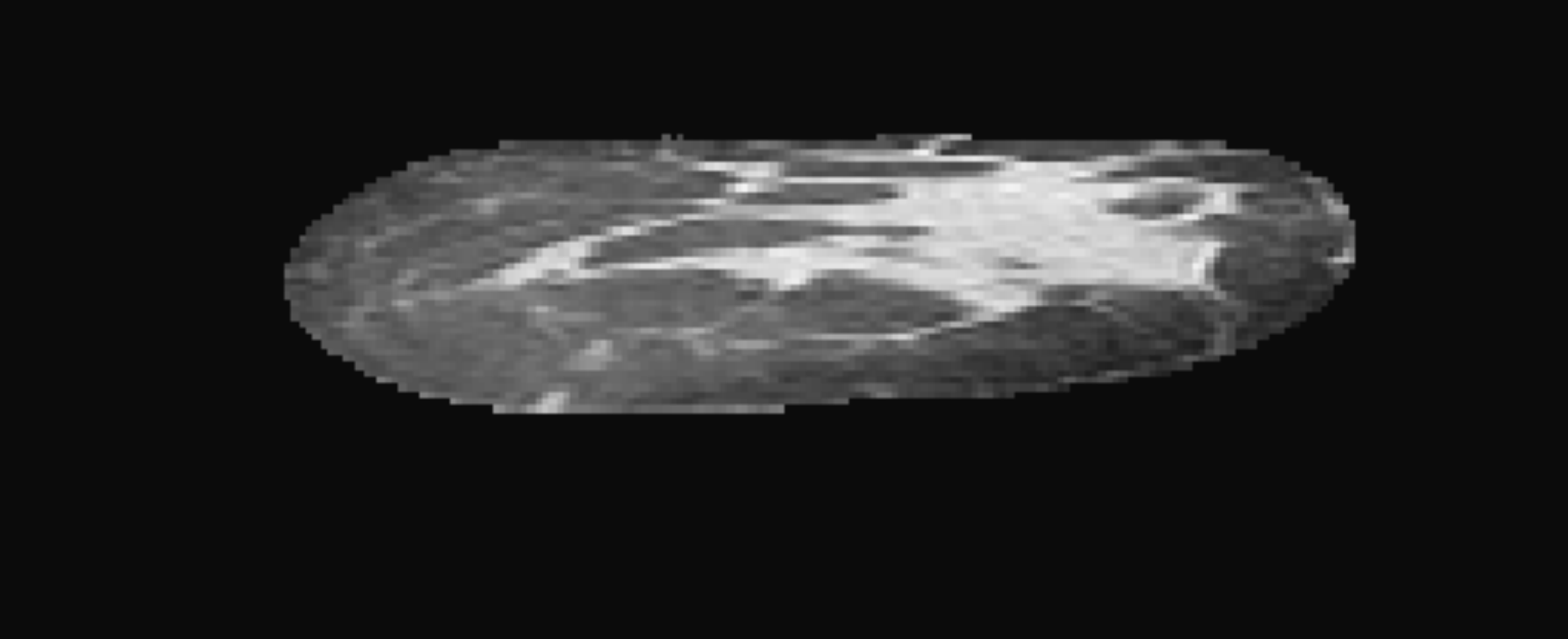}}
        \subfigure[]{
    \includegraphics[width=0.096\textwidth]{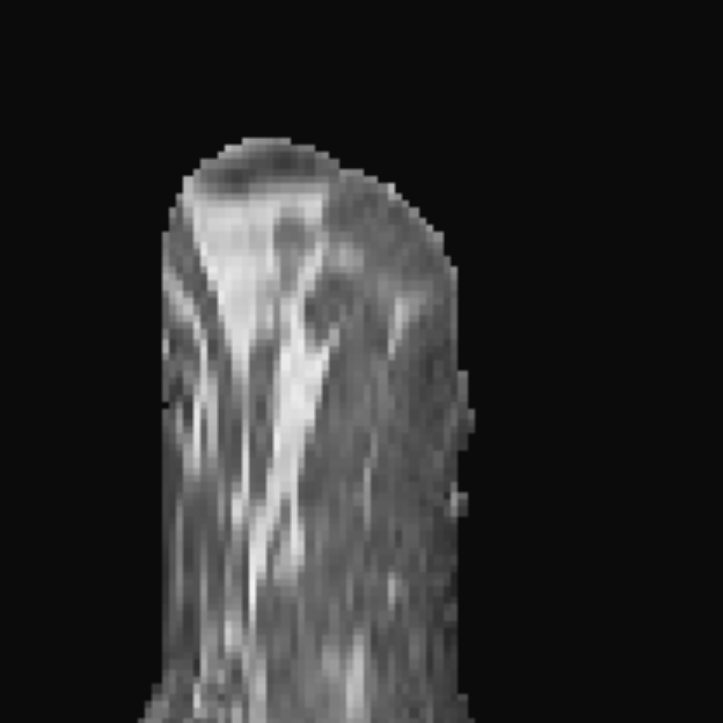}}  \\
        \subfigure[]{
    \includegraphics[width=0.238\textwidth]{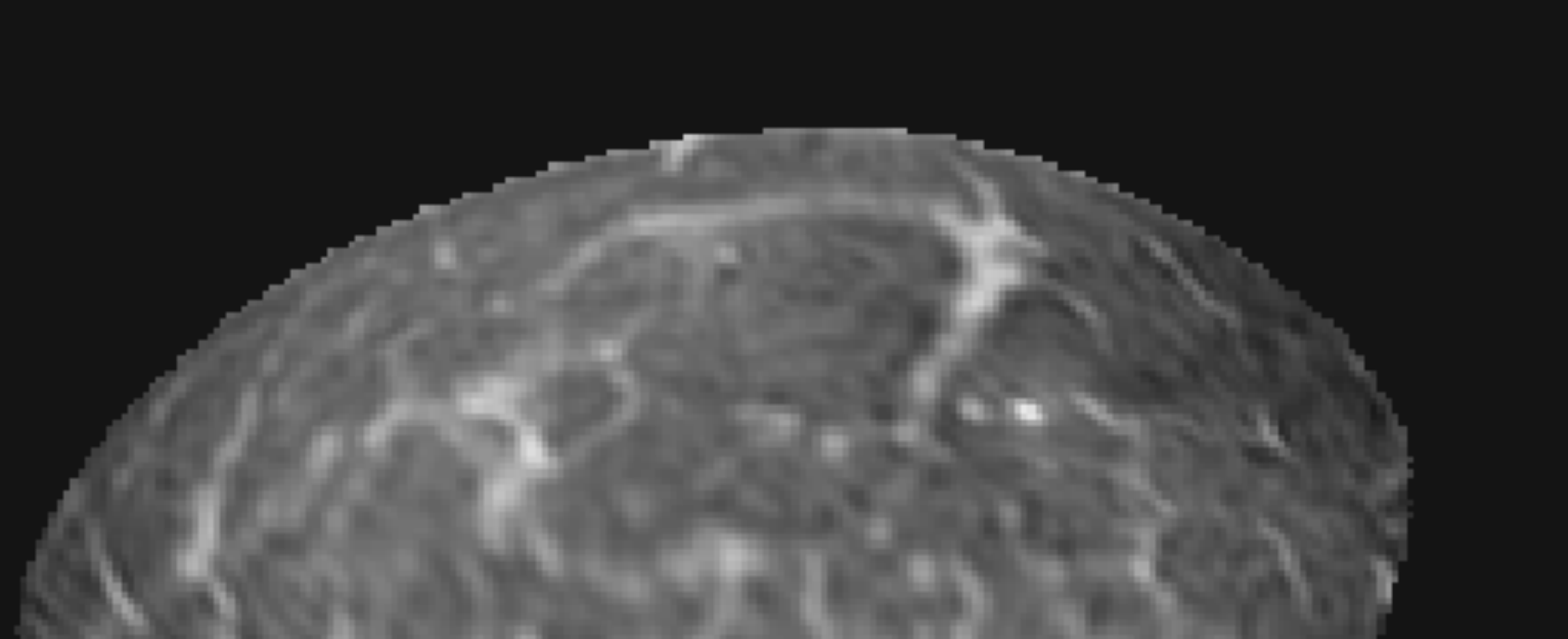}}
        \subfigure[]{
    \includegraphics[width=0.118\textwidth]{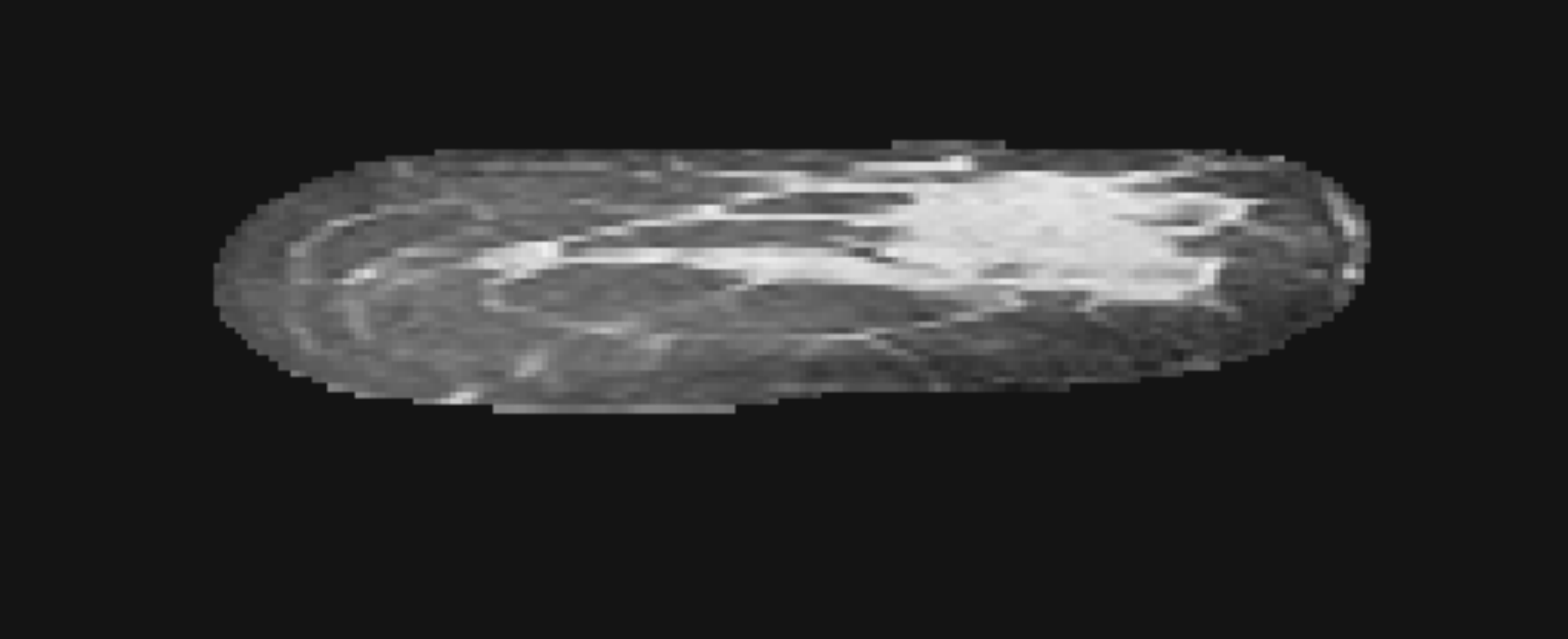}}
        \subfigure[]{
    \includegraphics[width=0.096\textwidth]{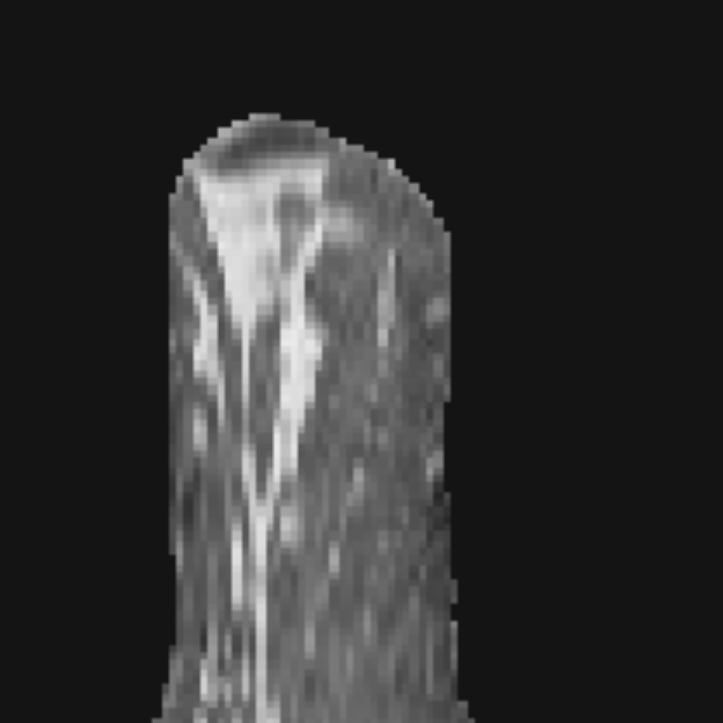}} \\
        \subfigure[]{
    \includegraphics[width=0.238\textwidth]{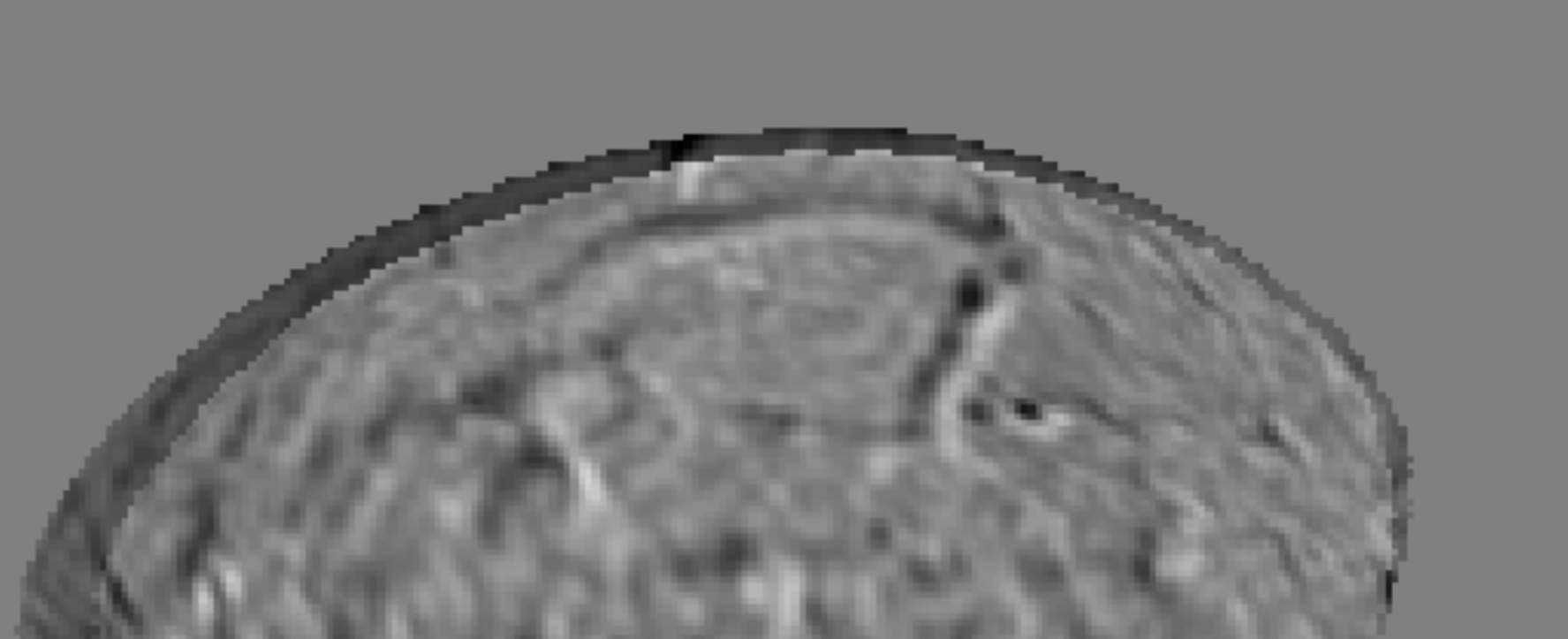}}
        \subfigure[]{
    \includegraphics[width=0.118\textwidth]{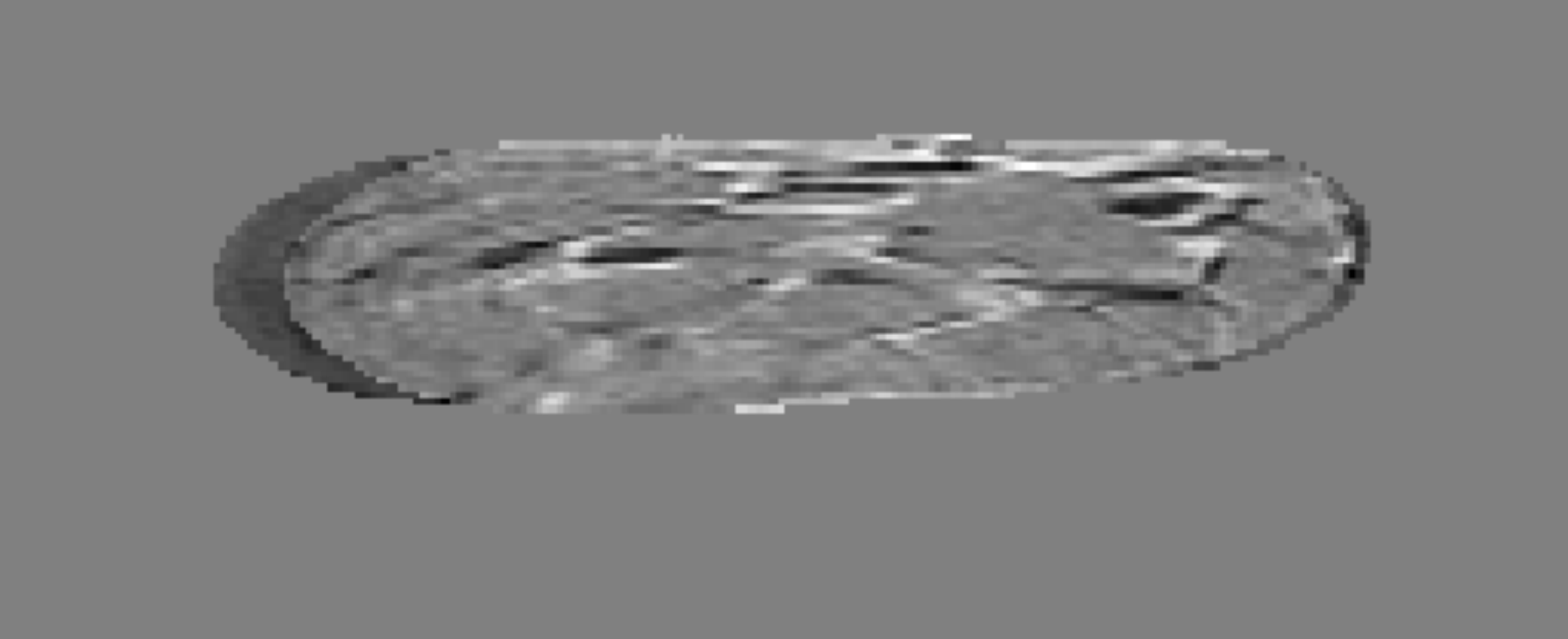}}
        \subfigure[]{
    \includegraphics[width=0.096\textwidth]{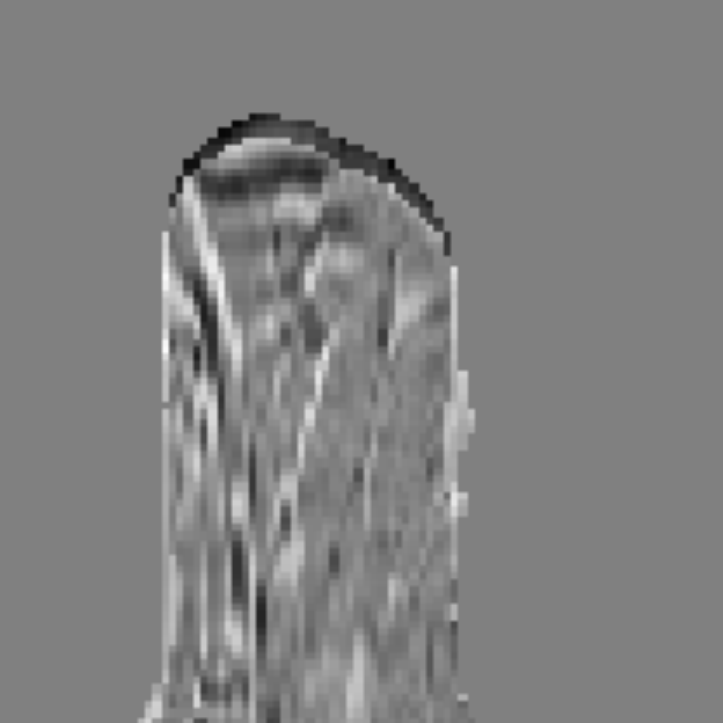}}  \\
        \subfigure{
    \includegraphics[width=0.45\textwidth]{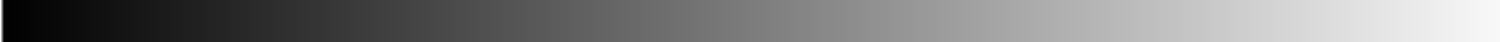}}
	\figcaption{\small\bf\it Two DBT simulations were created using in vivo MR acquisitions of a breast with two different real plate compressions to mimic the temporal imaging (Images have been segmented and mapped to effective X-ray attenuation). (a)-(c): Fixed image; (d)-(f): Moving image; (g)-(i): Initial difference between the fixed and moving images. (Left: Transverse view; Middle: Coronal view; Right: Sagittal view.)}
  \label{fig:FEM_220_8}
  \end{center}
  \end{figure}

Two temporal DBT simulations ($511\times208\times208$ voxels with a spatial resolution $0.215$mm in each direction) were created for this experiment, {\it i.e.,} a pair of MRI acquisitions obtained with two different real plate compressions of the breast (Figure \ref{fig:FEM_220_8} (a)-(c) and (d)-(f) with initial difference image in (g)-(i)).

\begin{figure}[!htb]
  \begin{center}
  \centering % \setlength{\floatsep}{10pt plus 3pt minus 2pt}
        \subfigure[]{
    \includegraphics[width=0.238\textwidth]{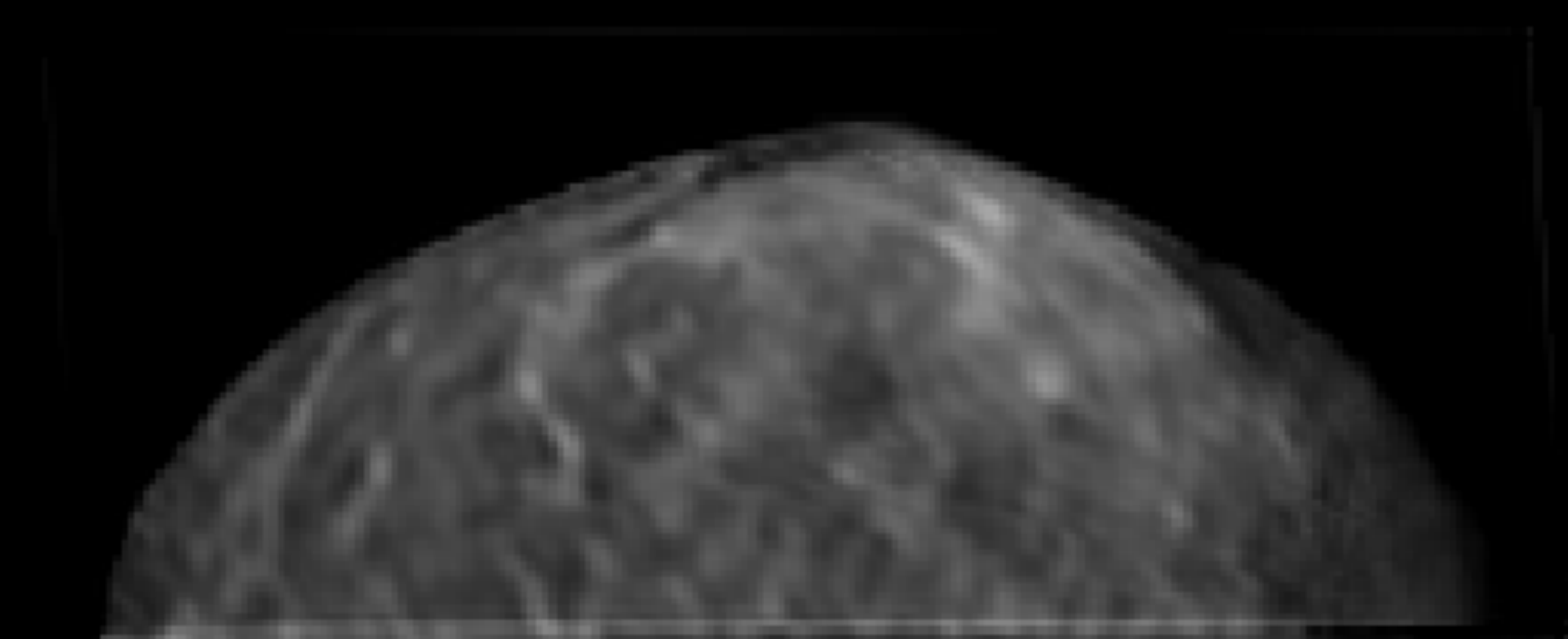}}
        \subfigure[]{
    \includegraphics[width=0.118\textwidth]{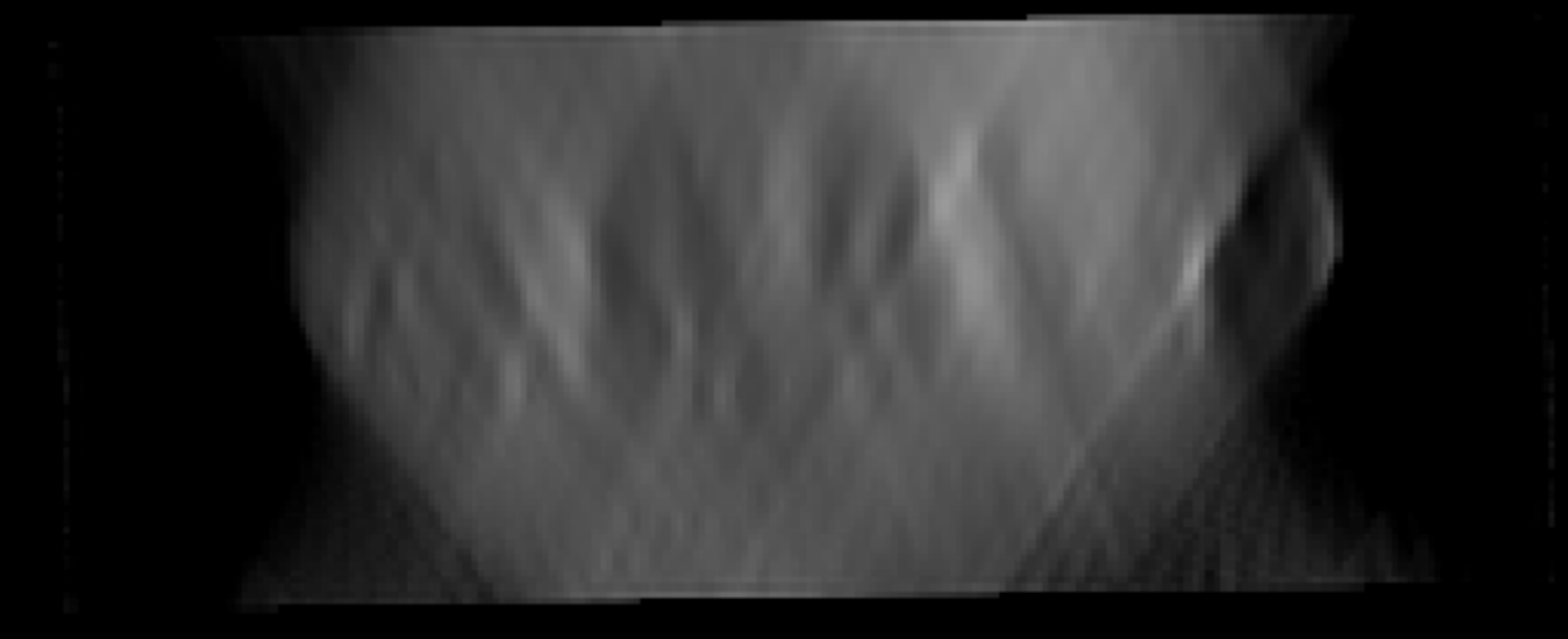}}
        \subfigure[]{
    \includegraphics[width=0.096\textwidth]{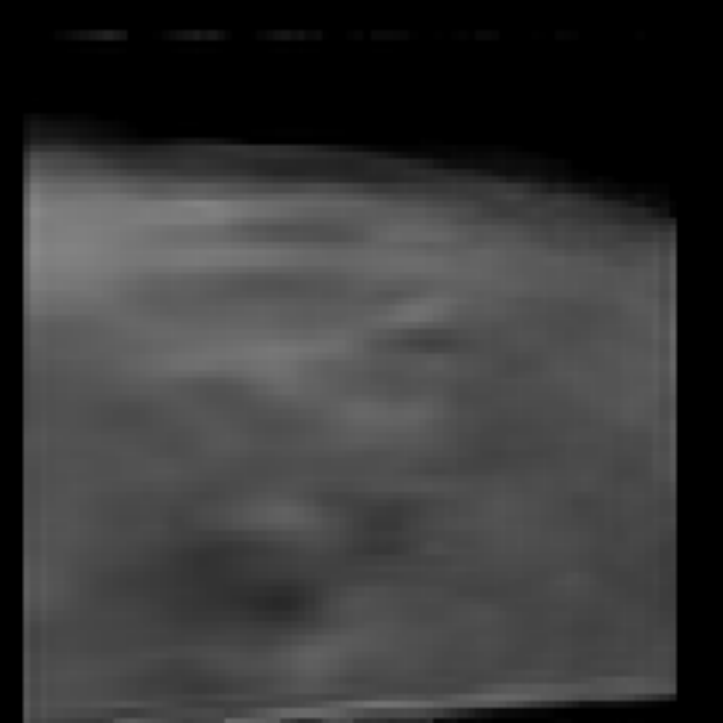}}  \\
        \subfigure[]{
    \includegraphics[width=0.238\textwidth]{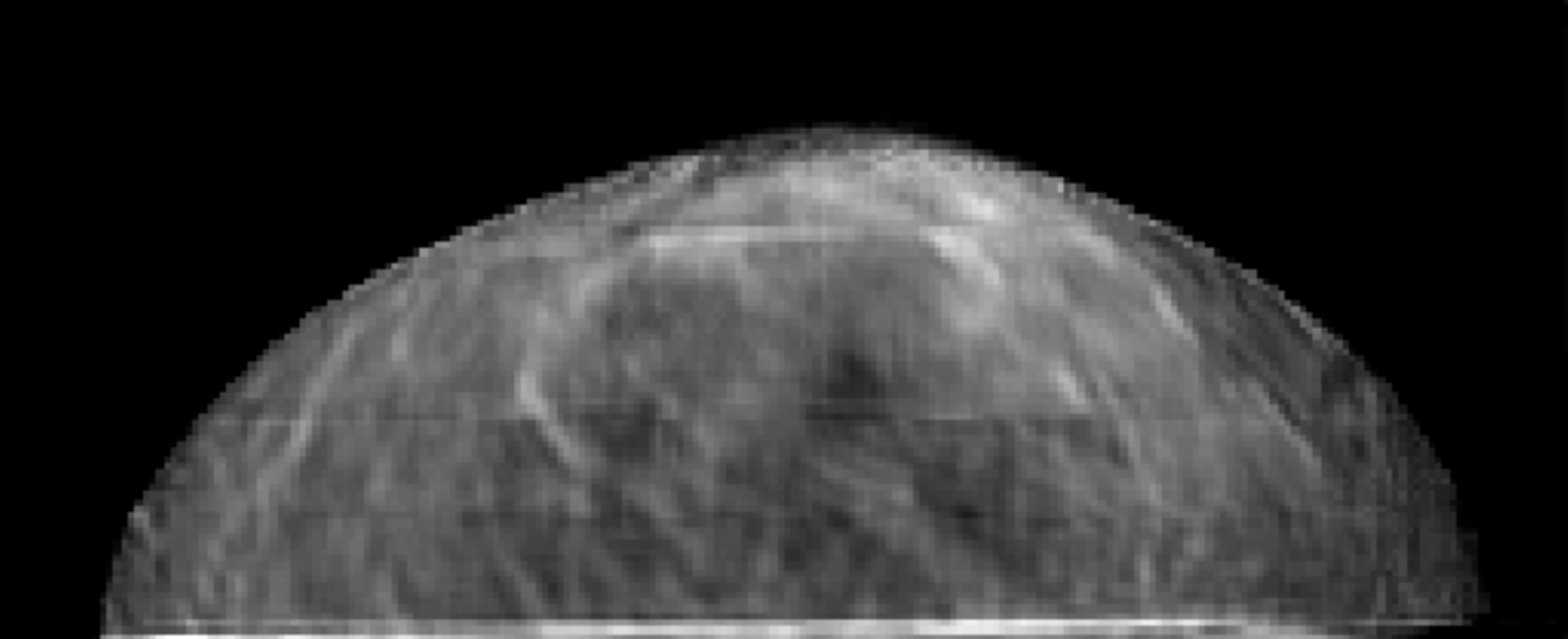}}
        \subfigure[]{
    \includegraphics[width=0.118\textwidth]{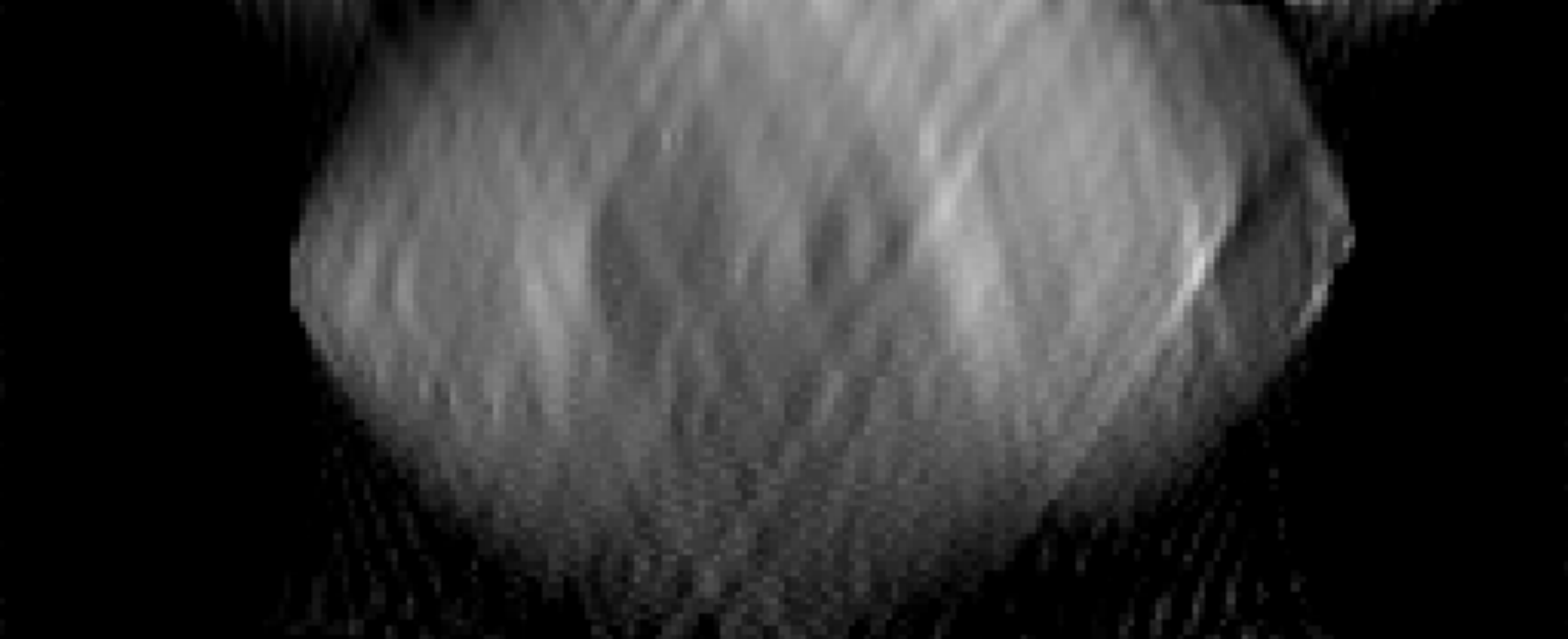}}
        \subfigure[]{
    \includegraphics[width=0.096\textwidth]{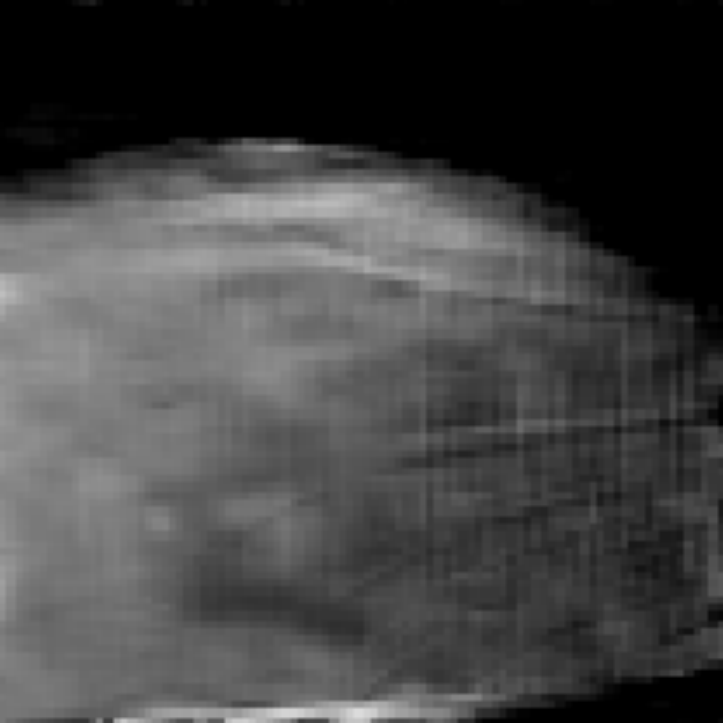}} \\
        \subfigure[]{
    \includegraphics[width=0.238\textwidth]{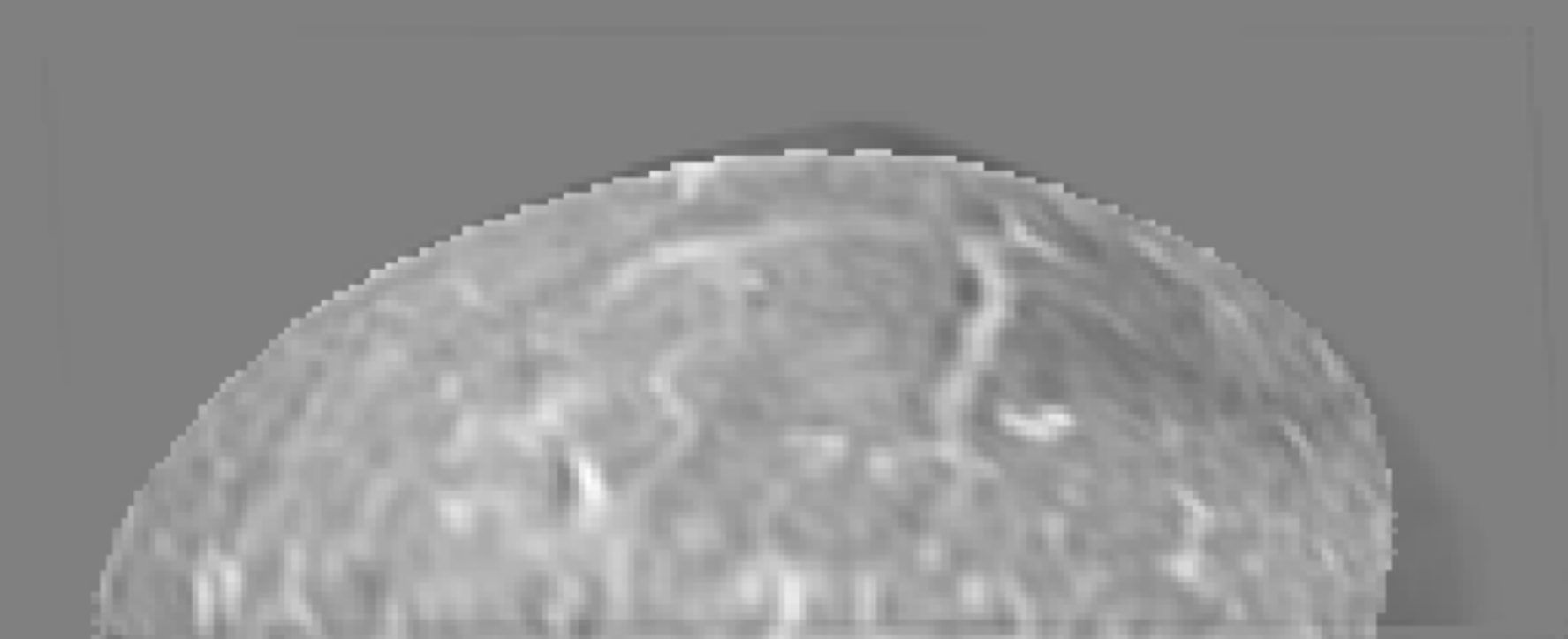}}
        \subfigure[]{
    \includegraphics[width=0.118\textwidth]{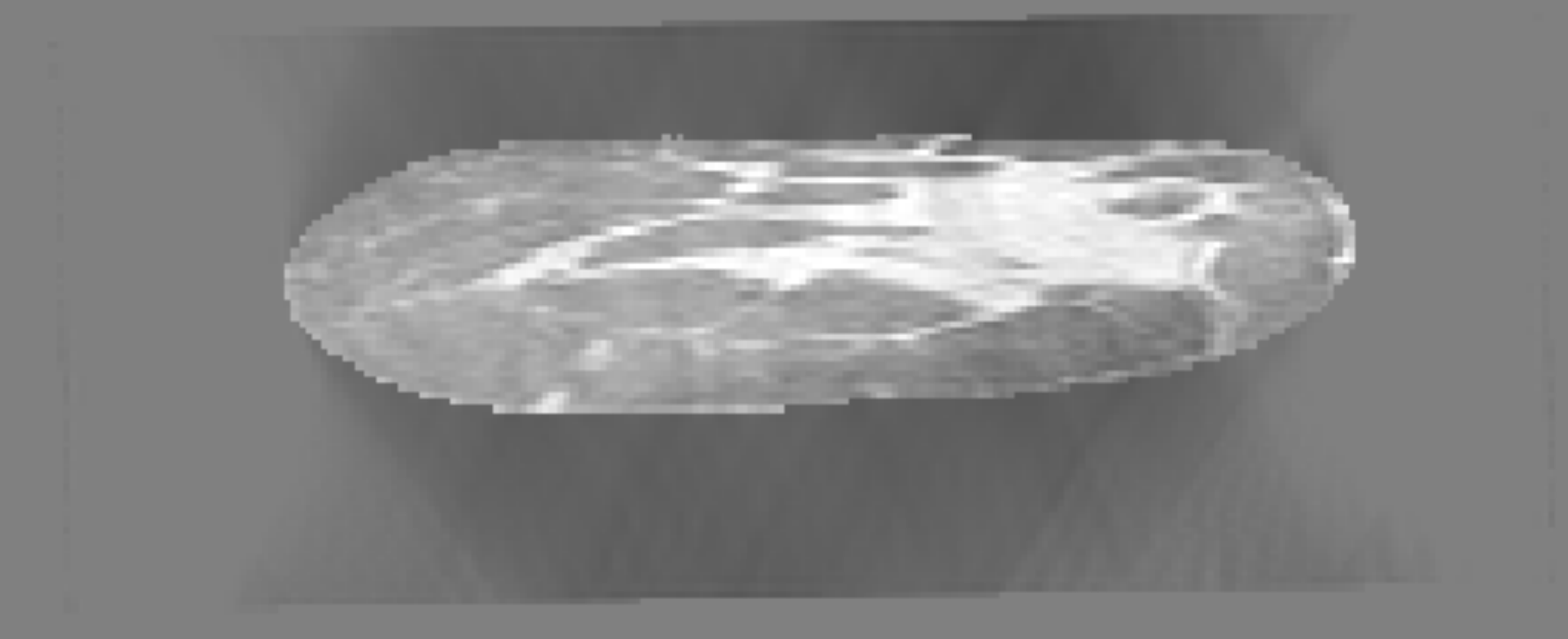}}
        \subfigure[]{
    \includegraphics[width=0.096\textwidth]{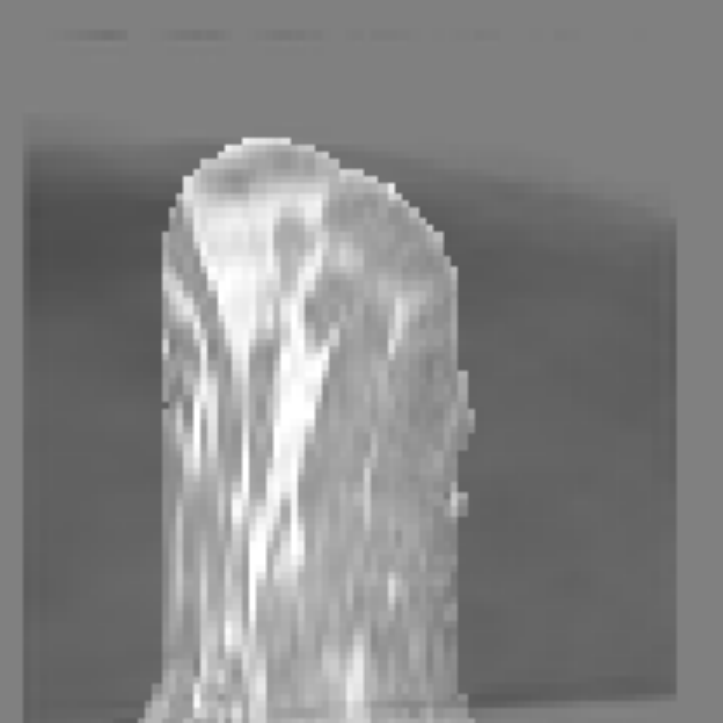}}  \\
        \subfigure[]{
    \includegraphics[width=0.238\textwidth]{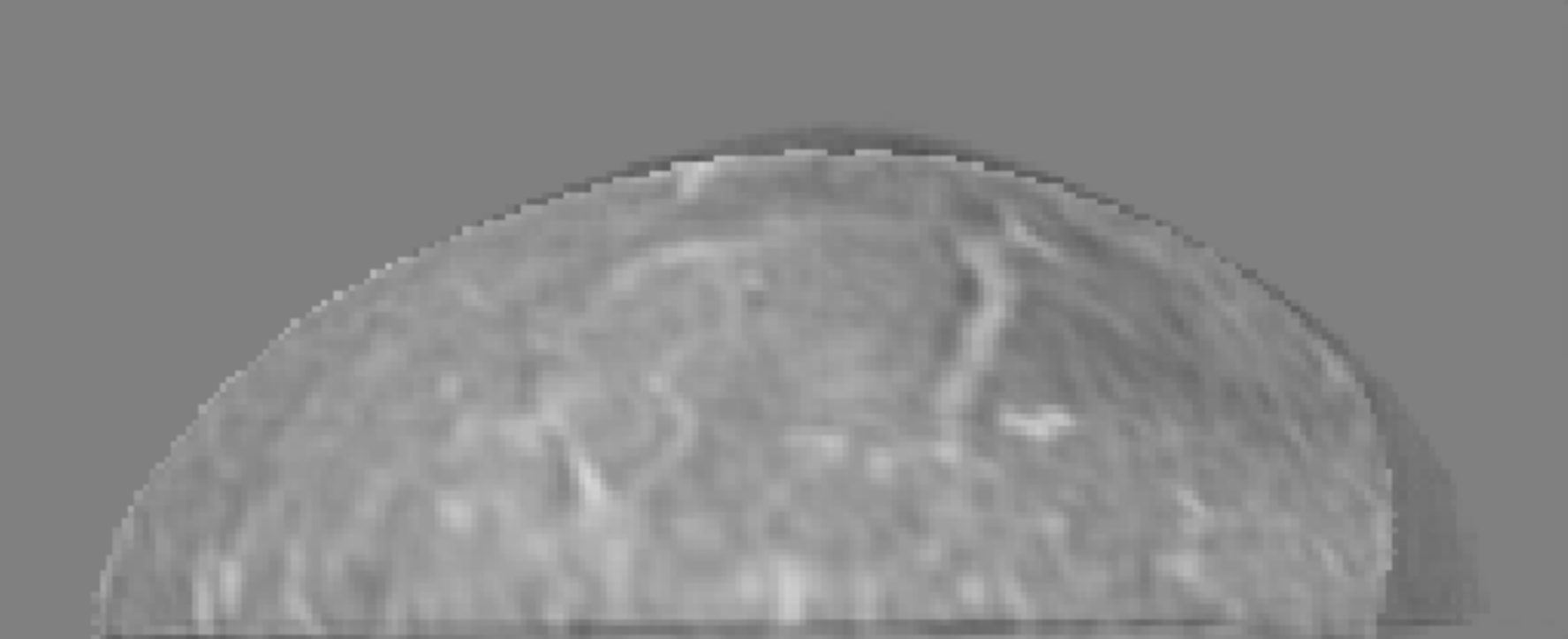}}
        \subfigure[]{
    \includegraphics[width=0.118\textwidth]{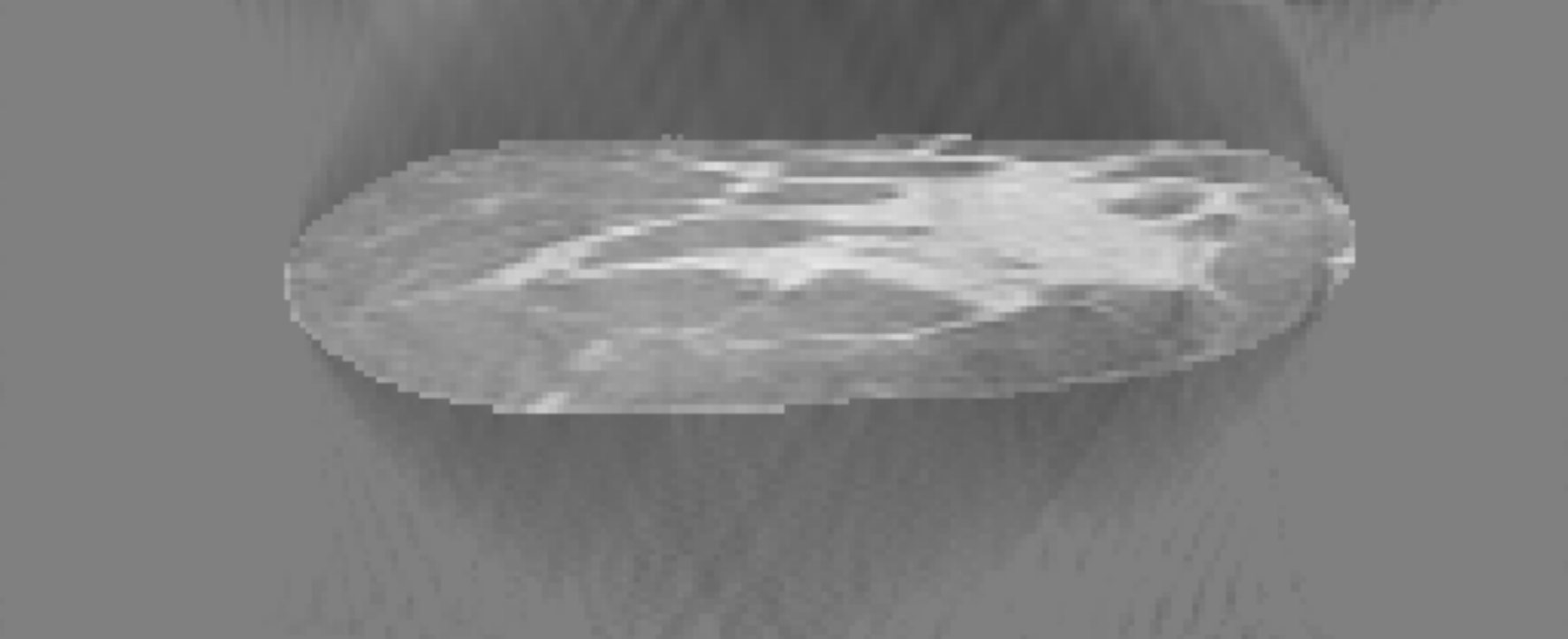}}
        \subfigure[]{
    \includegraphics[width=0.096\textwidth]{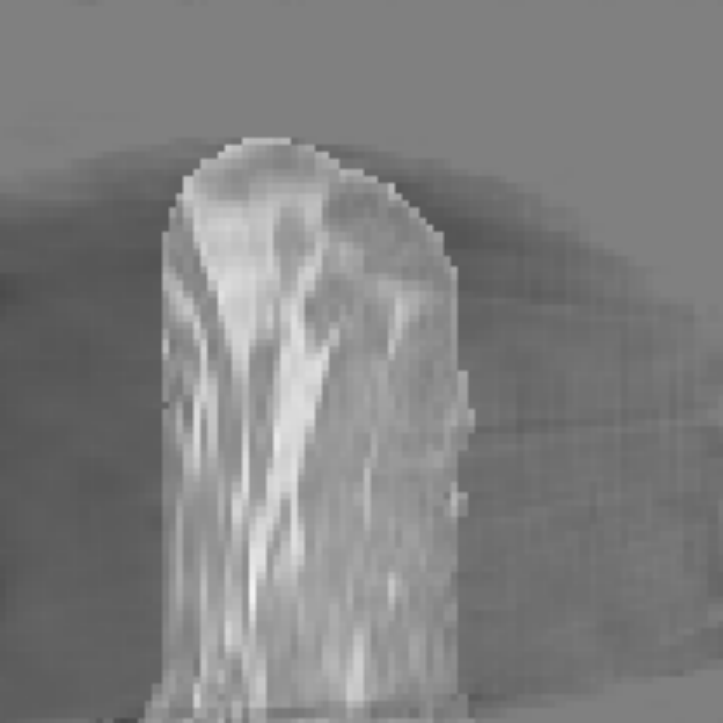}} \\
        \subfigure{
    \includegraphics[width=0.45\textwidth]{ReconRegnImage_220_11_8_4DBT.pdf}}
	\figcaption{\small\bf\it Sequential results {\it vs.} simultaneous results. (a)-(c): Sequential result; (d)-(f): Simultaneous result; (g)-(i): Differences between the sequential result and the fixed image; (j)-(l): Differences between the simultaneous result and the fixed image.}
  \label{fig:FEM_220_8_Results}
  \end{center}
  \end{figure}

From the comparison of the results (Figure \ref{fig:FEM_220_8_Results} (a)-(c) {\it vs.} (d)-(f)), our simultaneous method also outperformed the sequential method. First, the blurring effect of the sequential method was mitigated. Second, the out-of-plane structure was more compact with fewer radial artefacts. Third, there were no zero intensity (black regions) due to data truncation. There was no ground truth deformation available in this case because this pair of images was acquired in vivo. We applied our affine transformation based simultaneous framework to determine how well it could approximate the real non-rigid deformation. Despite this approximation, the difference images shows a fair reconstruction with appropriate registration using both methods (Figure \ref{fig:FEM_220_8_Results} (g)-(i) {\it vs.} (j)-(l)). There is no ground truth for the deformation of this data set, however from both the image appearance and the mean squared error (MSE in Table \ref{table:Numerical_Results2}), we can conclude that our joint method has successfully reconstructed the data with reasonable registration.

% Results of the experiment 3

\subsection{Simultaneous Method with Non-rigid B-spline} \label{sec:Simult_Bspline}

\begin{figure}[!htb]
  \centering
        \subfigure[]{
    \includegraphics[width=0.4818\textwidth]{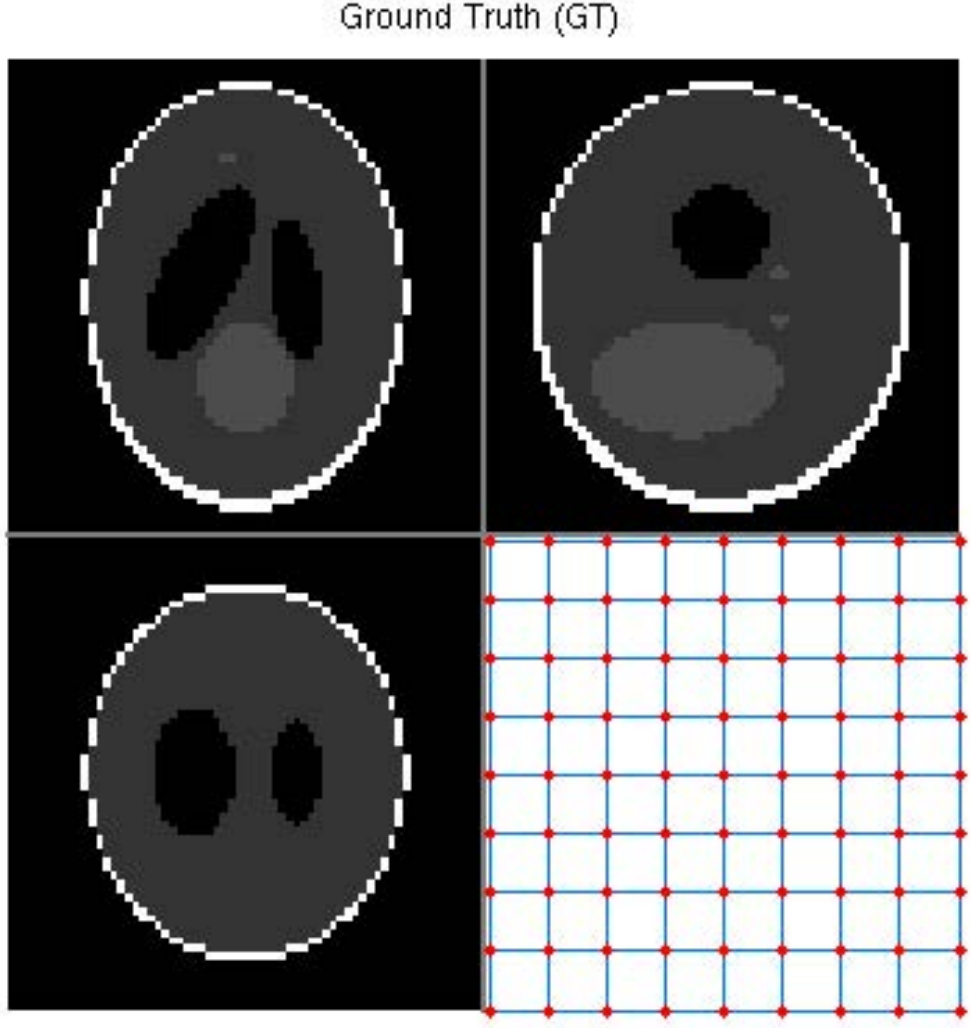}} \\
        \subfigure[]{
    \includegraphics[width=0.4818\textwidth]{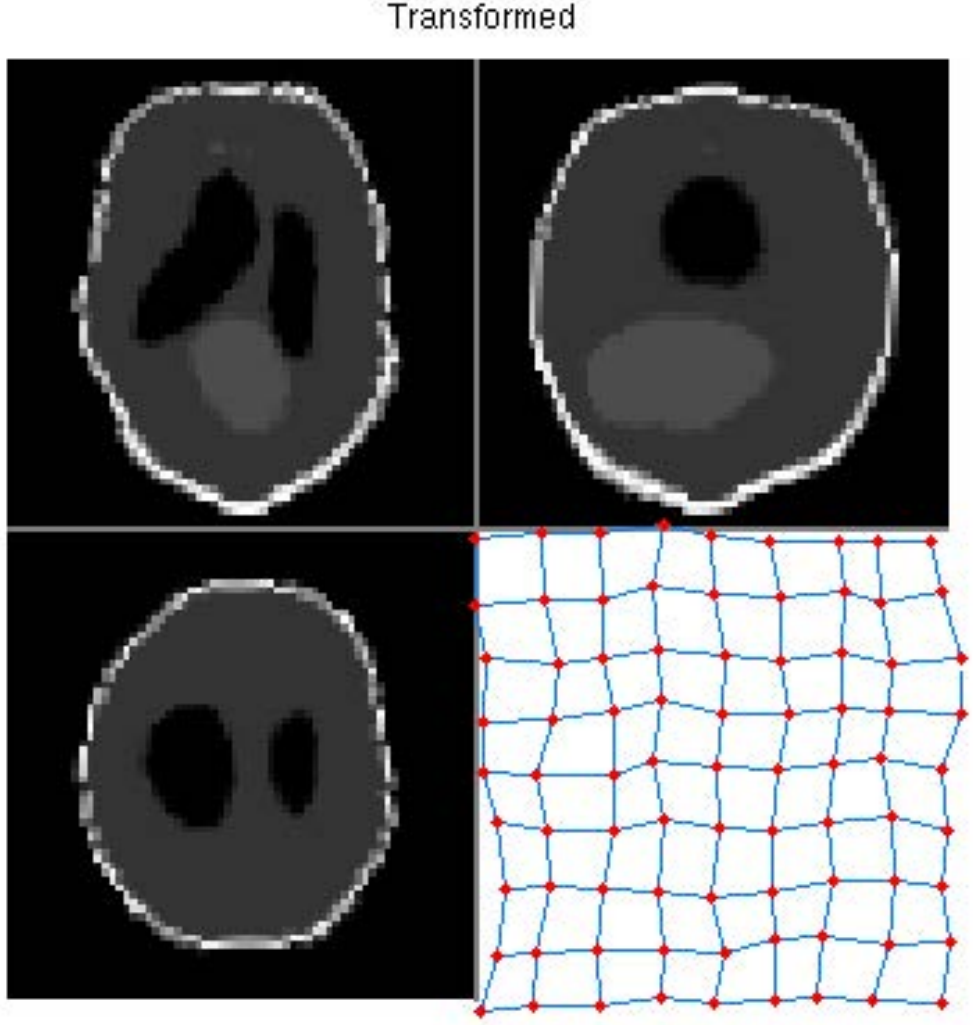}}
  \figcaption{\footnotesize\emph{(a): Original fixed 3D Shepp-Logan phantom and its regular grid for the central slice; (b): Transformed 3D Shepp-Logan phantom and its deformed grid for the central slice, {\it i.e.,} ground truth of the transformation. (Four sub-figures from top to bottom and from left to right are: Transverse view; Coronal view; Sagittal view; Grid of the central slice of the transverse view.)}}
  \label{fig:SL_Phantom_with_Grid}
\end{figure}

For the B-spline transformation model, we initially use a 3D Shepp-Logan phantom to test our simultaneous method. Figure \ref{fig:SL_Phantom_with_Grid}(a) shows the 3D Shepp-Logan phantom ($65\times65\times65$ in voxels), and also shows the central slice of each plane and the regular B-spline control point grid for the central slice of the transverse plane. The transformed phantom is shown in Figure \ref{fig:SL_Phantom_with_Grid}(b), and illustrates the ground truth of the transformation. This ground truth deformation is  simulated  with the B-spline transformation model using $9 \times 9 \times 9$ control points randomly offset in each dimension.

\begin{figure}[!htb]
  \centering
        \subfigure[]{
    \includegraphics[width=0.4818\textwidth]{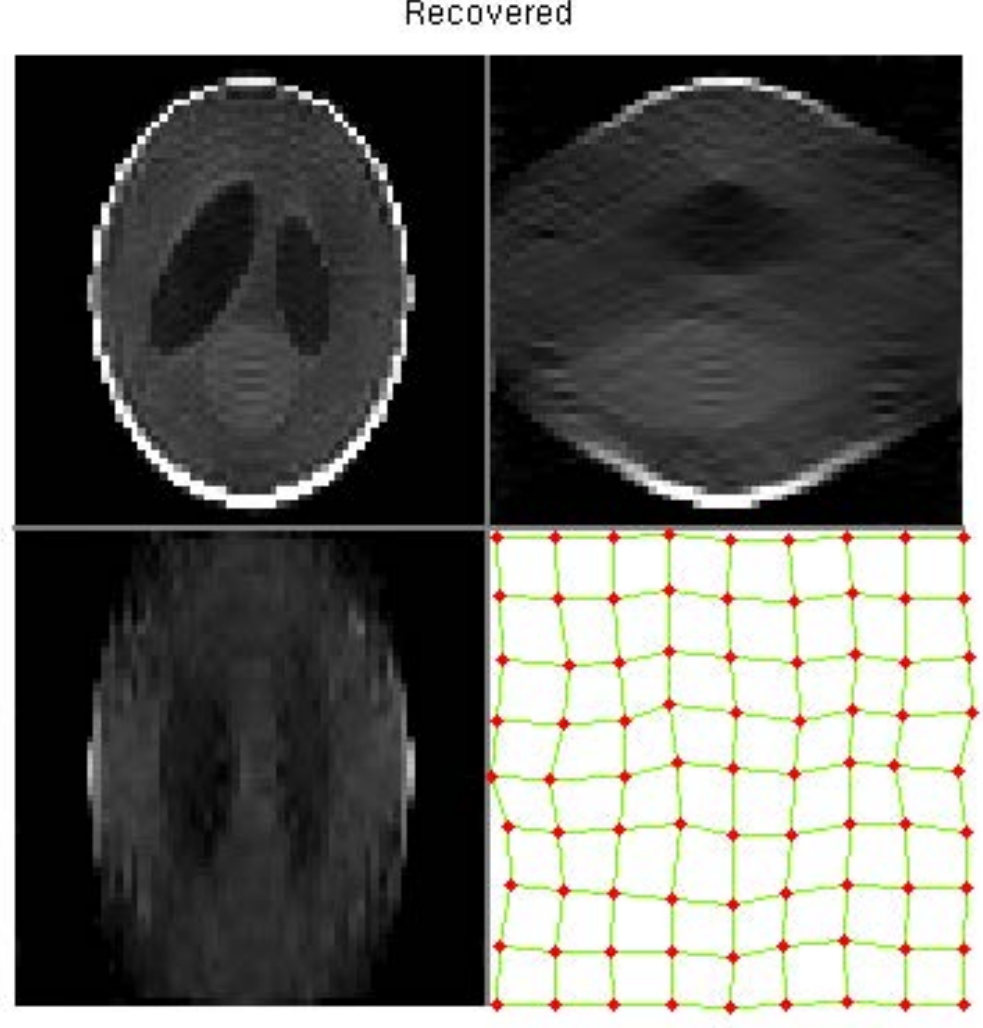}}  \\
        \subfigure[]{
    \includegraphics[width=0.4818\textwidth]{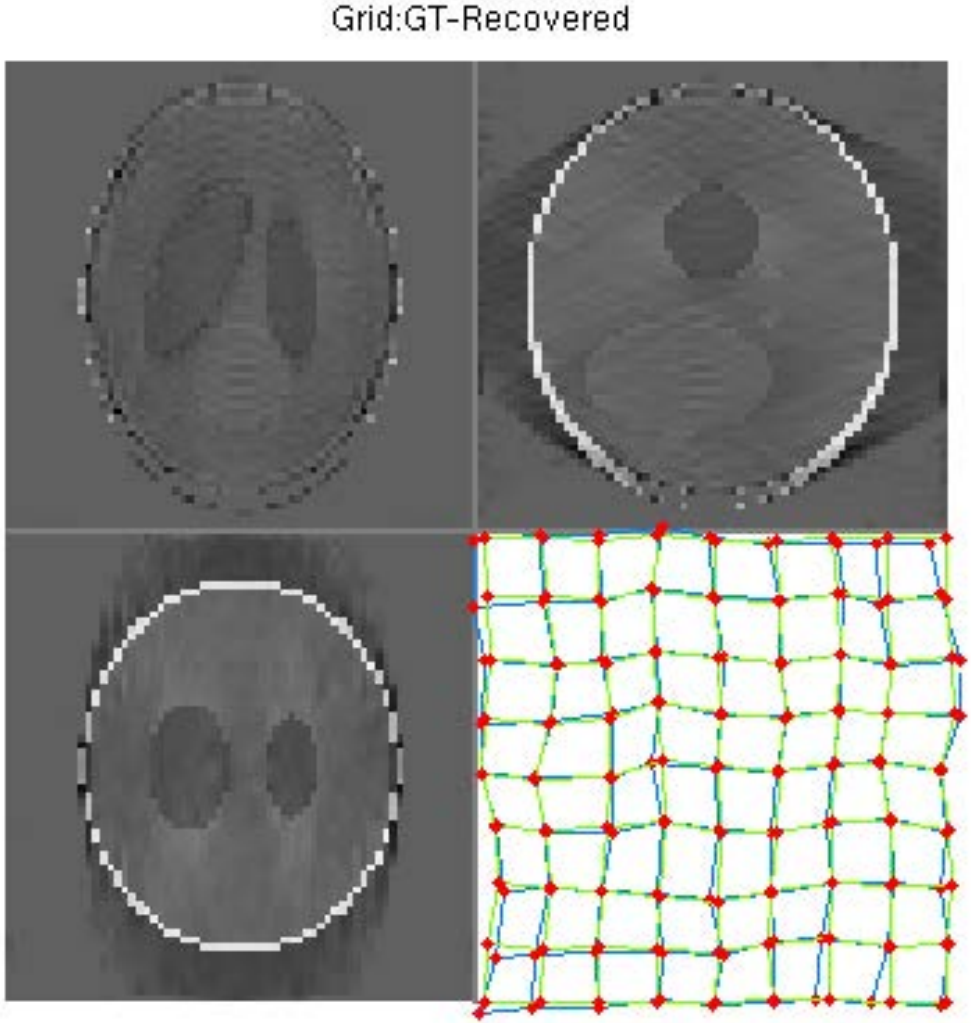}}
  \figcaption{\footnotesize\emph{(a): Simultaneous reconstruction and registration result and the registered control point grid for the central slice; (b): Difference image between the simultaneous result and the original fixed phantom. The registered control point grid is superimposed on the ground truth transformation and indicates that the ground truth transformation has been recovered well for the majority of control points.}}
  \label{fig:RecoveredGrid}
\end{figure}

\begin{figure*}
  \begin{center}
  \centering
 \includegraphics[width=0.698\textwidth]{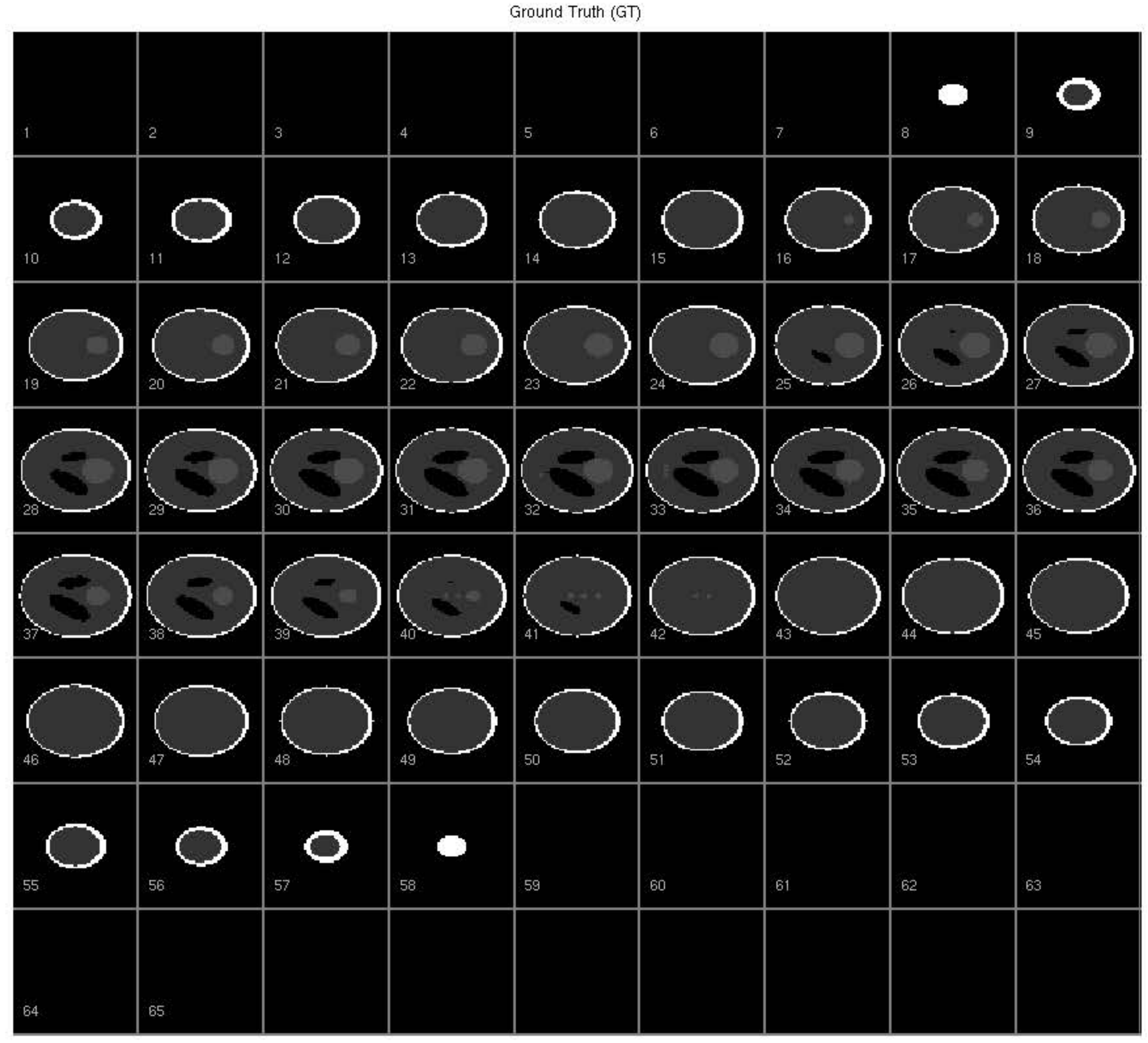}
  \figcaption{\small\bf\it A montage view of the original fixed 3D Shepp-Logan phantom.}
 \label{fig:FixedGroundTruthMontage}
  \end{center}
  \end{figure*}

\begin{figure*}
  \begin{center}
  \centering
 \includegraphics[width=0.698\textwidth]{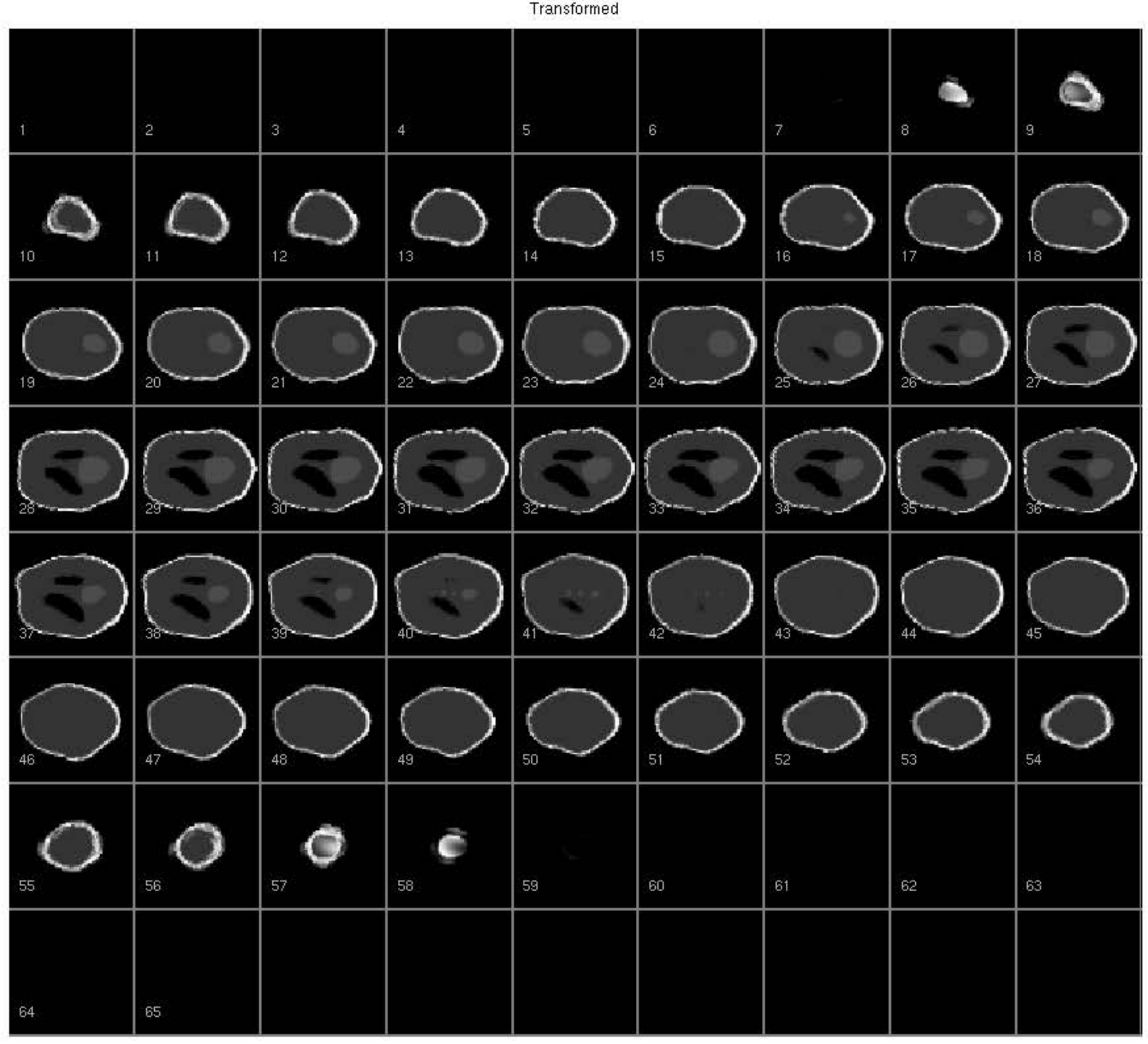}
  \figcaption{\small\bf\it A montage view of the transformed 3D Shepp-Logan phantom.}
  \label{fig:TransformedMontage}
  \end{center}
  \end{figure*}

\begin{figure*}
  \begin{center}
  \centering
 \includegraphics[width=0.698\textwidth]{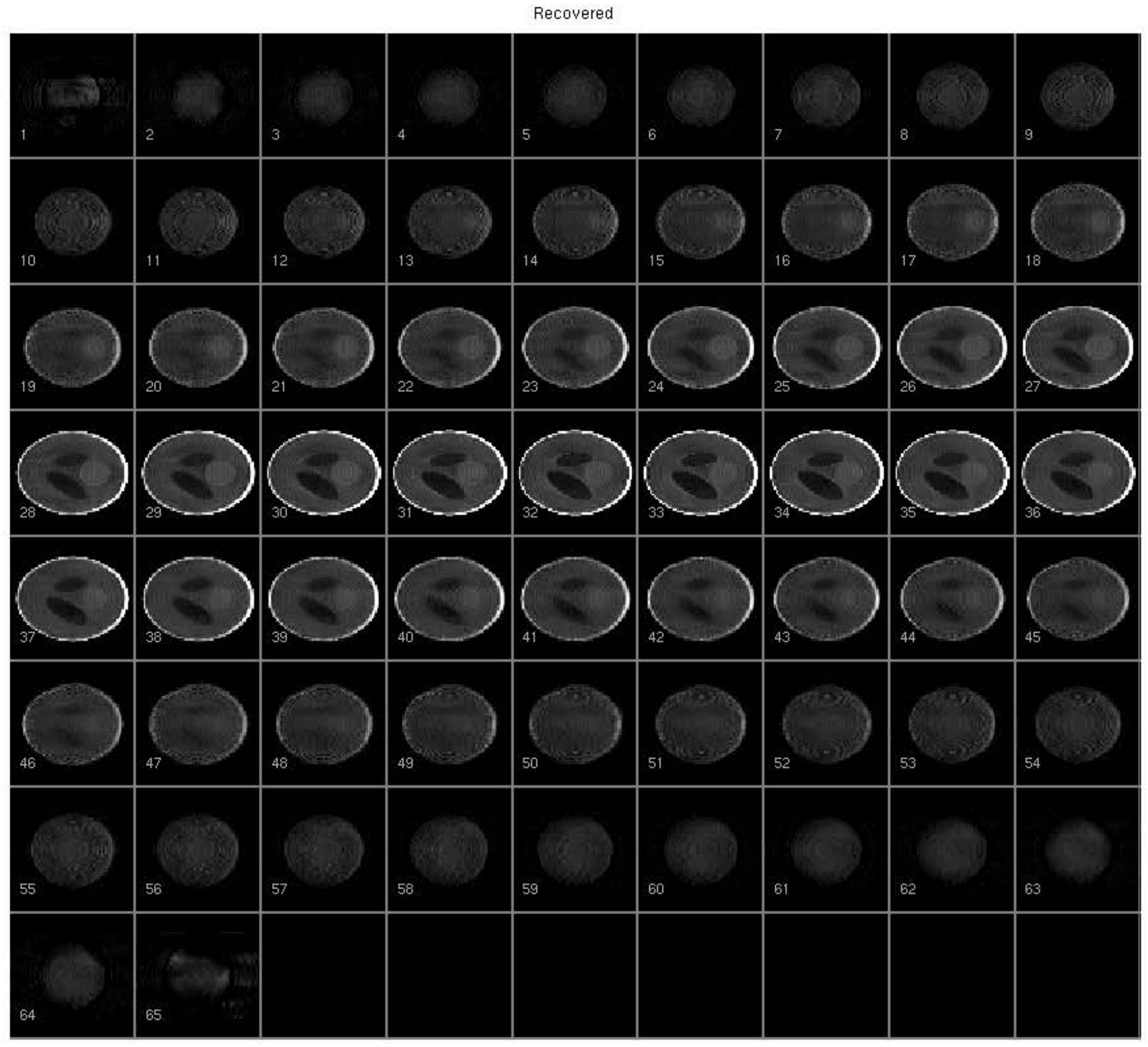}
  \figcaption{\small\bf\it A montage view of the joint reconstruction and registration result for the known B-spline transformation shown in Figure \ref{fig:TransformedMontage}.}
  \label{fig:RecoveredMontage}
  \end{center}
  \end{figure*}

\begin{figure*}
  \begin{center}
  \centering
 \includegraphics[width=0.698\textwidth]{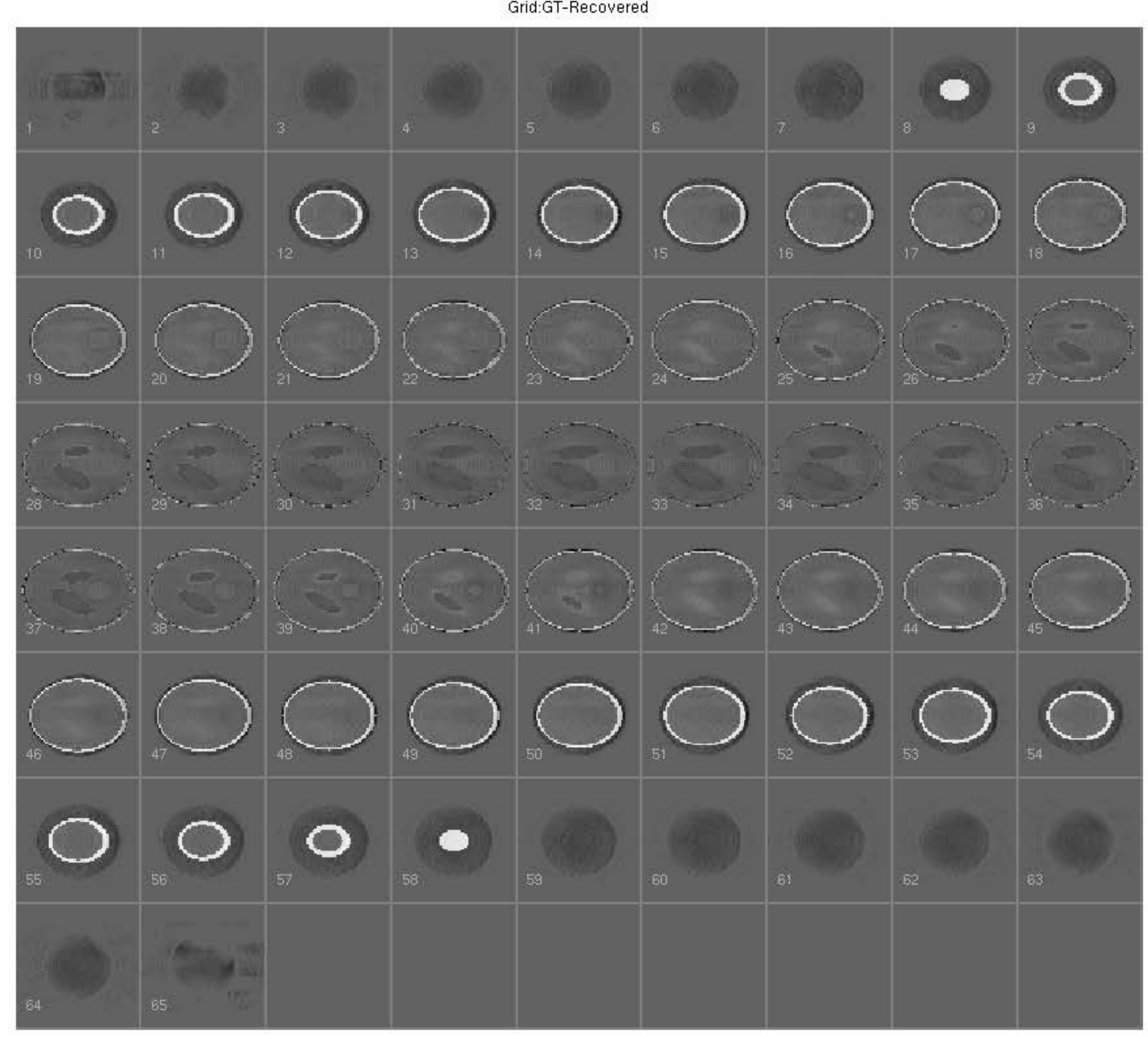}
  \figcaption{\small\bf\it A montage view of the difference image between the result of the joint method (Figure \ref{fig:RecoveredMontage}) and the original fixed phantom (Figure \ref{fig:FixedGroundTruthMontage}).}
  \label{fig:DiffImageMontage}
  \end{center}
  \end{figure*}

In this experiment, we set different ranges of perturbation for each direction (x-, y- and z-axis), {\it e.g.,} $[\mathrm{Range}_\mathrm{a},\mathrm{Range}_\mathrm{b}] = [-8,8]$, $= [-4,4]$, and $= [-2,2]$ (voxels) have been used respectively. Therefore, there are larger deformations for the in-plane slices and smaller ones for the out-of-plane. From the results shown in Figures \ref{fig:RecoveredGrid}(a) and (b), we can conclude that our simultaneous method has obtained an accurate reconstruction with a reasonable recovery of the non-rigid deformations. The montage views of the fixed, transformed, recovered, and difference images are shown in Figures \ref{fig:FixedGroundTruthMontage}, \ref{fig:TransformedMontage}, \ref{fig:RecoveredMontage}, and \ref{fig:DiffImageMontage}.

\section{Discussion}
\label{sec:discussion}

This work presented a threefold investigation of a joint framework in DBT reconstruction and registration using a pair of temporal data sets. This framework jointly considers these two tasks, and it is capable of recovering both the parametric transformations, and an enhanced, reconstructed image. We proposed a partially coupled iterative method, and further devised a fully coupled simultaneous method. By integrating the registration directly into the framework of the reconstruction problem, we are able to fully explore the interdependence between the transformation parameters and the 3D volume to be reconstructed.

Significantly, compared to the previous research on combining reconstruction and registration (or motion correction), our combined limited angle DBT problem has a much larger null space and is severely ill-posed, which makes the inverse problem more intriguing and more challenging. From Table \ref{table:LiteratureCompare}, we can see that for a typical 2D super-resolution problem previous studies used $5$ low resolution images to restore a high resolution image recovering only rotations and translations, and $32$ low resolution images for the affine registration. In general 3D problems, authors have used at least $60$ and up to $799$ forward projections covering a full-range of views, {\it i.e.,} $180^{\mathrm{o}}$ or $360^{\mathrm{o}}$, to perform the joint estimations. However, for our DBT application, we have two sets of data, which are observed at two time-points. Each of the data is acquired using only $11$ forward projections covering just $50^{\mathrm{o}}$ ($\pm25^{\mathrm{o}}$), and the two data sets overlap to a certain degree according to the original unknown deformations.

Section \ref{sec:Seqen_vs_Iter} demonstrated that our iterative method  outperformed the conventional sequential method using a software synthetic phantom image with affine transformation model. Inevitably, there are reconstruction artefacts showing up in the out-of-plane reconstructions; however, the in-plane structures have been reconstructed to a high precision. Compared to the transformed moving reconstruction using the sequential method, {\it i.e.,} the registration result of the reconstructed volume in Figure \ref{fig:ReconNoRegnToroid3D} (d)-(f) to the fixed image reconstruction in Figure \ref{fig:ReconNoRegnToroid3D} (a)-(c), our iterative method produced a more compact result to mitigate out-of-plane blurring.

In section \ref{sec:Seqen_vs_Simult}, we analysed our simultaneous method with various data sets using an affine transformation model, and the simultaneous method has achieved clearly superior results compared to the conventional sequential method. The experiment on the 3D toroid image clearly revealed that this approach has an advantage over the conventional method. The results of the breast MR image have further strengthened our confidence in the hypothesis that the reconstruction and registration have a reciprocal relationship. Importantly, plots of the cross-sectional line profiles confirmed that our combined method produced a superior reconstruction to the conventional method. In addition, the recovery of the transformation parameters was consistently accurate for both the 3D toroid and the breast MR data sets. Next, we attempted to reconstruct and register simulated DBT data sets created from real medio-lateral compressions of a breast imaged using MRI. As anticipated, the simultaneous approach still outperformed the conventional sequential method as demonstrated by the image appearance and MSE comparison (Figure \ref{fig:FEM_220_8_Results} and Table \ref{table:Numerical_Results2}). Although the improvements were limited in this experiment, this can be attributed, at least in part, to the fact that the affine transformation, which is a global parametric model, is insufficient to capture such a non-rigid breast deformation.

  \begin{figure}[!htb]
  \begin{center}
  \centering % \setlength{\floatsep}{10pt plus 3pt minus 2pt}
  \includegraphics[width=0.8\textwidth]{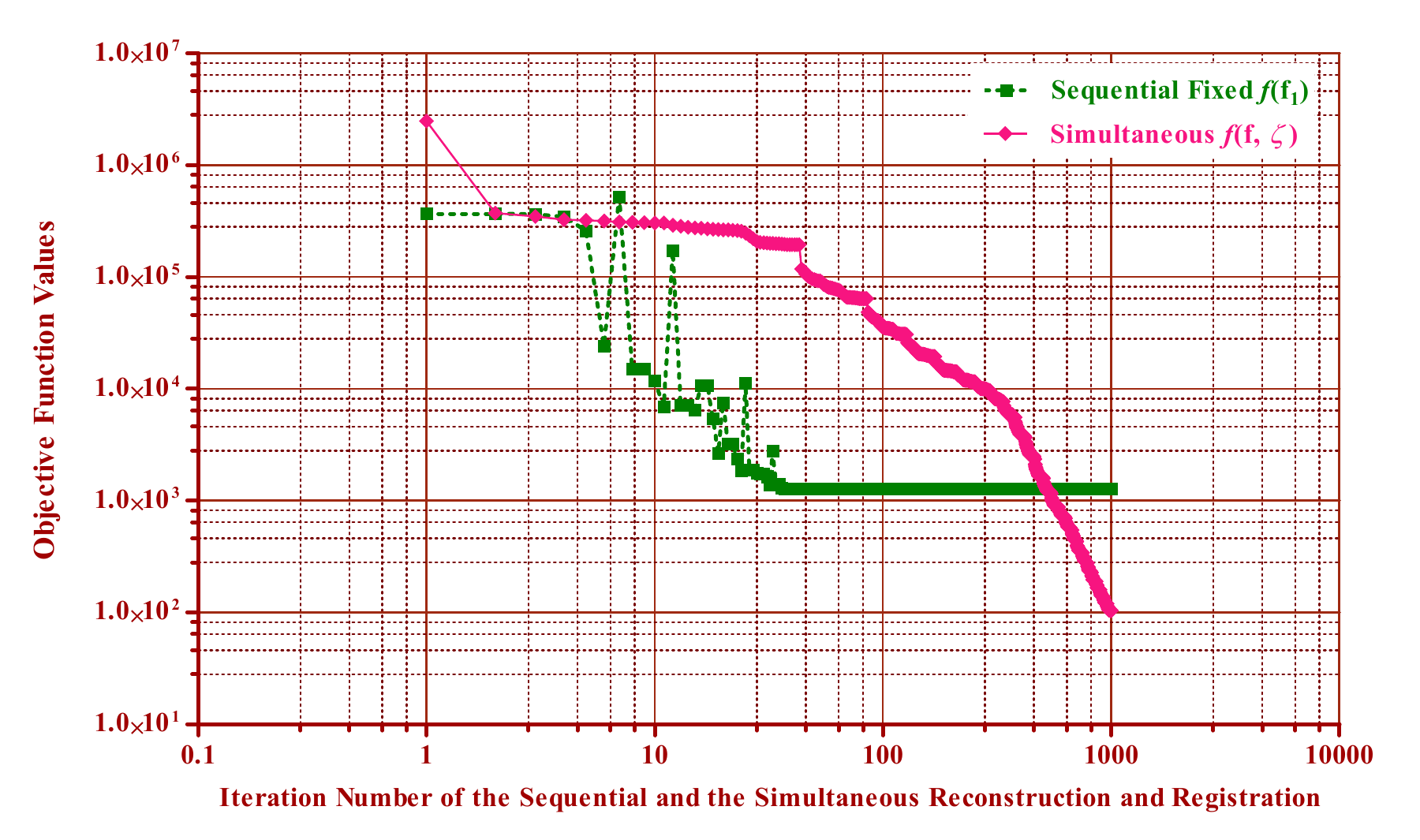}
  \figcaption{\small\bf\it The 3D toroid test case 1. Comparison of the objective function of the fixed image reconstruction using the sequential method $f(\mathrm{f}_1) = \frac{1}{2}\big\|A\mathrm{f}_1-\mathrm{p}_1\big\|^2$, and the objective function of the simultaneous method $f({\mathrm{f},\upzeta}) = \frac{1}{2} \Big(\big\| A\mathrm{f}-\mathrm{p}_1\big\|^2 + \big\| A\mathcal{T}_{{\upzeta}}\mathrm{f}-\mathrm{p}_2\big\|^2\Big)$.}
  \label{fig:Toroid_70_CostFun_Compare}
  \end{center}
  \end{figure}

  \begin{figure}[!htb]
  \begin{center}
  \centering % \setlength{\floatsep}{10pt plus 3pt minus 2pt}
  \includegraphics[width=0.8\textwidth]{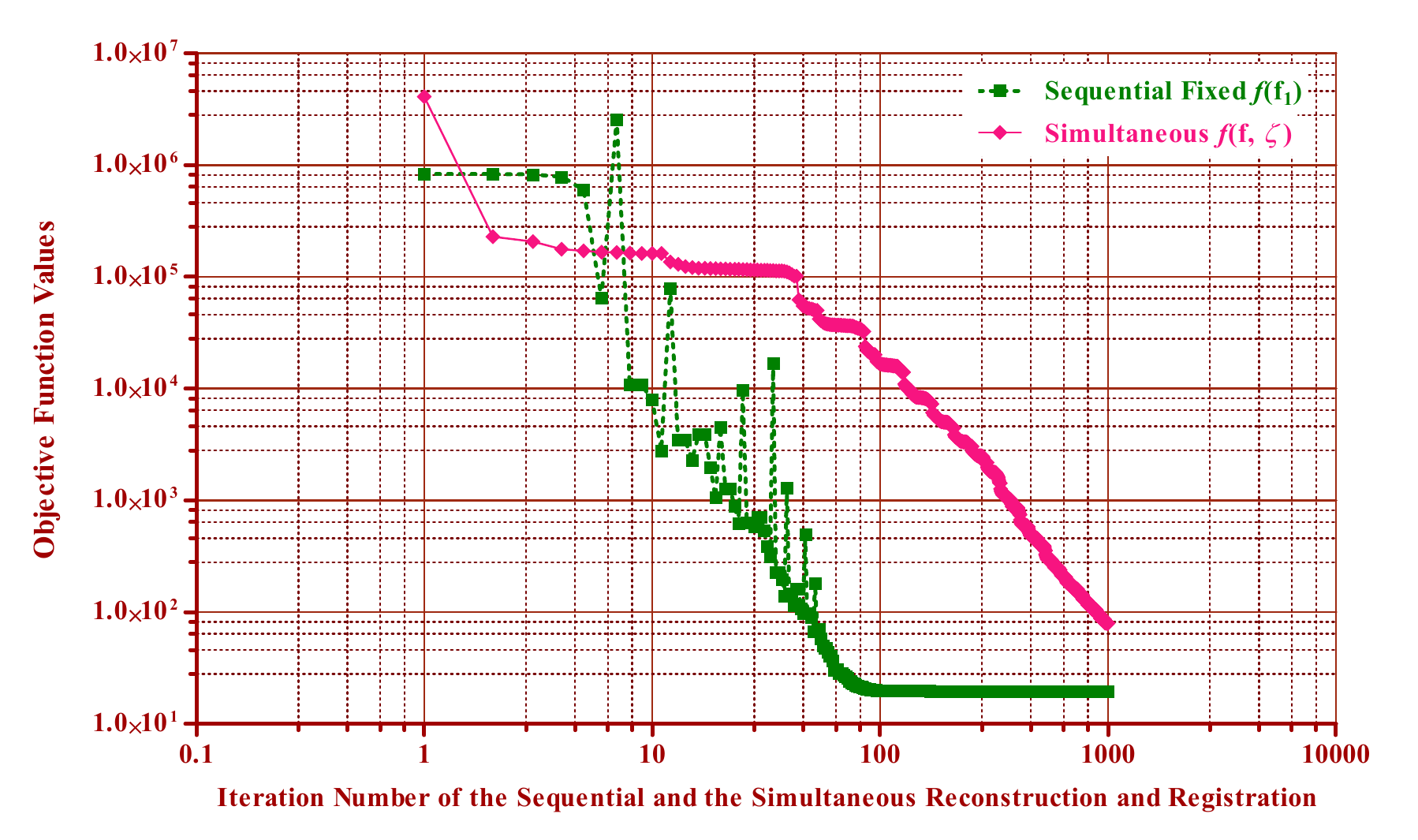}
  \figcaption{\small\bf\it The breast MRI test case 8. Comparison of the objective function of the fixed image reconstruction using the sequential method $f(\mathrm{f}_1) = \frac{1}{2}\big\|A\mathrm{f}_1-\mathrm{p}_1\big\|^2$, and the objective function of the simultaneous method $f({\mathrm{f},\upzeta}) = \frac{1}{2} \Big(\big\| A\mathrm{f}-\mathrm{p}_1\big\|^2 + \big\| A\mathcal{T}_{{\upzeta}}\mathrm{f}-\mathrm{p}_2\big\|^2\Big)$.}
  \label{fig:P1_128_CostFun_Compare}
  \end{center}
  \end{figure}

Figure \ref{fig:Toroid_70_CostFun_Compare} and Figure \ref{fig:P1_128_CostFun_Compare} show that the objective function decreases smoothly using our simultaneous method, but the convergence rate is low. More iteration could improve the convergence but iterations more the 1000 will have diminished improvement. The total number of reconstruction and registration iterations was set to 1000 for both the simultaneous and sequential methods. Figure \ref{fig:Toroid_70_CostFun_Compare} shows that our simultaneous method obtained a better convergence than the sequential method, but Figure \ref{fig:P1_128_CostFun_Compare} shows a lower convergent value using the sequential method. In fact, contrary to the objective function of the simultaneous method, we only displayed the $f(\mathrm{f}_1)$ (one of the two reconstruction objective functions), which represents the reconstruction of the fixed image of the sequential method. Additionally, the trend of the objective function was still downwards using the simultaneous method.

A different number of inner iterations for the two decoupled optimisation steps would affect convergence. The experiments pointed to the likelihood that a smaller number of inner iterations of Equations \ref{ReconDecoupledG} and \ref{RegnDecoupledG} gave better results, but it would slow down the convergence to keep the total number of iterations constant. A common way to compare the convergence rate of optimisation algorithms is to plot the objective function value as a function of the iteration. Since each method in this study optimised a different objective function, direct comparison of the individual objective function values would not be appropriate.

We also calculated the MSE error and the relative error between the reconstruction and registration results and the original fixed image, {\it i.e.,} ground truth, for the two methods (Tables \ref{table:Numerical_Results2} and \ref{table:Relative_Errors_2}). These comparison indicates that our simultaneous method has consistently produced a better result, and it is in line with our hypothesis that combining the two tasks leads to an improvement in the reconstruction, which in turn enables a more accurate registration.

\begin{table}[!htb]
  \caption{\small\emph{Comparison of the MSE. The MSE of the sequential method is $\frac{1}{\mathrm{D}_3} \| {\mathrm{f}}_1^\star - {\mathrm{f}}_1^{\mathrm{g}} \|^2$ (Difference between the result of the transformed moving image reconstruction and the original fixed image), and the MSE of the simultaneous method is given by $\frac{1}{\mathrm{D}_3}\| {\mathrm{f}}^\star - {\mathrm{f}}_1^{\mathrm{g}} \|^2$.}}
  \centering
  \scalebox{0.858}{
  \setlength{\floatsep}{10pt plus 3pt minus 2pt} % Set the length between the table and the text body
  \begin{tabular}{p{4.8cm}^p{2.1cm}^p{4.1cm}^p{4.5cm}}
  \addlinespace
  \toprule\rowstyle{\bfseries}
  \footnotesize
  ~                             & Initial               & Sequential Method       & Simultaneous Method         \\
  \midrule\rowstyle{\mdseries}
  Toroid Phantom                & $1.31\times10^{6}$    & $7.46\times10^{3}$    & $0.24\times10^{3}$        \\
  \midrule\rowstyle{\mdseries}
  Uncompressed Breast MRI       & $1.18\times10^{6}$    & $6.04\times10^{3}$    & $3.01\times10^{3}$        \\
  \midrule\rowstyle{\mdseries}
  In vivo DBT simulation        & $5.32\times10^{6}$    & $3.68\times10^{4}$    & $3.22\times10^{4}$        \\
  \bottomrule
  \end{tabular}
  }
  \label{table:Numerical_Results2}
\end{table}

\begin{table}[!htb]
  \caption{\small\emph{Comparison of the relative error, which is defined by $ \frac{ \| {\mathrm{f}}_1^\star - {\mathrm{f}}_1^{\mathrm{g}} \|^2 }{ \| {\mathrm{f}}_1^{\mathrm{g}} \|^2 }$ and $ \frac{ \| {\mathrm{f}}^\star - {\mathrm{f}}_1^{\mathrm{g}} \|^2 }{ \| {\mathrm{f}}_1^{\mathrm{g}} \|^2 }$ for the sequential and simultaneous method respectively.}}
  \centering
  \scalebox{0.858}{
  \setlength{\floatsep}{10pt plus 3pt minus 2pt} % Set the length between the table and the text body
  \begin{tabular}{p{4.8cm}^p{2.1cm}^p{4.1cm}^p{4.5cm}}
  \addlinespace
  \toprule\rowstyle{\bfseries}
  \footnotesize
  ~                             & Initial            & Sequential Method       	& Simultaneous Method         	\\
  \midrule\rowstyle{\mdseries}
  Toroid Phantom                & $1$   				& $0.0057$    				& $0.0002$         				\\
  \midrule\rowstyle{\mdseries}
  Uncompressed Breast MRI       & $1$   				& $0.0051$    				& $0.0026$         				\\
  \midrule\rowstyle{\mdseries}
  In-vivo DBT simulation        & $1$   				& $0.0058$   				& $0.0051$        				\\
  \bottomrule
  \end{tabular}
  }
  \label{table:Relative_Errors_2}
\end{table}

The third investigation evaluated incorporating a non-rigid transformation model into our combined framework (Section \ref{sec:Simult_Bspline}). In particular, we employed the B-spline transformation model and tested it with our simultaneous method. Results generated using a 3D Shepp-Logan phantom image offer compelling evidence that our simultaneous method has successfully reconstructed the volume with accurate recovery of the non-rigid deformations.

The construction of the objective function of our joint framework assumes that there is no change in the breast (such as the growth of a tumour or due to the differences in image acquisition parameters) between the two time-points being reconstructed and registered. We envisage a subsequent step where we compare the reconstructed and registered volume $\mathrm{f}^\star$ with the original acquisitions $\mathrm{p}_1$ and $\mathrm{p}_2$, to detect the change.

%The construction of the objective function of our joint framework assumes that there is no change in the breast (such as the growth of a tumour or due to the differences in image acquisition parameters) between the two time-points being reconstructed and registered. Consequently we use SSD as the registration similarity metric. We tested the efficacy of our framework when there are differences between the two temporal data sets due to changes in the breast  in Section \ref{sec:Simult_Detect_Changes}. Interestingly, compared to the sequential results, we could observe simulated micro-calcifications and a mass in our simultaneous results. This suggests that the unified reconstruction could be interpreted as an intermediate stage of the breast changing. We envisage a subsequent step where we compare the reconstructed and registered volume $\mathrm{f}^\star$ with the original acquisitions $\mathrm{p}_1$ and $\mathrm{p}_2$, to detect the change.

%______________________________________________________________________________________________________________________
% Section 8: Conclusion and Perspectives
\section{Conclusion and Perspectives}
\label{sec:conclusion_and_perspectives}

As far as we aware this is the first time that the joint reconstruction and registration framework has been proposed in DBT. In this work, we have presented two novel methods, {\it i.e.,} iterative and simultaneous methods, under the joint reconstruction and registration framework.  Affine and non-linear B-spline registration transformation models have been incorporated as plug-ins. These methods were  motivated by the goal of detecting changes between the two sets of temporal data. Essentially, the embedding of registration provides more information for the reconstruction; however, it does not simply increase the number of forward projections because the  temporal data sets overlap significantly (assuming both views are either CC or MLO). By experimenting with various simulation data sets, we conclude that this framework produced satisfactory results in both registration accuracy and reconstruction appearance.

This research has also raised many interesting points to explore in future work. First, we would like to implement GPU acceleration for some components of our framework, {\it e.g.,} forward and backward projectors. Second, we could tackle this large-scale optimisation problem using multi-scale and multi-resolution techniques to reduce execution time further and avoid convergence to local minima. Using our framework it will be straightforward to incorporate other non-rigid transformation models and priors to regularise the solution. It may also be applied to the combined reconstruction and registration of two view (cranial-caudal (CC) and Mediolateral-oblique (MLO)) DBT data sets, to overcome the null-space limitation of the individual views and produce a single reconstructed volume with improved depth resolution.

\bibliography{DecoupledMethodBib}

\begin{thebibliography}{41}
\providecommand{\natexlab}[1]{#1}
\providecommand{\url}[1]{\texttt{#1}}
\expandafter\ifx\csname urlstyle\endcsname\relax
  \providecommand{\doi}[1]{doi: #1}\else
  \providecommand{\doi}{doi: \begingroup \urlstyle{rm}\Url}\fi

\bibitem[Cand\`{e}s et~al.(2006)Cand\`{e}s, Romberg, and Tao]{Candes2006}
E.J. Cand\`{e}s, J.~Romberg, and T.~Tao.
\newblock Robust uncertainty principles: exact signal reconstruction from
  highly incomplete frequency information.
\newblock \emph{Information Theory, IEEE Transactions on}, 52\penalty0
  (2):\penalty0 489--509, 2006.
\newblock ISSN 0018-9448.
\newblock \doi{10.1109/TIT.2005.862083}.

\bibitem[Chung et~al.(2006)Chung, Haber, and Nagy]{Chung2006}
Julianne Chung, Eldad Haber, and James Nagy.
\newblock Numerical methods for coupled super-resolution.
\newblock \emph{Inverse Problems}, 22\penalty0 (4):\penalty0 1261, 2006.
\newblock URL \url{http://stacks.iop.org/0266-5611/22/i=4/a=009}.

\bibitem[Chung et~al.(2010)Chung, Sternberg, and Yang]{Chung2010}
Julianne Chung, Philip Sternberg, and Chao Yang.
\newblock High-performance three-dimensional image reconstruction for molecular
  structure determination.
\newblock \emph{International Journal of High Performance Computing
  Applications}, 24\penalty0 (2):\penalty0 117--135, 2010.
\newblock \doi{10.1177/1094342009106293}.
\newblock URL \url{http://hpc.sagepub.com/content/24/2/117.abstract}.

\bibitem[CRUK(2010)]{CRUK2010}
CRUK.
\newblock Breast screening in the {UK}: A brief history., 2010.
\newblock URL
  \url{http://info.cancerresearchuk.org/cancerstats/types/breast/screening/history/}.

\bibitem[Dempster et~al.(1977)Dempster, Laird, and Rubin]{Dempster1977}
A.~P. Dempster, N.~M. Laird, and D.~B. Rubin.
\newblock Maximum likelihood from incomplete data via the {EM} algorithm.
\newblock \emph{Journal of the Royal Statistical Society. Series B
  (Methodological)}, 39\penalty0 (1):\penalty0 1--38, 1977.
\newblock ISSN 00359246.
\newblock URL \url{http://www.jstor.org/stable/2984875}.

\bibitem[DobbinsIII and Godfrey(2003)]{DobbinsIII2003}
James~T DobbinsIII and Devon~J Godfrey.
\newblock Digital x-ray tomosynthesis: current state of the art and clinical
  potential.
\newblock \emph{Physics in Medicine and Biology}, 48\penalty0 (19):\penalty0
  R65, 2003.
\newblock URL \url{http://stacks.iop.org/0031-9155/48/i=19/a=R01}.

\bibitem[Fessler(1994)]{Fessler1994}
J.A. Fessler.
\newblock Penalized weighted least-squares image reconstruction for positron
  emission tomography.
\newblock \emph{Medical Imaging, IEEE Transactions on}, 13\penalty0
  (2):\penalty0 290--300, 1994.
\newblock ISSN 0278-0062.
\newblock \doi{10.1109/42.293921}.

\bibitem[Fessler(2010)]{Fessler2010}
J.A. Fessler.
\newblock Optimization transfer approach to joint registration / reconstruction
  for motion-compensated image reconstruction.
\newblock In \emph{Biomedical Imaging: From Nano to Macro, 2010 IEEE
  International Symposium on}, pages 596--599, 2010.
\newblock \doi{10.1109/ISBI.2010.5490108}.

\bibitem[Gur et~al.(2009)Gur, Abrams, Chough, Ganott, Hakim, Perrin, Rathfon,
  Sumkin, Zuley, and Bandos]{Gur2009}
David Gur, Gordon~S. Abrams, Denise~M. Chough, Marie~A. Ganott, Christiane~M.
  Hakim, Ronald~L. Perrin, Grace~Y. Rathfon, Jules~H. Sumkin, Margarita~L.
  Zuley, and Andriy~I. Bandos.
\newblock Digital breast tomosynthesis: Observer performance study.
\newblock \emph{American Journal of Roentgenology}, 193\penalty0 (2):\penalty0
  586--591, 2009.
\newblock \doi{10.2214/AJR.08.2031}.
\newblock URL \url{http://www.ajronline.org/content/193/2/586.abstract}.

\bibitem[He et~al.(2007)He, Yap, Chen, and Chau]{He2007}
Yu~He, Kim-Hui Yap, Li~Chen, and Lap-Pui Chau.
\newblock A nonlinear least square technique for simultaneous image
  registration and super-resolution.
\newblock \emph{Image Processing, IEEE Transactions on}, 16\penalty0
  (11):\penalty0 2830--2841, 2007.
\newblock ISSN 1057-7149.
\newblock \doi{10.1109/TIP.2007.908074}.

\bibitem[Herman(2010)]{Herman2010}
Gabor~T. Herman.
\newblock \emph{Fundamentals of Computerized Tomography: Image Reconstruction
  from Projections}.
\newblock Springer London, second edition, 2010.
\newblock ISBN 978-1-85233-617-2.

\bibitem[Hudson and Larkin(1994)]{Hudson1994}
H.M. Hudson and R.S. Larkin.
\newblock Accelerated image reconstruction using ordered subsets of projection
  data.
\newblock \emph{Medical Imaging, IEEE Transactions on}, 13\penalty0
  (4):\penalty0 601--609, 1994.
\newblock ISSN 0278-0062.
\newblock \doi{10.1109/42.363108}.

\bibitem[Jacobson and Fessler(2003)]{Jacobson2003}
M.W. Jacobson and J.A. Fessler.
\newblock Joint estimation of image and deformation parameters in
  motion-corrected {PET}.
\newblock In \emph{Nuclear Science Symposium Conference Record, 2003 IEEE},
  volume~5, pages 3290--3294, 2003.
\newblock \doi{10.1109/NSSMIC.2003.1352599}.

\bibitem[Kak and Slaney(2001)]{Kak2001}
A.~C. Kak and M.~Slaney.
\newblock \emph{Principles of computerized tomographic imaging}.
\newblock Society for Industrial and Applied Mathematics, Philadelphia, PA,
  USA, 2001.
\newblock ISBN 0-89871-494-X.

\bibitem[Kastanis et~al.(2008)Kastanis, Arridge, Stewart, Gunn, Ullberg, and
  Francke]{Kastanis2008}
Iason Kastanis, Simon Arridge, Alex Stewart, Spencer Gunn, Christer Ullberg,
  and Tom Francke.
\newblock {3D} digital breast tomosynthesis using total variation
  regularization.
\newblock In Elizabeth Krupinski, editor, \emph{Digital Mammography}, volume
  5116 of \emph{Lecture Notes in Computer Science}, pages 621--627. Springer
  Berlin / Heidelberg, 2008.
\newblock URL \url{http://dx.doi.org/10.1007/978-3-540-70538-3_86}.

\bibitem[Kolitsi et~al.(1992)Kolitsi, Panayiotakis, Anastassopoulos, Scodras,
  and Pallikarakis]{Kolitsi1992}
Z.~Kolitsi, G.~Panayiotakis, V.~Anastassopoulos, A.~Scodras, and
  N.~Pallikarakis.
\newblock A multiple projection method for digital tomosynthesis.
\newblock \emph{Medical Physics}, 19\penalty0 (4):\penalty0 1045--1050, 1992.
\newblock \doi{10.1118/1.596822}.
\newblock URL \url{http://link.aip.org/link/?MPH/19/1045/1}.

\bibitem[Matsuo et~al.(1993)Matsuo, Iwata, Horiba, and Suzumura]{Matsuo1993}
H.~Matsuo, A.~Iwata, I.~Horiba, and N.~Suzumura.
\newblock Three-dimensional image reconstruction by digital tomo-synthesis
  using inverse filtering.
\newblock \emph{Medical Imaging, IEEE Transactions on}, 12\penalty0
  (2):\penalty0 307--313, 1993.
\newblock ISSN 0278-0062.
\newblock \doi{10.1109/42.232260}.

\bibitem[Mertelmeier et~al.(2006)Mertelmeier, Orman, Haerer, and
  Dudam]{Mertelmeier2006}
Thomas Mertelmeier, Jasmina Orman, Wolfgang Haerer, and Mithun~K. Dudam.
\newblock Optimizing filtered backprojection reconstruction for a breast
  tomosynthesis prototype device.
\newblock In Michael~J. Flynn and Jiang Hsieh, editors, \emph{Medical Imaging
  2006: Physics of Medical Imaging}, volume 6142, page 61420F. SPIE, 2006.
\newblock \doi{10.1117/12.651380}.
\newblock URL \url{http://link.aip.org/link/?PSI/6142/61420F/1}.

\bibitem[Mueller et~al.(1999)Mueller, Yagel, and Wheller]{Mueller1999}
K.~Mueller, R.~Yagel, and J.J. Wheller.
\newblock Anti-aliased three-dimensional cone-beam reconstruction of
  low-contrast objects with algebraic methods.
\newblock \emph{Medical Imaging, IEEE Transactions on}, 18\penalty0
  (6):\penalty0 519--537, 1999.
\newblock ISSN 0278-0062.
\newblock \doi{10.1109/42.781017}.

\bibitem[Odille et~al.(2008)Odille, Vuissoz, Marie, and Felblinger]{Odille2008}
Freddy Odille, Pierre-André Vuissoz, Pierre-Yves Marie, and Jacques
  Felblinger.
\newblock Generalized reconstruction by inversion of coupled systems ({GRICS})
  applied to free-breathing mri.
\newblock \emph{Magnetic Resonance in Medicine}, 60\penalty0 (1):\penalty0
  146--157, 2008.
\newblock ISSN 1522-2594.
\newblock \doi{10.1002/mrm.21623}.
\newblock URL \url{http://dx.doi.org/10.1002/mrm.21623}.

\bibitem[Poplack et~al.(2007)Poplack, Tosteson, Kogel, and Nagy]{Poplack2007}
Steven~P. Poplack, Tor~D. Tosteson, Christine~A. Kogel, and Helene~M. Nagy.
\newblock Digital breast tomosynthesis: Initial experience in 98 women with
  abnormal digital screening mammography.
\newblock \emph{American Journal of Roentgenology}, 189\penalty0 (3):\penalty0
  616--623, 2007.
\newblock \doi{10.2214/AJR.07.2231}.
\newblock URL \url{http://www.ajronline.org/content/189/3/616.abstract}.

\bibitem[Schumacher et~al.(2009)Schumacher, Modersitzki, and
  Fischer]{Schumacher2009}
H.~Schumacher, J.~Modersitzki, and B.~Fischer.
\newblock Combined reconstruction and motion correction in {SPECT} imaging.
\newblock \emph{Nuclear Science, IEEE Transactions on}, 56\penalty0
  (1):\penalty0 73--80, 2009.
\newblock ISSN 0018-9499.
\newblock \doi{10.1109/TNS.2008.2007907}.

\bibitem[Shepp and Vardi(1982)]{Shepp1982}
L.~A. Shepp and Y.~Vardi.
\newblock Maximum likelihood reconstruction for emission tomography.
\newblock \emph{Medical Imaging, IEEE Transactions on}, 1\penalty0
  (2):\penalty0 113--122, 1982.
\newblock ISSN 0278-0062.
\newblock \doi{10.1109/TMI.1982.4307558}.

\bibitem[Sidky et~al.(2009)Sidky, Pan, Reiser, Nishikawa, Moore, and
  Kopans]{Sidky2009}
Emil~Y. Sidky, Xiaochuan Pan, Ingrid~S. Reiser, Robert~M. Nishikawa, Richard~H.
  Moore, and Daniel~B. Kopans.
\newblock Enhanced imaging of microcalcifications in digital breast
  tomosynthesis through improved image-reconstruction algorithms.
\newblock \emph{Medical Physics}, 36\penalty0 (11):\penalty0 4920--4932, 2009.
\newblock \doi{10.1118/1.3232211}.
\newblock URL \url{http://link.aip.org/link/?MPH/36/4920/1}.

\bibitem[Sinha et~al.(2009)Sinha, Narayanan, Ma, Roubidoux, Liu, and
  Carson]{Sinha2009}
Sumedha~P. Sinha, Ramkrishnan Narayanan, Bing Ma, Marilyn~A. Roubidoux, He~Liu,
  and Paul~L. Carson.
\newblock Image registration for detection and quantification of change on
  digital tomosynthesis mammographic volumes.
\newblock \emph{American Journal of Roentgenology}, 192\penalty0 (2):\penalty0
  384--387, 2009.
\newblock \doi{10.2214/AJR.08.1388}.
\newblock URL \url{http://www.ajronline.org/cgi/content/abstract/192/2/384}.

\bibitem[Spangler et~al.(2011)Spangler, Zuley, Sumkin, Abrams, Ganott, Hakim,
  Perrin, Chough, Shah, and Gur]{Spangler2011}
M.~Lee Spangler, Margarita~L. Zuley, Jules~H. Sumkin, Gordan Abrams, Marie~A.
  Ganott, Christiane Hakim, Ronald Perrin, Denise~M. Chough, Ratan Shah, and
  David Gur.
\newblock Detection and classification of calcifications on digital breast
  tomosynthesis and 2d digital mammography: A comparison.
\newblock \emph{American Journal of Roentgenology}, 196\penalty0 (2):\penalty0
  320--324, 2011.
\newblock \doi{10.2214/AJR.10.4656}.
\newblock URL \url{http://www.ajronline.org/content/196/2/320.abstract}.

\bibitem[Stevens et~al.(2001)Stevens, Fahrig, and Pelc]{Stevens2001}
Grant~M. Stevens, Rebecca Fahrig, and Norbert~J. Pelc.
\newblock Filtered backprojection for modifying the impulse response of
  circular tomosynthesis.
\newblock \emph{Medical Physics}, 28\penalty0 (3):\penalty0 372--380, 2001.
\newblock \doi{10.1118/1.1350588}.
\newblock URL \url{http://link.aip.org/link/?MPH/28/372/1}.

\bibitem[Van~de Sompel et~al.(2011)Van~de Sompel, Brady, and
  Boone]{VandeSompel2011}
Dominique Van~de Sompel, Sir~Michael Brady, and John Boone.
\newblock Task-based performance analysis of {FBP}, {SART} and {ML} for digital
  breast tomosynthesis using signal {CNR} and channelised hotelling observers.
\newblock \emph{Medical Image Analysis}, 15\penalty0 (1):\penalty0 53--70,
  2011.
\newblock ISSN 1361-8415.
\newblock \doi{DOI: 10.1016/j.media.2010.07.004}.
\newblock URL
  \url{http://www.sciencedirect.com/science/article/pii/S1361841510000964}.

\bibitem[Wu et~al.(2004{\natexlab{a}})Wu, Moore, Rafferty, and Kopans]{Wu2004}
Tao Wu, Richard~H. Moore, Elizabeth~A. Rafferty, and Daniel~B. Kopans.
\newblock A comparison of reconstruction algorithms for breast tomosynthesis.
\newblock \emph{Medical Physics}, 31\penalty0 (9):\penalty0 2636--2647,
  2004{\natexlab{a}}.
\newblock \doi{10.1118/1.1786692}.
\newblock URL \url{http://link.aip.org/link/?MPH/31/2636/1}.

\bibitem[Wu et~al.(2004{\natexlab{b}})Wu, Zhang, Moore, Rafferty, Kopans,
  Meleis, and Kaeli]{Wu2004a}
Tao Wu, Juemin Zhang, Richard Moore, Elizabeth Rafferty, Daniel Kopans, Waleed
  Meleis, and David Kaeli.
\newblock Digital tomosynthesis mammography using a parallel maximum-likelihood
  reconstruction method.
\newblock In Martin~J. Yaffe and Michael~J. Flynn, editors, \emph{Medical
  Imaging 2004: Physics of Medical Imaging}, volume 5368, pages 1--11. SPIE,
  2004{\natexlab{b}}.
\newblock \doi{10.1117/12.534446}.
\newblock URL \url{http://link.aip.org/link/?PSI/5368/1/1}.

\bibitem[Yan et~al.(2007)Yan, Ren, Godfrey, and Yin]{Yan2007}
Hui Yan, Lei Ren, Devon~J. Godfrey, and Fang-Fang Yin.
\newblock Accelerating reconstruction of reference digital tomosynthesis using
  graphics hardware.
\newblock \emph{Medical Physics}, 34\penalty0 (10):\penalty0 3768--3776, 2007.
\newblock \doi{10.1118/1.2779945}.
\newblock URL \url{http://link.aip.org/link/?MPH/34/3768/1}.

\bibitem[Yang et~al.(2005)Yang, Ng, and Penczek]{Yang2005}
Chao Yang, Esmond~G. Ng, and Pawel~A. Penczek.
\newblock Unified 3-d structure and projection orientation refinement using
  quasi-newton algorithm.
\newblock \emph{Journal of Structural Biology}, 149\penalty0 (1):\penalty0
  53--64, 2005.
\newblock ISSN 1047-8477.
\newblock \doi{DOI: 10.1016/j.jsb.2004.08.010}.
\newblock URL
  \url{http://www.sciencedirect.com/science/article/pii/S1047847704001716}.

\bibitem[Yang(2012)]{Yang2012c}
Guang Yang.
\newblock \emph{Numerical Approaches for Solving the Combined Reconstruction
  and Registration of Digital Breast Tomosynthesis}.
\newblock Doctoral thesis, UCL (University College London)., Feb 2012.

\bibitem[Yang et~al.(2010{\natexlab{a}})Yang, Hipwell, Clarkson, Tanner,
  Mertzanidou, Gunn, Ourselin, Hawkes, and Arridge]{Yang2010a}
Guang Yang, John Hipwell, Matthew Clarkson, Christine Tanner, Thomy
  Mertzanidou, Spencer Gunn, Sebastien Ourselin, David Hawkes, and Simon
  Arridge.
\newblock Combined reconstruction and registration of digital breast
  tomosynthesis.
\newblock In \emph{Digital Mammography}, volume 6136 of \emph{Lecture Notes in
  Computer Science}, pages 760--768. Springer Berlin/Heidelberg,
  2010{\natexlab{a}}.
\newblock URL \url{http://dx.doi.org/10.1007/978-3-642-13666-5_102}.

\bibitem[Yang et~al.(2010{\natexlab{b}})Yang, Hipwell, Clarkson, Tanner,
  Mertzanidou, Gunn, Ourselin, Hawkes, and Arridge]{Yang2010b}
Guang Yang, John~H. Hipwell, Matthew~J. Clarkson, Christine Tanner, Thomy
  Mertzanidou, Spencer Gunn, Sebastien Ourselin, David~J. Hawkes, and Simon~R.
  Arridge.
\newblock Combined reconstruction and registration of digital breast
  tomosynthesis: Sequential method versus iterative method.
\newblock In \emph{Medical Image Understanding and Analysis}, pages 1--5,
  University of Warwick, Coventry., 2010{\natexlab{b}}.

\bibitem[Yang et~al.(2011)Yang, Hipwell, Hawkes, and Arridge]{Yang2011}
Guang Yang, John~H. Hipwell, David~J. Hawkes, and Simon~R. Arridge.
\newblock Unconstrained simultaneous scheme to fully couple reconstruction and
  registration for digital breast tomosynthesis: A feasible study.
\newblock In \emph{Workshop on Breast Image Analysis}, pages 25--32, In
  conjunction with MICCAI 2011, September 2011.
\newblock ISBN 978-87-981270-9-3.
\newblock URL \url{http://www.diku.dk/BIA2011proceedings.pdf}.

\bibitem[Yang et~al.(2012{\natexlab{a}})Yang, Hipwell, Tanner, Hawkes, and
  Arridge]{Yang2012a}
Guang Yang, John Hipwell, Christine Tanner, David Hawkes, and Simon Arridge.
\newblock Joint registration and limited-angle reconstruction of digital breast
  tomosynthesis.
\newblock In \emph{Digital Mammography, IWDM'12}, Lecture Notes in Computer
  Science. Springer Berlin / Heidelberg, 2012{\natexlab{a}}.
\newblock Submitted.

\bibitem[Yang et~al.(2012{\natexlab{b}})Yang, Hipwell, Hawkes, and
  Arridge]{Yang2012b}
Guang Yang, John~H. Hipwell, David~J. Hawkes, and Simon~R. Arridge.
\newblock A nonlinear least squares method for solving the joint reconstruction
  and registration problem in digital breast tomosynthesis.
\newblock In \emph{Medical Image Understanding and Analysis}, pages 75--80,
  Swansea University, UK, 2012{\natexlab{b}}.

\bibitem[Yap et~al.(2009)Yap, He, Tian, and Chau]{Yap2009}
Kim-Hui Yap, Yu~He, Yushuang Tian, and Lap-Pui Chau.
\newblock A nonlinear {$L_{1}$}-norm approach for joint image registration and
  super-resolution.
\newblock \emph{Signal Processing Letters, IEEE}, 16\penalty0 (11):\penalty0
  981--984, 2009.
\newblock ISSN 1070-9908.
\newblock \doi{10.1109/LSP.2009.2028106}.

\bibitem[Zhang and Brady(2010)]{Zhang2010}
Weiwei Zhang and Sir Brady.
\newblock Feature point detection for non-rigid registration of digital breast
  tomosynthesis images.
\newblock In Joan Martí, Arnau Oliver, Jordi Freixenet, and Robert Martí,
  editors, \emph{Digital Mammography}, volume 6136 of \emph{Lecture Notes in
  Computer Science}, pages 296--303. Springer Berlin / Heidelberg, 2010.
\newblock URL \url{http://dx.doi.org/10.1007/978-3-642-13666-5_40}.

\bibitem[Zhang et~al.(2006)Zhang, Chan, Sahiner, Wei, Goodsitt, Hadjiiski, Ge,
  and Zhou]{Zhang2006}
Yiheng Zhang, Heang-Ping Chan, Berkman Sahiner, Jun Wei, Mitchell~M. Goodsitt,
  Lubomir~M. Hadjiiski, Jun Ge, and Chuan Zhou.
\newblock A comparative study of limited-angle cone-beam reconstruction methods
  for breast tomosynthesis.
\newblock \emph{Medical Physics}, 33\penalty0 (10):\penalty0 3781--3795, 2006.
\newblock \doi{10.1118/1.2237543}.
\newblock URL \url{http://link.aip.org/link/?MPH/33/3781/1}.

\end{thebibliography}
%% \bibliography{<your-bib-database>}

%% not want to use model2-names.bst.



%% \bibitem must have one of the following forms:
%%   \bibitem[Jones et al.(1990)]{key}...
%%   \bibitem[Jones et al.(1990)Jones, Baker, and Williams]{key}...
%%   \bibitem[Jones et al., 1990]{key}...
%%   \bibitem[\protect\citeauthoryear{Jones, Baker, and Williams}{Jones
%%       et al.}{1990}]{key}...
%%   \bibitem[\protect\citeauthoryear{Jones et al.}{1990}]{key}...
%%   \bibitem[\protect\astroncite{Jones et al.}{1990}]{key}...
%%   \bibitem[\protect\citename{Jones et al., }1990]{key}...
%%   \harvarditem[Jones et al.]{Jones, Baker, and Williams}{1990}{key}...
%%


% \end{thebibliography}

%______________________________________________________________________________________________________________________
% End of the Document
\end{document}